%% file: Thesis.tex
\documentclass[a4paper,12pt]{report}

\usepackage{hhline}
\usepackage{soul}
\usepackage{color}
\usepackage{cite}
\usepackage{graphicx}
\usepackage{verbatim}
\usepackage{amsmath}
\usepackage{amssymb}
\usepackage{amsthm}
\usepackage{amsfonts}
\usepackage{paralist}
\usepackage{url}
\usepackage{IEEEtrantools}
\usepackage{fancyhdr}
\usepackage[printonlyused]{acronym}
\usepackage{enumerate}
\usepackage{subfig}
\usepackage{listings}
\usepackage{nomencl}
\usepackage[percent]{overpic}

\makenomenclature

\usepackage{setspace}
\renewcommand{\bibname}{References}

\newtheorem{defin}{Definition}
\newtheorem{thrm}{Theorem}

\renewcommand{\v}[1]{\ensuremath{\text{\bf #1}}}

\begin{document}

\author{Wesam Elshamy}
\title{Using Artificial Intelligence Models in System Identification}
\date{\today}

\include{cover}

\include{abstract}

\pagenumbering{roman} \setcounter{page}{3}

\include{ack}

\tableofcontents

\listoffigures \addcontentsline{toc}{chapter}{\listfigurename}

\listoftables \addcontentsline{toc}{chapter}{\listtablename}

\chapter*{Acronyms\markboth{Acronyms}{Acronyms}}
\printnomenclature

\addcontentsline{toc}{chapter}{Acronyms}

\begin{acronym}[FDR-PSO]%
\acro{ACO}{Ant Colony Optimization}%
\acro{AI}{Artificial Intelligence}%
\acro{ANN}{Artificial Neural Network}%
\acro{ARMA}{Auto-Regressive Moving Average}%
\acro{CI}{Computational Intelligence}%
\acro{C-PSO}{Clubs-based PSO}%
\acro{DAS}{Dominant-alleles-set}%
\acro{DM}{Decision Maker}%
\acro{EA}{Evolutionary Algorithm}%
\acro{EC}{Evolutionary Computation}%
\acro{EMO}{Evolutionary Multi-objective Optimization}%
\acro{EP}{Evolutionary Programming}%
\acro{ES}{Evolutionary Strategies}%
\acro{FDR-PSO}{Fitness-Distance-Ratio PSO}%
\acro{FIPS}{Fully Informed Particle Swarm}%
\acro{FPS}{Fitness Proportionate Selection}%
\acro{GA}{Genetic Algorithm}%
\acro{GP}{Genetic Programming}%
\acro{H-PSO}{Hierarchical PSO}%
\acro{LCS}{Learning Classifier Systems}%
\acro{LQG}{Linear Quadratic Gaussian}%
\acro{LS}{Line Search}%
\acro{MOEA}{Multi-Objective Evolutionary Algorithm}%
\acro{MLP}{Multi-Layered Perceptron}%
\acro{MOP}{Multiobjective Optimization Problem}%
\acro{NARMAX}{Non-linear Auto-Regressive Moving Average model with eXogenous inputs}%
\acro{NFL}{No Free Lunch}%
\acro{NSGA-II}{Non-dominated Sorting Genetic Algorithm II}%
\acro{PF}{Pareto-optimal Front}%
\acro{PID}{Proportional-Integral-Derivative}%
\acro{PSO}{Particle Swarm Optimization}%
\acro{RBF}{Radial Basis Function}%
\acro{RMS}{Root-Mean-Square}%
\acro{SAE}{Sum of Absolute Error}%
\acro{SD}{Steepest Descent}%
\acro{SDS}{Stochastic Diffusion Search}%
\acro{SI}{Swarm Intelligence}%
\acro{SOP}{Single-objective Optimization Problem}%
\acro{SAT}{Boolean Satisfiability Problem}%
\acro{SBX}{Simulated Binary Crossover}%
\acro{TSP}{Traveling Salesman Problem}%
\end{acronym}

\include{intro/intro}

\include{optimization/optimization}

\include{EA/EA}

\include{SI/SI}

\include{ACE/ACE}

\include{conc/conc}

\appendix
\include{app/im}

\pagebreak
\addcontentsline{toc}{chapter}{\bibname}
\bibliography{thesis_bib}
\bibliographystyle{IEEEtran}

\end{document}

%% file: cover.tex
\begin{titlepage}%

\begin{center}
\begin{Large}{\sc Using Artificial Intelligence Models in System Identification}
\end{Large}

by

Wesam Samy Mohammed Elshamy%

\vspace{5ex}%

A Thesis Submitted to the\\
Faculty of Engineering at Cairo University\\
in Partial Fulfillment of the\\
Requirements for the Degree of\\
{\sc Master of Science}\\
in\\
Electrical Power and Machines

\vspace{9ex}

Under the Supervision of\\

\vspace{3ex}

\begin{minipage}[t]{0.45\textwidth}
\centering%
Dr. Ahmed Bahgat Gamal Bahgat\\
{\footnotesize Professor, Department of Electrical Power and
Machines}\end{minipage} \hfill
\begin{minipage}[t]{0.45\textwidth}
\centering%
Dr. Hassan Mohammed Rashad\\
{\footnotesize Assoc. Professor, Department of Electrical Power and
Machines}
\end{minipage}

\vfill

{\sc Faculty of Engineering, Cairo University\\
Giza, Egypt}\\
22 May 2007

\end{center}
\end{titlepage}

%% file: abstract.tex
\begin{abstract}
\pagenumbering{roman} \setcounter{page}{2}
\addcontentsline{toc}{chapter}{\abstractname}%

\ac{AI} techniques are known for its ability in tackling problems
found to be unyielding to traditional mathematical methods. A recent
addition to these techniques are the \ac{CI} techniques which, in
most cases, are nature or biologically inspired techniques.
Different \ac{CI} techniques found their way to many control
engineering applications, including system identification, and the
results obtained by many researchers were encouraging. However, most
control engineers and researchers used the basic \ac{CI} models
\emph{as is} or slightly modified them to match their needs.
Henceforth, the merits of one model over the other was not clear,
and full potential of these models was not exploited.

In this research, \ac{GA} and \ac{PSO} methods, which are different
\ac{CI} techniques, are modified to best suit the multimodal problem
of system identification. In the first case of \ac{GA}, an extension
to the basic algorithm, which is inspired from nature as well, was
deployed by introducing redundant genetic material. This extension,
which come in handy in living organisms, did not result in
significant performance improvement to the basic algorithm. In the
second case, the \ac{C-PSO} dynamic neighborhood structure was
introduced to replace the basic static structure used in canonical
\ac{PSO} algorithms. This modification of the neighborhood structure
resulted in significant performance of the algorithm regarding
convergence speed, and equipped it with a tool to handle multimodal
problems.

To understand the suitability of different \ac{GA} and \ac{PSO}
techniques in the problem of system identification, they were used
in an induction motor's parameter identification problem. The
results enforced previous conclusions and showed the superiority of
\ac{PSO} in general over the \ac{GA} in such a multimodal problem.
In addition, the \ac{C-PSO} topology used significantly outperformed
the two other static topologies in all performance measures used in
this problem.

\end{abstract}

%% file: ack.tex
\chapter*{Acknowledgements}
\addcontentsline{toc}{chapter}{Acknowledgements}

I would like to thank Dr. Ahmed Bahgat for being my teacher and
supervisor.

Special thanks go to Dr. Hassan Rashad who spared no effort in
teaching and supervising me. I learned a lot from the valuable
discussions I had with him and his valuable comments.

Finally I cannot forget the support I had from my family, not only
during my academic studies, but throughout my life. I am deeply
indebted to them.

\vspace{1cm}

\begin{flushright}
  \emph{``We are what we repeatedly do.\\
   Excellence, therefore, is not an act,\\
    but a habit''}\\
  --- Aristotle 384--322 BC
\end{flushright}

%% file: intro/intro.tex
\chapter{Introduction}
\label{ch:intro}

\pagenumbering{arabic}

\section{Motivation}
The field of \ac{AI} inevitably emerged as computers started to find
their way in many applications. Engineers and computer scientists
who worked on developing \ac{AI} models, started using these models
to solve many problems they faced. The success of many \ac{AI}
applications in computer science and computer engineering were
exciting, it became possible to find the meaning of a word according
to its context, computers and machines became able to understand
spoken language to some extent, and recently were used to match
fingerprints. Until recently, many of the \ac{AI} models were
developed to solve problems in computer engineering and computer
science in the first place. These models were later used \emph{as
is} or slightly modified by researchers in other fields.

Control engineers and researchers were enthusiastic about the
results obtained by their fellows in the computer science field.
They were scrutinizing the \ac{AI} models developed by their fellows
because back in their labs, they faced complex problems unyielding
to traditional mathematical techniques. Among these problems is the
system identification problem. The problem of system identification
with its hard nonlinearity, multimodality, and constraints is
especially unsuitable for traditional mathematical techniques, and
the results obtained using these techniques are unsatisfactory for
most real life applications. Henceforth, control engineers started
using the models developed by their fellows to solve system
identification problems. The results obtained were encouraging, and
many \ac{AI} models became the method of choice for many control
engineers.

The author of this thesis belongs to both groups of researchers. He
was unsatisfied by the off-the-shelf \ac{AI} models used by control
engineers, so he used his knowledge in both fields to test, modify,
and develop new models with the problem of system identification in
mind.

A new wave of \ac{AI} models is the \ac{CI} techniques, which are in
most cases, are nature inspired, or biologically inspired
techniques. Recent research have shown promising results in their
applications in many control engineering problems, and specifically,
system identification problems \cite{Fleming01}. Due to their
inherent capability of handling many of the difficulties encountered
in control engineering problems, and because of the encouraging
results reported by many researchers, it was found by the author of
this thesis that developing these techniques is the next logical
step to pursue.

\section{Thesis Outline}
After this brief introduction, the optimization problem with its
various types is presented in Chapter~\ref{ch:OptProb}. After the
Single and Multiobjective problems are presented and their
terminology and definitions are explained in
Section~\ref{ch:opt_sec:SOP} and \ref{ch:opt_sec:MOP}, respectively,
the difficulties faced in solving these problems are detailed in
Section~\ref{ch:opt_sec:diff}.

The thesis moves to non-traditional
techniques by presenting the \acfp{EA} in Chapter~\ref{ch:EAs}.
After an introduction in Section~\ref{ch:EAs_sec:intro}, the logic
behind \acp{EA} and an analysis of their behavior is detailed in
Section~\ref{ch:EAs_sec:HW?}, and an example of these techniques,
which is the \acp{GA}, presented in detail in
Section~\ref{ch:EAs_sec:GA}. Finally, a proposed extension to the
\ac{GA} model is explained, tested on a set of benchmark problems,
and a conclusion about its efficiency is given in
Section~\ref{ch:EAs_sec:poly}.

More \ac{CI} models follow in Chapter~\ref{ch:SI} where the \ac{SI}
methods are presented. The chapter starts by an introduction then
explains some theoretical aspects about these techniques in
Section~\ref{ch:SI_sec:HW?}. An example of the \ac{SI} techniques
which is the \ac{PSO} is presented followed by some of its
variations in Section~\ref{ch:SI_sec:PSO} and \ref{ch:SI_sec:var},
respectively. A proposed modification to some aspects of the
\ac{PSO} model is explained, analyzed, tested on many test problems,
and the results were furnished and followed by a conclusion in
Section~\ref{ch:SI_sec:C-PSO}. Finally, a comparison between the
\ac{GA} and \ac{PSO} models concludes the chapter.

Chapter~\ref{ch:EA_pro} is concerned about the applications of
\ac{CI} techniques in control engineering, and particularly in
system identification. It starts by outlining the advantages and
disadvantages of using \ac{CI} methods in control engineering in
Section~\ref{ch:ACE_sec:why} and \ref{ch:ACE_sec:optout},
respectively. Different applications in control engineering are
presented in Section~\ref{ch:ACE_sec:apps} and the problem of system
identification is given in more detail in Section~\ref{ch:sysident}.
The problem of parameter identification of an induction motor is
explained, its model is driven, the algorithms used to solve it are
presented, and the experiment is carried out with its results
explained and a conclusion is furnished in
Section~\ref{ch:ACE_sec:paridentind}.

Chapter~\ref{ch:conc} concludes this thesis by analyzing the results
reached in previous chapters and proposes future research
directions.

%% file: optimization/optimization.tex
\chapter{The Optimization Problem}
\label{ch:OptProb}

The Optimization Problem is encountered in every day life. A man
driving to work usually chooses a route that minimizes his travel
time. An investor makes many decisions on daily basis to minimize
his business risks and increase his profits. Even electrons tend to
occupy the lowest energy level available \cite{bohr1913}. The
problem of optimization becomes a matter of life or death in some
situations. A bad utilization of energy or food reserves of a nation
may lead to crises and loss of life.

The Optimization Problem could be as simple as shopping for a good
looking shirt with a reasonable price, or as complicated as
scheduling air flights for a major airline company. Although those
two examples are at the extremes, they do have the characteristics
of the optimization problem. Both of them have \emph{objectives};
the objectives of the first problem is to find and buy a shirt that
\emph{looks as good as possible} and \emph{is as cheap as possible},
while for the second problem the objective is to \emph{increase
profits as much as possible}. Each one of those problems has
\emph{parameters} or \emph{decision variables} which by tuning them
properly the \emph{objectives} are optimized. For the first problem
these \emph{parameters} include the shop location, brand name and
shirt fabric etc., while for the second problem these
\emph{parameters} include flight destinations, ticket prices, pilots
and crew salaries and flights schedule among many others.
Figure~\ref{fig:dec_obj_sps} shows the mapping between the decision
space, which contains the \emph{parameters}, and the objective
space. Based on this mapping, the \ac{DM} chooses the parameters,
which make-up the decision vector, that maps to the desired point in
the objective space. Neither the decision space nor the objective
space has to be continuous or connected. Optimization problems can
be classified into two categories regarding the number of
objectives;
\begin{inparaenum}[i)]
\item \acp{SOP} and
\item \acp{MOP}
\end{inparaenum}

\section{\aclp{SOP}}
\label{ch:opt_sec:SOP}

\aclp{SOP}, as their name implies, are problems that have single
objective to be optimized (maximized or minimized) by varying their
parameters. A \ac{SOP} can be defined as follows \cite{Liu03,
Zitzler99f}:
\begin{defin}[\acl{SOP}]
\begin{align}
\text{optimize}\quad &y=f(\v{x}) \in \v{Y} \\
           s.t.\quad &h_i(\v{x})= 0   \quad i=1,2\dotsc m\\
               \quad &g_j(\v{x})\ge 0 \quad j=1,2,\dotsc l\\
               \quad &\v{x}=[x_{1}\dotsm x_{n}]^{T} \in \v{X}&
\end{align}
where $x_i$ is a \emph{decision variable}, \v{x} is a \emph{decision
vector}, \v{X} is the \emph{decision space}, $y$ is an
\emph{objective function}, \v{Y} is the \emph{objective space} and
$h_i(\v{x})$ and $g_j(\v{x})$ are equality and inequality constraint
functions respectively.
\end{defin}
The constraint functions determine the \emph{feasible set}.
\begin{defin}[Feasible Set]
\label{defin:feas_set} The \emph{feasible set} $\v{X}_f$ is the set
of \emph{decision vectors} \v{x} that satisfy the constraints
$h_i(\v{x})$ and $g_j(\v{x})$.\\
The image of the \emph{feasible set} in the \emph{objective space}
is known as the \emph{feasible region}.
\end{defin}

\begin{figure}
\hfill
\begin{minipage}[t]{0.45\textwidth}
\centering
\includegraphics{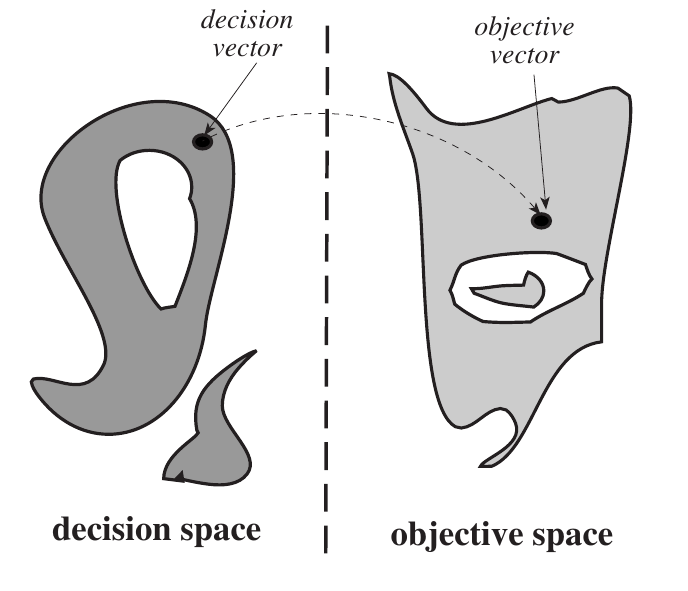}
\caption{Decision space---objective space mapping
\label{fig:dec_obj_sps}}
\end{minipage}
\hfill
\begin{minipage}[t]{0.45\textwidth}
\centering
\includegraphics{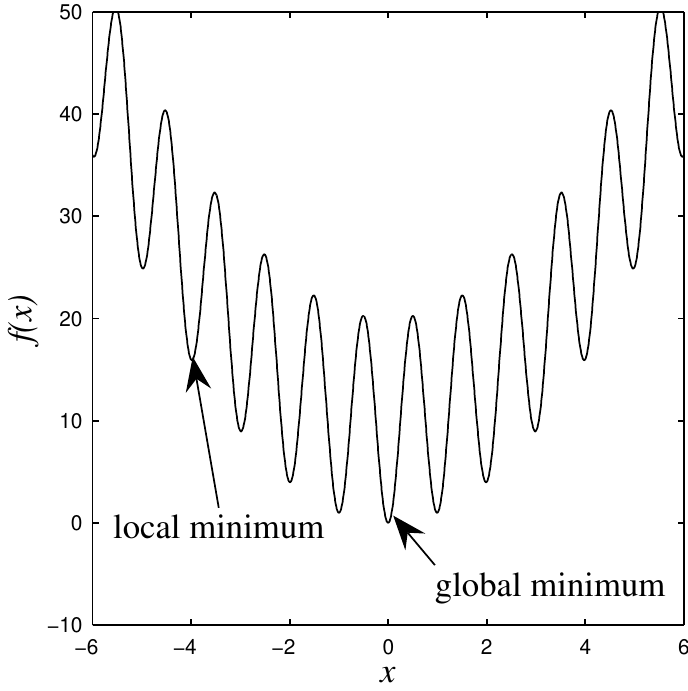}%
\caption[Landscape of Rastrigin problem]{Landscape of Rastrigin
problem ($n=1$) \label{fig:Rast_land}}
\end{minipage}
\hfill
\end{figure}

A well known \ac{SOP} is the Rastrigin test problem
\cite{toern:1989, Muhlenbein91a}. It is defined as follows:
\begin{align}
\text{minimize} \quad &f(\v{x})=\sum_{i=1}^{n}\left[x_i^2 -
10\cos(2\pi
x_i)+10\right]\\
\text{s.t.} \quad &\v{x}\in[-6,6]^n \label{eq:opt_rast_range}
\end{align}
where $n$ is the number of decision variables.
Figure~\ref{fig:Rast_land} shows the landscape of this problem for
$n=1$. The minimum value of the objective function in the feasible
region is achieved at the \emph{global minimum}, or more generally,
the \emph{global optimum}.
\begin{defin}[Global Optimum]
\label{defin:Glob_opt}%
A global optimum is a point in the feasible region whose value is
better than all other points in that region
\end{defin}
For the Rastrigin problem shown in Figure~\ref{fig:Rast_land} the
global minimum is achieved at $x=0$ with a value of $f(\v{x})=0$.
Excluding the valley where the global minimum lies at its bottom,
there are 12 valleys in this problem's landscape. The point at the
bottom of each one of them is known as a \emph{local minimum}, or
more generally, a \emph{local optimum}.
\begin{defin}[Local Optimum]
\label{defin:Loc_opt}%
A local optimum is a point in the feasible region whose value is
better than all other points in its vicinity in the region, and is
worse than the global optimum.
\end{defin}

\section{\aclp{MOP}}
\label{ch:opt_sec:MOP}

Unlike \acp{SOP}, \acp{MOP} have many objectives to be optimized
concurrently, and most of the time these objectives are conflicting.
A \ac{MOP} can be defined as follows \cite{Liu03, Zitzler99f}:
\begin{defin}[\acl{MOP}]
\begin{align}
\text{optimize} \quad &\v{y} = \v{F}(\v{x}) = \{f_{1}(\v{x}), f_{2}(\v{x}),\dotsc, f_{k}(\v{x})\} \in \v{Y}\\
        s.t.    \quad &\v{x} \in \v{X}\\
                \quad &\v{X} = \left\{ \; \v{x} \left| \begin{aligned}&h_{i}(\v{x}) = 0 \quad i = 1,\dotsc m\\
&g_{j}(\v{x}) \le 0 \quad j = 1,\dotsc l\\
&\v{x}=[x_{1}\dotsm x_{n}]^{T}
\end{aligned}\right. \right\}
\end{align}
where $x_{i}$ is a \emph{decision variable}, \v{x} is a
\emph{decision vector}, \v{X} is the \emph{decision space}, \v{y} is
a vector of $k$ objective functions, \v{Y} is the \emph{objective
space} and $h_{i}(\v{x})$ and $g_{j}(\v{x})$ are equality and
inequality constraint functions respectively.
\end{defin}

This definition can be illustrated using the following classic
example \cite{Schaffer85a, VeldhuizenL99}:

\begin{align}
\text{minimize} \quad &\v{F}(\v{x})=\{f_1(\v{x}), f_2(\v{x})\}, \quad \text{where} \label{eq:opt_MOP_F(x)}\\
\quad &f_1(\v{x})= x^{2} \label{eq:opt_f_1}\\
\quad &f_2(\v{x})=(x-2)^{2} \label{eq:opt_f_2}
\end{align}

\begin{figure}
\hfill
\begin{minipage}[t]{0.45\textwidth}
\centering
\includegraphics{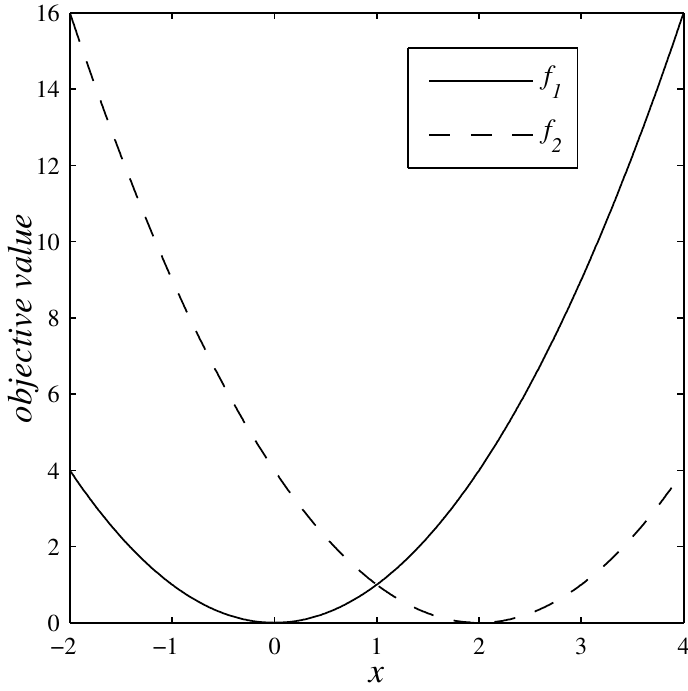}
\caption{Decision parameter's value against Objectives' values
\label{fig:opt_MOP_f1f2_vs_x}}
\end{minipage}
\hfill
\begin{minipage}[t]{0.45\textwidth}
\centering
\includegraphics{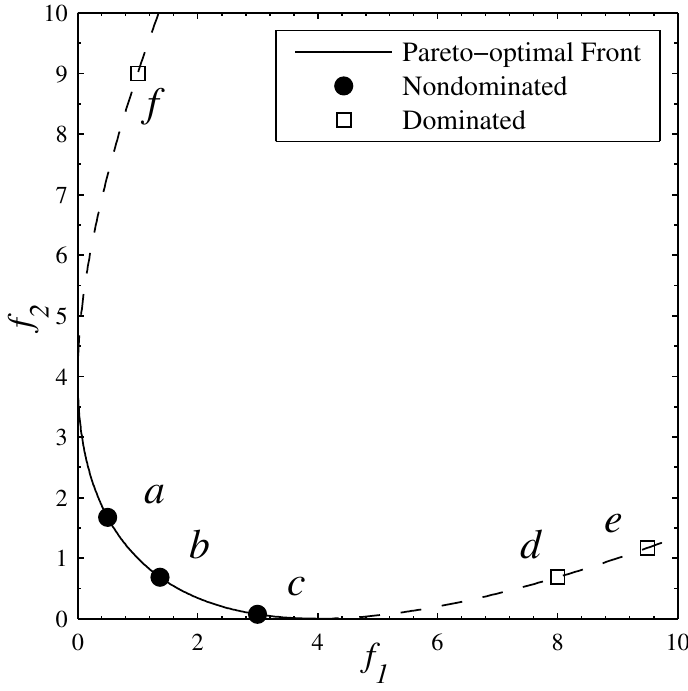}
\caption{Objective space of a MOP \label{fig:opt_MOP_obj}}
\end{minipage}
\hfill
\end{figure}

Figure~\ref{fig:opt_MOP_f1f2_vs_x} shows the values of the objective
functions $f_1$ and $f_2$ while varying the decision variable $x$
value. $f_1$ and $f_2$ are monotonically decreasing with $x$ in the
range $x\in(-\infty,0)$, so the two objectives are in \emph{harmony}
\cite{Purshouse03}, which means that an improvement in one of them
is rewarded with simultaneous improvement in the other. The higher
the value of $x$ in this range the better (the lower) the value of
the two objectives become. A similar situation happens in the range
$x\in(2,\infty)$. The two objective functions are in \emph{harmony}
and monotonically decreasing with $x$ in this range; the smallest
possible value of $x$ in this range is translated to the best (the
lowest) value for the two objectives in that range. However, The two
objectives are in \emph{conflict} in the range $x\in[0,2]$; An
increase in $x$ value is accompanied by improvement of $f_1$ and
deterioration of $f_2$.

A mapping of Figure~\ref{fig:opt_MOP_f1f2_vs_x} to the objective
space gives Figure~\ref{fig:opt_MOP_obj}. The regions
$x\in(-\infty~0)$, $x\in[0~2]$ and $x\in(2~\infty)$ in
Figure~\ref{fig:opt_MOP_f1f2_vs_x} are mapped to the upper-left
dashed segment, solid segment and the lower-right dashed segment in
Figure~\ref{fig:opt_MOP_obj} respectively. It is clear that points
\v{d}, \v{e} and \v{f} are in the harmony region, while points
\v{a}, \v{b} and \v{c} are in the conflict region.

Using logical comparison, the following relationships are
established: Objective vector \v{b} is better than objective vector
\v{d} because although they have the same $f_2$ value, \v{b} has a
lower $f_1$ value. Objective vector \v{c} is better than objective
vector \v{d} as well; it has lower $f_1$ and $f_2$ values. The
following definition is used to put these relationships in
mathematical notation \cite{Zitzler99f}.

\begin{defin}
For any two objective vectors $\boldsymbol{u}$ and $\boldsymbol{v}$,
\begin{alignat}{4}
\v{u} & =   \v{v} &\quad&\text{iff} &\quad&\forall i \in \{1,2,\dotsc, k \}: &\quad& u_i =   v_i\\
\v{u} & \le \v{v} &&\text{iff} &&\forall i \in \{1,2,\dotsc, k \}: && u_i \le v_i\\
\v{u} & <   \v{v} &&\text{iff} &&\v{u}\le \v{v} \wedge \v{u}\ne
\v{v} &&
\end{alignat}
\end{defin}

In a minimization \ac{SOP}, a solution \v{r} is better than a
solution \v{s}, \emph{iff} $p(\v{r})<p(\v{s})$, where $p$ is the
objective function. But in a \ac{MOP} this comparison mechanism does
not hold because there are more than one objective to be
concurrently optimized. In Figure~\ref{fig:opt_MOP_obj}, $c<d$ and
$c\nless a$, but $a\nless d$. This may seem illogical when using
\acp{SOP} reasoning, but in \acp{MOP} this situation is quite
common. The points $a$ and $c$ represent two different solutions yet
none of them is superior to the other; although the solution
represented by $a$ has lower $f_1$ value than that of $c$, the
solution represented by $c$ has lower $f_2$ value than that of $a$.
This means that a new relationship is needed to compare two
different decision vectors \v{a} and \v{b} in a \ac{MOP} when
$\v{F}(\v{a}) \nleq \v{F}(\v{b}) \wedge \v{F}(\v{b}) \nleq
\v{F}(\v{a})$. This relationship is described using \emph{Pareto
Dominance} \cite{Zitzler99f, Hwang79:MultiOpt, Ste:86}.
\begin{defin}[Pareto Dominance]
For any two decision vectors \v{a} and \v{b} in a minimization
problem without loss of generality,
\begin{alignat}{5}
\v{a} &\prec   \v{b} &\quad&(\v{a} \text{\emph{ dominates }} \v{b})        &\quad&\text{iff} &\quad&\v{F}(\v{a}) <     \v{F}(\v{b})\\
\v{a} &\preceq \v{b}      &&(\v{a} \text{\emph{ weakly dominates }} \v{b})      &&\text{iff}      &&\v{F}(\v{a}) \leq  \v{F}(\v{b})\\
\v{a} &\sim    \v{b}      &&(\v{a} \text{\emph{ is indifferent to }}
\v{b})     &&\text{iff} &&\v{F}(\v{a}) \nleq \v{F}(\v{b}) \wedge
\v{F}(\v{b}) \nleq \v{F}(\v{a})
\end{alignat}
\end{defin}

Pareto Dominance is attributed to the Italian sociologist, economist
and philosopher Vilfredo Pareto (1848--1923) \cite{pareto:1896:cdp}.
It is used to compare the \emph{partially ordered} solutions of
\acp{MOP}, compared to the \emph{completely ordered} solutions of
\acp{SOP}. Using Pareto dominance to compare solutions represented
in Figure~\ref{fig:opt_MOP_obj}, the following relationships are
established; \v{b} \emph{dominates} \v{d} because they have the same
$f_2$ value, and \v{b} has a lower $f_1$ value, while \v{b} \emph{is
indifferent to} \v{a} because although \v{b} has a lower $f_2$
value, \v{a} has a lower $f_1$ value. The solutions represented by
\v{a}, \v{b} and \v{c} are known as \emph{Pareto Optimal} solutions.
These solutions are optimal in the sense that none of their
objectives can be improved without simultaneously degrading another
objective.

\begin{defin}[Pareto Optimality]
A decision vector $\v{x}\in\v{X}_f$ is said to be nondominated
regarding a set $\v{A}\subseteq \v{X}_f$ iff
\begin{equation}
\nexists \: \v{a} \in \v{A} : \v{a} \prec \v{x}
\end{equation}
If it is clear within the context which set \v{A} is meant, it is
simply left out. Moreover, \v{x} is said to be \emph{Pareto optimal}
iff \v{x} is nondominated regarding $\v{X}_f$
\end{defin}

By applying this definition to the example presented in
Figure~\ref{fig:opt_MOP_obj}. It is clear that the solutions
represented by $d$, $e$ and $f$ are \emph{dominated}, while $a$, $b$
and $c$ are nondominated. The set of all Pareto-optimal solutions is
known as the \emph{Pareto-optimal set}, and its image in the
objective space is known as the \emph{\ac{PF}}.

\begin{defin}[Nondominated Sets and Fronts]
Let $\v{x}\subseteq \v{X}_f$. The function $p(\v{A})$ gives the set
of nondominated decision vectors in $\v{A}$:
\begin{equation}
p(\v{A})=\{\v{a}\in \v{A} | \v{a} \text{ is nondominated regarding }
\v{A}\}
\end{equation}
The set $p(\v{A})$ is the \emph{nondominated set} regarding \v{A},
the corresponding set of objective vectors $\v{f}(p(\v{A}))$ is the
\emph{nondominated front} regarding \v{A}. Furthermore, the set
$\v{X}_p=p(\v{X}_f)$ is called the \emph{Pareto-optimal set} and the
set $\v{Y}_p=\v{f}(\v{X}_p)$ is denoted as the \emph{Pareto-optimal
front}.
\end{defin}

The Pareto-optimal set of the example given in
Equations~(\ref{eq:opt_MOP_F(x)})--(\ref{eq:opt_f_2}) is
$x\in[0~2]$, and its corresponding image in the objective space is
the \ac{PF} shown as a solid segment in
Figure~\ref{fig:opt_MOP_obj}.

\begin{figure}
  \centering
  \includegraphics[width = 0.7\textwidth]{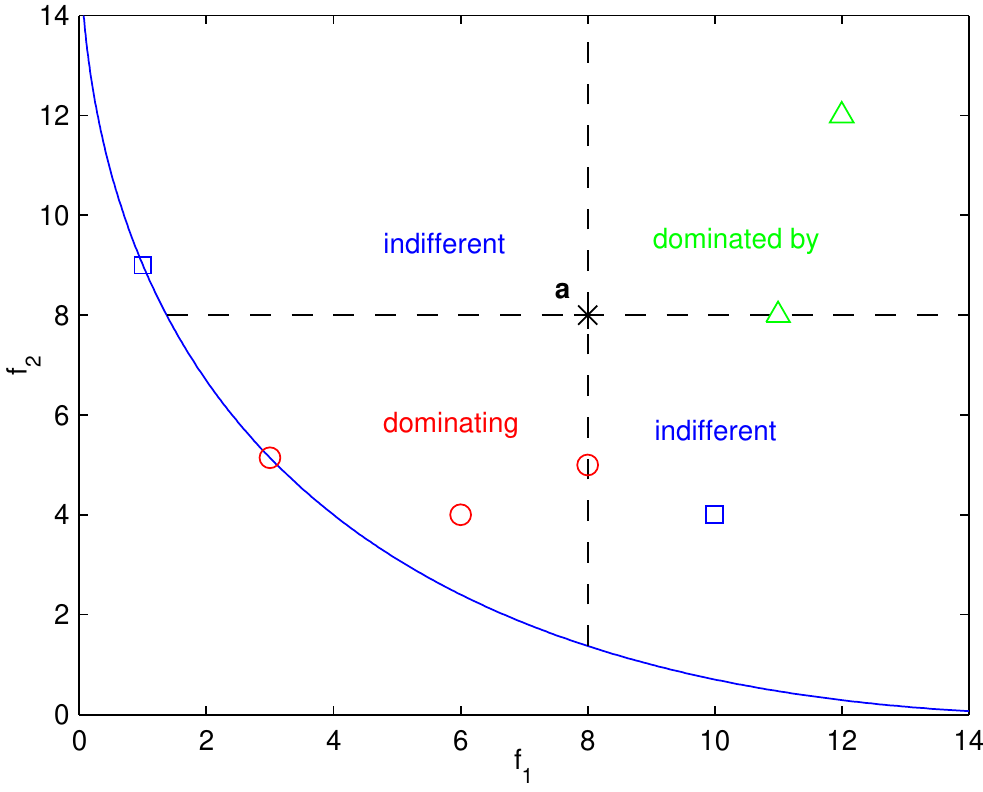}
  \caption{Pareto dominance relationships \label{fig:opt_pareto}}
\end{figure}

To further explain the Pareto dominance relationships, the example
in Figure~\ref{fig:opt_pareto} is given where the two objectives
$f_1$ and $f_2$ are to be minimized. The relationships given in this
example show how the solution vector represented by the point $a$ at
position (8,8) is seen by the other solution vectors. So, the
objective space is divided by a vertical and a horizontal lines
passing through point $a$ into four regions.

\begin{itemize}
\item[-] All the points in the first region at the northeast of point $a$
(including the borderlines) have higher $f_1$ or $f_2$ values, or
both, so point $a$ is \emph{better} than them and these points are
\emph{dominated by} point $a$.

\item[-] All the points in the second and fourth regions at the
southeast and northwest of point $a$ (excluding the borderlines)
have lower values of one objective and higher values for the other
objective compared to the objective values of point $a$. So, All the
points in these two regions are \emph{indifferent} to point $a$.
Note that although the point at position (1,9) is on the Pareto
front while point $a$ is far from that front, they are
\emph{indifferent} to each other.

\item[-] All the points on the third region at the southwest of point
$a$ (including the borderlines) have lower $f_1$ or $f_2$ values, or
both. Which means that they are \emph{better} than point $a$ and
they are \emph{dominating} point $a$.
\end{itemize}

\section{Difficulties in Optimization Problems}
\label{ch:opt_sec:diff}

A researcher or a problem solver could easily be overwhelmed by the
great number of algorithms found in literature dealing with
optimization. Some of them dated back to 1600 BC
\cite{wiki:alg_tmln}, while others are being developed as this text
is being typed. These myriads of algorithms are developed to deal
with different types of difficulties in optimization problems. The
performance of each these optimizers depends on the characteristics
of the problem it optimizes such as being linear/nonlinear,
static/dynamic, \ac{SOP}/\ac{MOP},
combinatorial/discrete/continuous, types and number of constraints,
size of the search space\ldots etc. According to the \ac{NFL}
theorems \cite{Wolpert:1995:NFL}, all algorithms perform exactly the
same when averaged over all possible problems. So, as much as
possible knowledge about the problem should be incorporated in
selecting the problem optimizer, because, according to \ac{NFL}
theorems, there is no such algorithm that will perform better than
all other algorithms on all possible problems. The first step in
understanding optimization problems' characteristics is to answer
the question: \emph{Why are some problems difficult to
solve?}\cite{Michal00, macready:1995, jones95fitness,
stadler95towards, naudts00comparison, jel99c, mitchell92royal,
forrest93what, wright00computational, He:2004, Guo:2003:gecco}

\subsection{The Search Space}
\label{ch:opt_sec:dif_subsec:srch}

The search space of a problem is the set of all possible solutions
to that problem. To tune a radio set to a station, a reasonable man
may work out an exhaustive search by scanning the entire available
bandwidth in his set, but this reasonable man will never resort to
exhaustive search to find the best medication for his heart, or to
look up a word in a dictionary; For the medication case, the penalty
of trying-out all possibilities is too high, it may lead to certain
death, while exhaustively looking up a word in a dictionary takes a
lot of time\footnote{Oxford English Dictionary contains over half a
million words}. To imagine how exhaustive search can easily be a
laborious task, the following example is given.

A classic and one of the most used combinatorial optimization
problems in \ac{AI} literature is the \ac{SAT}
\cite{Davis:Putman:1960}. The problem is to make a compound
statement of boolean variables evaluate to TRUE \cite{Michal00}. For
example, find the truth assignment for the variables $[x_1,
x_2,\cdots, x_{100}]$ that evaluate the following function to TRUE:

\begin{equation}
F(\v{x})=(x_{12} \lor \overline{x}_3 \lor x_{59}) \land
(\overline{x}_{73} \lor x_{28} \lor \overline{x}_9) \land \dotsm
\land (\overline{x}_{69} \lor x_{92} \lor x_5)
\end{equation}
where $\overline{x}_i$ is the complement of $x_i$. In real world
situations, such a function with 100 variables is a reasonable one,
yet its search space is extremely huge; There are $2^{100}$ possible
solutions for this problem. Given a computer that can test 1000
solutions of this problem per second, and that it has started its
trials at the beginning of time itself, 15 billion years ago, it
would have examined less than one percent of all possibilities by
now \cite{Michal00}.

The \ac{SAT} problem is a combinatorial optimization problem, which
means that its search space contains a finite set of solutions, and
a problem solver can theoretically test all of them. But the
Rastrigin problem given earlier in this chapter is a problem with a
continuous search space, which means there are an infinite number of
possible solutions, there is no way to test them all.

\begin{figure}
\centering
\includegraphics{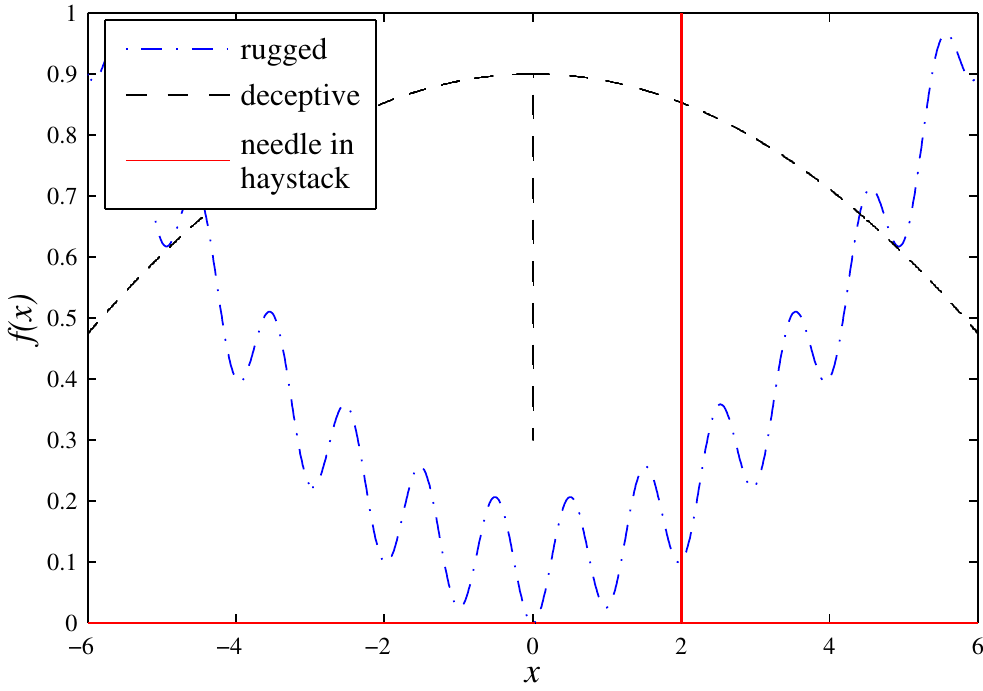}
\caption{Landscapes of different optimization
problems\label{fig:opt_land_typ}}
\end{figure}

Because most people are not willing to wait for a computer to
exhaustively search for a solution to the given \ac{SAT} problem,
other search techniques have been devised to facilitate the search
task. For the radio tuning example given earlier, one can make use
of the feedback sound he gets from the radio to narrow down the
search space and fine-tune the set. But for the given \ac{SAT}
problem this mechanism is useless because the feedback, which is the
value of $F(\v{x})$, is always FALSE except for a single solution
resulting in TRUE output. Such problems are known as \emph{needle in
a haystack} problems \cite{jones95fitness} as shown in
Figure~\ref{fig:opt_land_typ}, where $f(x)$ equals zero in the range
$x\in[-6,6]$ except for a single solution ($x=0$) where $f(x)=1$.

The landscape of the problem could be more difficult than the
\emph{needle in a haystack} case. It could be a \emph{misleading} or
\emph{deceptive} landscape \cite{goldberg87, whitley91fundamental,
Goldberg92c, grefenstette93deception}. In this situation the
feedback (fitness value) drives the problem solver away from the
global minimum, as shown in Figure~\ref{fig:opt_land_typ}. While
searching this landscape for the global minimum, a higher value of
$x$ is rewarded by a lower value of $f(x)$ when $x>0$, and a lower
value of $x$ is rewarded by a lower value of $f(x)$ when $x<0$. This
reward encourages the problem solver to move away from the global
minimum at $x=0$

Another type of difficulty regarding the search space is the
landscape \emph{multimodality} \cite{horn95genetic, Goldberg92c,
SinghDeb06} or \emph{ruggedness} \cite{kolarov97landscape,
kallel01properties}. A rugged landscape may trap the problem solver
in a local minimum, as shown in Figure~\ref{fig:opt_land_typ}. For
this minimization problem, a problem solver can easily get trapped
in any of the local minima, and the odds are high that it will fall
in another one if it managed to escape the first.

\subsection{Modeling the Problem}
The fist step in solving an optimization problem is building a model
for it. After this step, the real problem is put aside and the
problem solver becomes concerned with the model not the problem
itself. So a solution of an optimization problem is a solution to
its \emph{model} \cite{Michal00}, and an optimal solution to
inaccurate model, is a right solution to the wrong problem.
Inaccuracies arise from wrong \emph{assumptions} and
\emph{simplifications}.

For example, a major airline company is deciding on its carriers
destinations, flights schedule and pricing policy. In this
complicated task, the company may \emph{assume} that customers in
Brazil and Argentina are willing to pay the same price for the same
service because they have similar average national income. But this
\emph{assumption} is wrong because customers in Argentina enjoy a
high quality flights with a competitive price on their national
airline company. After the company has considered all factors to
build a model for their problem, its highly likely that this model
will be extremely complicated to be solved by most optimization
tools. So the company is faced with one of two possibilities
\cite{Michal00}.

\begin{enumerate}[i)]
\item Find a \emph{precise} solution to a \emph{simplified} version of the
model.
\item Find an \emph{approximate} solution to the \emph{precise} model.
\end{enumerate}

The first method uses traditional optimization techniques, such as
linear or dynamic programming, to find a solution to the approximate
model. while the second method uses non-traditional techniques, such
as \acp{EA}, \ac{PSO} and \ac{ACO}, to find an approximate solution
to the precise model.

\subsection{Constraints}
Another source of difficulty in optimization problems is the
constraints imposed on the problem. At first glance, these
constraints may be seen as an aid to the problem solver because they
do limit the search space. But in many cases they become a major
source of headache and make it hard to find a single feasible
solution, let alone an optimal one. The following example borrowed
from \cite{Michal00} is used for illustration:

A common problem found in all universities is making a timetable for
classes offered to students each semester. The first step in solving
this problem is collecting the required data, which includes,
courses offered and students registered for them, professors
teaching these courses and their assistances, tools required for
instruction such as projectors, computers and special blackboards,
available laboratories \ldots etc. The next step is to define the
\emph{hard constraints}, which are constraints that must all be met
by a solution to be a feasible solution. These constraints may
include:
\begin{itemize}
\item[-] Every class must be assigned to an available room that has
enough seats for all students and has all the tools required for
students to carry experiments if any, and all special instruction
tools.
\item[-] Students enrolled in more than one course must not have
their courses held at the same time.
\item[-] Professors must not have two classes with overlapping time.
\item[-] Classes must not start by 8 a.m. and must not end after 10
p.m.
\item[-] Classes for the Fall semester must not start by September
$1^{\text{st}}$ and must not end after December $31^{\text{st}}$.
\end{itemize}

These constraints are \emph{hard} in the sense that violating them
will severely hinder the education process. However in addition to
these \emph{hard} constraints, there are \emph{soft} constraints
which a solution that violates any, some or even all of them would
still be feasible, but it's highly encouraged to satisfy them. These
constraints may include:
\begin{itemize}
\item[-] Its preferable that undergraduate classes be held from 8 a.m.
to 4 p.m., while postgraduate classes be held from 4 p.m. to 9 p.m.
\item[-] Its preferable that lectures be held before their
corresponding exercise classes.
\item[-] Its preferable that students have more than 3 classes and
less than 9 classes per day.
\item[-] Its desirable not to roam students across opposite ends of
the campus to take their classes.
\item[-] Its preferable not to assign more than 5 classes for each professor per
day.
\end{itemize}

Although these constraints are all \emph{soft} constraints, some of
them are more important than others. For example, its more important
not to hold an exercise class before its corresponding lecture than
having to make students go to opposite ends of campus. To achieve
this, each constraint is assigned a weight that reflects its
importance and acts as a penalty in the \emph{fitness function}.

After all the constraints have been defined, the problem of finding
a solution that meets all the hard constraints and optimizes the
soft constraints becomes a challenging task indeed.

\subsection{Change Over Time}
Almost all real-world problems are dynamic problems, they change
over time. The only thing fixed about them is the fact that they are
in continuous change. A man neglecting this fact is like a stock
exchange investor who assumes that share prices will remain fixed or
a man who assumes that the weather will always be perfect for a
picnic. To illustrate the significance of change over time, the
stock investor example will be further extended.

Kevin is a stock exchange investor who must realize that the stock
market is extremely dynamic. Share prices change every minute
affected by many factors.

Some factors are hard to predict, for example, the share price of a
company that is announcing its profits/losses is expected to rise or
fall. To simplify the situation, one of two possibilities must
happen. The company announces good profits so its share price will
rise from 60\$ to reach 110\$, or it will announce losses and its
share price will fall from 60\$ to 10\$. Assuming that Kevin knows
about these two possibilities, he is left with a hard decision,
either to sell his shares of this company or to buy more. Trying to
rely on statistics he may average the two possibilities; $\frac{10 +
110}{2} = 60\$$, which means that the share price will remain fixed,
and this is definitely not going to happen. However, some other
factors are biased and are predictable to some extent. A company
achieving high profits for the past ten years is expected to
announce profits for the current year and, consequently, its share
price will increase.

Some events which affect the stock market are associated with
particular periods of time; During Christmas and new year's holidays
share prices almost always fall, because investors need cash to
celebrate. A few days later share prices rise again.

On the other hand, some events are nonpredictable at all. A
terrorist attack on an oil field in Saudi Arabia or a fire that
break out in an oil refinery in Nigeria will cause oil prices to
soar. While a discovery of huge oil reserve in Venezuela will cause
prices to fall.

Although the above factors may act against Kevin sometimes, they are
not conspiring against him, but other investors do. Kevin is faced
with many investors who are acting against him because every penny
he makes is subtracted from their profits. He has to be aware that
the decisions he make are reciprocated by other investors' decisions
that ruin his gains and make his life harder. He in return has to
act in return and wait for their action, and so on.

\subsection{Diversity of Solutions}
A source of difficulty which is unique to \acp{MOP} is the diversity
of solutions. In addition to the above difficulties, the problem
solver has to present solutions that cover the entire \ac{PF},
moreover these solutions should be uniformly distributed over that
front to give the \ac{DM} a flexibility in decision making.

For example, the head of the design department in a mobile phone
manufacturing factory has asked four design engineers each to
present 21 phone speaker designs that adhere to speaker
manufacturing standards but have a varying trad-off between speaker
size and its cost. These four groups of 21 designs each are to be
handled to the marketing department to choose a single design to be
manufactured.

\begin{figure}
\centering%
\subfloat[non-uniform---biased to one
end]{\includegraphics{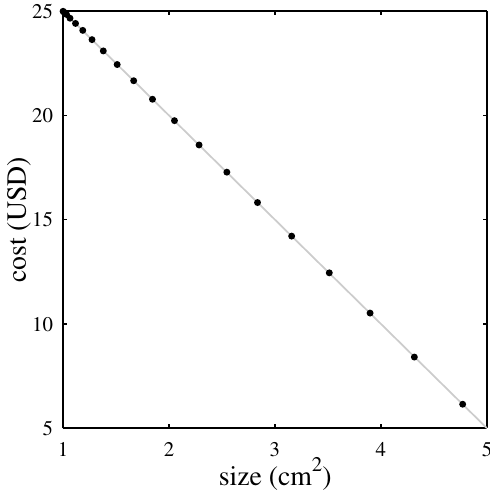}
\label{fig:opt_MOP_dist_a}}
\qquad %
\subfloat[non-uniform---biased to both ends]{\includegraphics{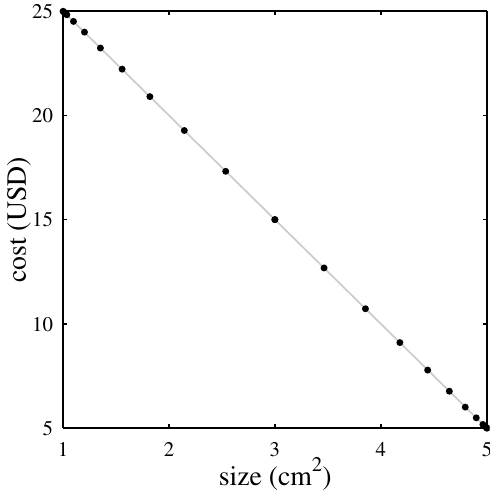} \label{fig:opt_MOP_dist_b}} \\
\subfloat[uniform---concentrated in the
middle]{\includegraphics{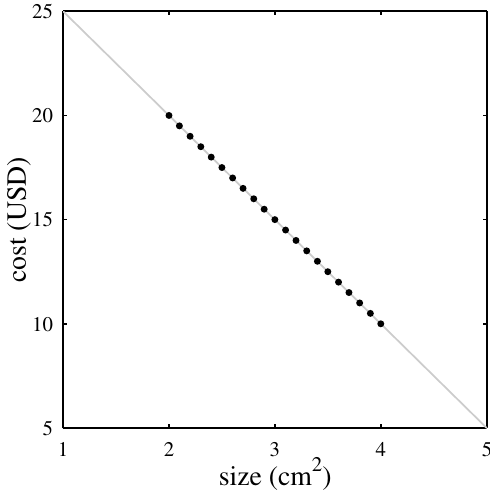}
\label{fig:opt_MOP_dist_c}}
\qquad %
\subfloat[uniform and cover the entire \ac{PF}]{\includegraphics{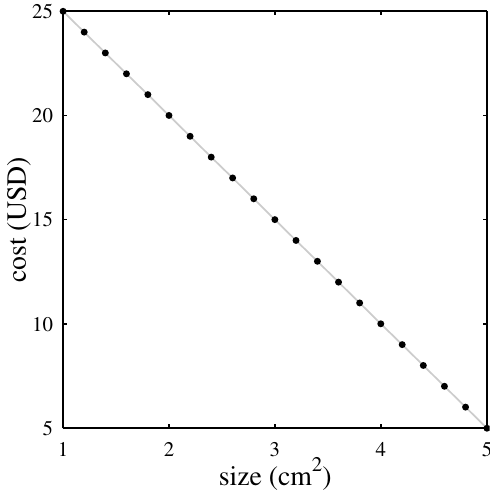} \label{fig:opt_MOP_dist_d}}%
\caption{Different distributions of solutions across the PF
\label{fig:opt_MOP_dist}}
\end{figure}

The first engineer presented the designs shown in
Figure~\ref{fig:opt_MOP_dist_a}. These designs are biased towards
one end of the \ac{PF}. As a consequence, the \ac{DM} has plenty of
designs with high cost and small size, but few designs with low cost
and relatively bigger size. As the speaker size increases, the
designs get separated by an increasing incremental step. \emph{These
designs are not uniformly distributed across the \ac{PF}} which is
represented by grey line as shown in Figures~\ref{fig:opt_MOP_dist}.

The situation with the second engineer was different. His designs,
presented in Figure~\ref{fig:opt_MOP_dist_b}, was biased towards
both ends of the \ac{PF}. Most of the designs are for big cheap
speakers or small expensive ones. The \ac{DM} is left with few
options in-between. \emph{These designs also are not uniformly
distributed across the \ac{PF}.}

However the third design engineer provided a different alternative.
Despite the designs he made are uniformly distributed as shown in
Figure~\ref{fig:opt_MOP_dist_c}, they only cover a medium range of
the \ac{PF}. The \ac{DM} does not have the option to choose a highly
expensive and tiny speakers or an overly big and cheap ones.
\emph{These designs does not cover the entire \ac{PF}.}

The fourth engineer presented the best solution. His designs, as
shown in Figure~\ref{fig:opt_MOP_dist_d} does \emph{cover the entire
\ac{PF}}. Moreover, they are \emph{uniformly distributed across that
front}. The \ac{DM} in this case can choose from a sample of designs
which fairly represent all possible designs adhering to speaker
manufacturing standards, and with a varying trade-off between
speaker size and its cost.

%% file: EA/EA.tex
\chapter{Evolutionary Algorithms}
\label{ch:EAs}

\section{Introduction}
\label{ch:EAs_sec:intro}

During \emph{The Voyage of the Beagle} (1831--1836), Charles Darwin
(1809--1882) noticed that most species produce more offspring than
can grow to adulthood, however the population size of these species
remains roughly stable. It was the struggle for survival that
stabilizes the population size given the limited, however, stable
food resources. He noticed also that among sexually reproductive
species no two individuals are identical, although many of the
characteristics they bear are inherited. It were the variations
among different individuals that directly or indirectly
distinguished them among other species and their peers, and rendered
some of them more suitable to their environment that the others.
Those who are more fit to their environment are more likely to
survive and reproduce than those who are less fit to their
environment. Which means that the more fit individuals have greater
influence on the characteristics of the following generations than
the less fit individuals. This process results in offspring that
\emph{evolve} and \emph{adapt} to their environment over time, which
ultimately leads to \emph{new species} \cite{Darwin}. Darwin has
documented his observations and hypotheses in his renowned yet
controversial book ``On the Origin of Species'' \cite{Darwin} which
constituted the \emph{Darwinian Principles}.

Although Darwinian principles brought a lot of contention due to
apparent conflict with some religious teachings to the point that
some clergymen considered it blasphemy, these principles captured
the interest of many naturists and scientists. It wasn't until the
1950's when computational power emerged and allowed researchers and
scientists to build serious simulation models based on Darwinian
principles \cite{box:1957:eomiip}. During the 1960's and 1970's, the
field of \ac{EC} started to take ground by the work of Ingo
Rechenberg \cite{Rechenberg65}, John Bagley \cite{Bagley:1967:BAS},
John Holland \cite{holland75:adapt} and Kenneth De Jong
\cite{dejong:1975}. Those researchers developed their algorithms and
worked separately for almost 15 years until early 1990's when the
term \ac{CI} and its subfields were formalized by the IEEE Neural
Network Council and the IEEE World Congress on Computational
Intelligence in Orlando, Florida in the summer of 1994
\cite{Levitin:CompIntRel:2007}. Since then, these various algorithms
are perceived as different representations of a single idea.
Currently, \ac{EC} along with \ac{SI}, \ac{ANN}, Fuzzy systems among
other emerging intelligent agents and technologies are considered
subfields of \ac{CI} \cite{Wang:CompIntMan}.

Researchers have utilized different \ac{EC} models in analyzing,
designing and developing intelligent systems to solve and model
problems in various fields ranging from engineering
\cite{Levitin:CompIntRel:2007, baeck:1992:aea}, industrial
\cite{karr:1999:iaga}, medical \cite{Bosman:medical,
Chambers:PractGAAPPs}, economic \cite{Geisendorf:GAEconomics} among
many others \cite{baeck:1997:ecchcs}. For a comprehensive record on
\ac{EC} history the reader may refer to \cite{Fogel98,
baeck:1997:ecchcs, dejong:1997:hec}

\section{How and Why They Work?}
\label{ch:EAs_sec:HW?}

\begin{figure}
\centering
\includegraphics{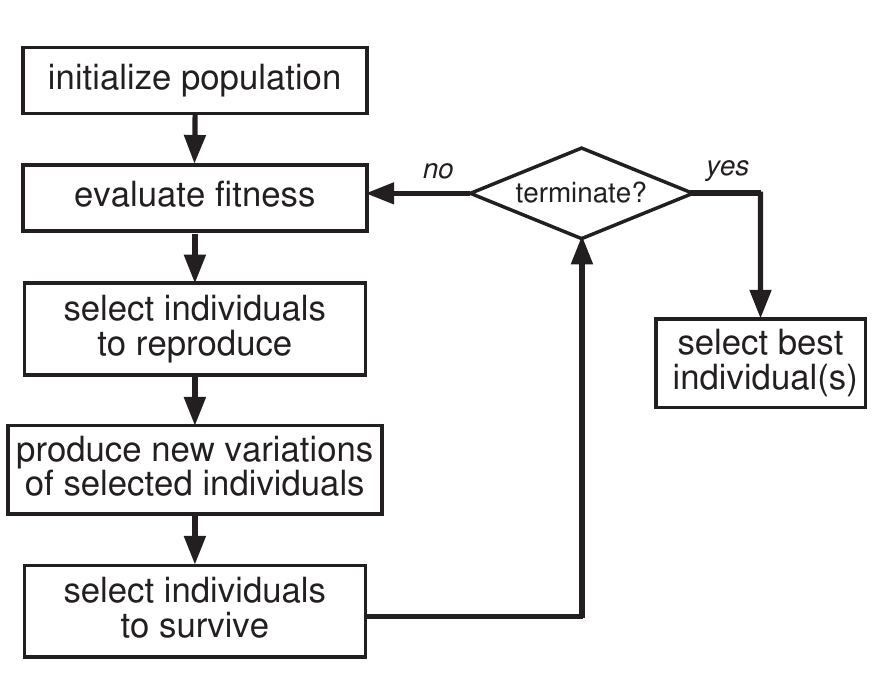}
\caption{Evolutionary Algorithms procedure \label{fig:EA_EAproc}}
\end{figure}

\acp{EA} are a subfield of \ac{EC}. They are biologically- and
nature-inspired algorithms. They work by \emph{evolving population
of potential solutions} to a problem, analogous to populations of
living organism. In real-world, individuals of a population vary in
their \emph{fitness} to their environment, some of them can defend
themselves against attacks, while others can't survive an attack and
perish. Some can provide food for themselves and their offspring all
the time, while others may not survive a short famine and starve to
death. In \acp{EA}, individuals of a population, likewise, are not
equally fit. Due to \emph{differences} in their
\emph{characteristics} some of them are more fit than the others,
and because the resources that keep them alive are limited, the more
fit is the individual the more likely it will survive and take part
in new \emph{generations}. Those who are fit enough to survive have
the chance to mate with other fit individuals and produce offspring
that carry a slightly modified version of their parents' \emph{good}
characteristics. But those fertile parents produce more offspring
than their environment resources can support. So the \emph{best} of
the \emph{good} individuals in the population will survive to the
next generation and start a new cycle of mating, reproducing and
fighting for survival. This cycle is shown in
Figure~\ref{fig:EA_EAproc}, where \emph{terminate} is the
termination condition; it could be a certain number of generations
or an error tolerance value or any other condition. As generations
pass by, the average fitness of individuals increases because they
\emph{adapt} to their environment. Which means that this population
of evolving solutions will explore the solution space looking for
the best value(s) and will progressively approach it(them).

The idea of using Darwinian selection to evolve a population of
potential solutions has a another great benefit in solving dynamic
problems. Because most real-life problems are dynamic, the
definition of fitness and the rules of the problem may change once
the problem has been formalized. So it will be useless to solve the
problem using the old rules because it means solving a problem that
does not exist any more. However, by using a population of evolving
solutions, individuals adapt to the new rules of their environment
over time.

\acp{EA} are a collection of algorithms that share the theme of
evolution. The mainstream instances of \acp{EA} comprise \acfp{GA}
\cite{holland75:adapt}, \acf{ES} \cite{Rechenberg65}, \acf{EP}
\cite{Fogel:1966:AIT}. In addition to those three methods, \acf{GP}
\cite{Koza:1989:HGA, koza:GPbook92}, \acf{LCS}
\cite{Goldberg:book89, Holland_etal:book86} and hybridizations of
\aclp{EA} with other techniques are classified under the umbrella of
\acp{EA} as well \cite{baeck:2000:intro}.

\section{Genetic Algorithms}
\label{ch:EAs_sec:GA}

\begin{figure}
\centering
\includegraphics{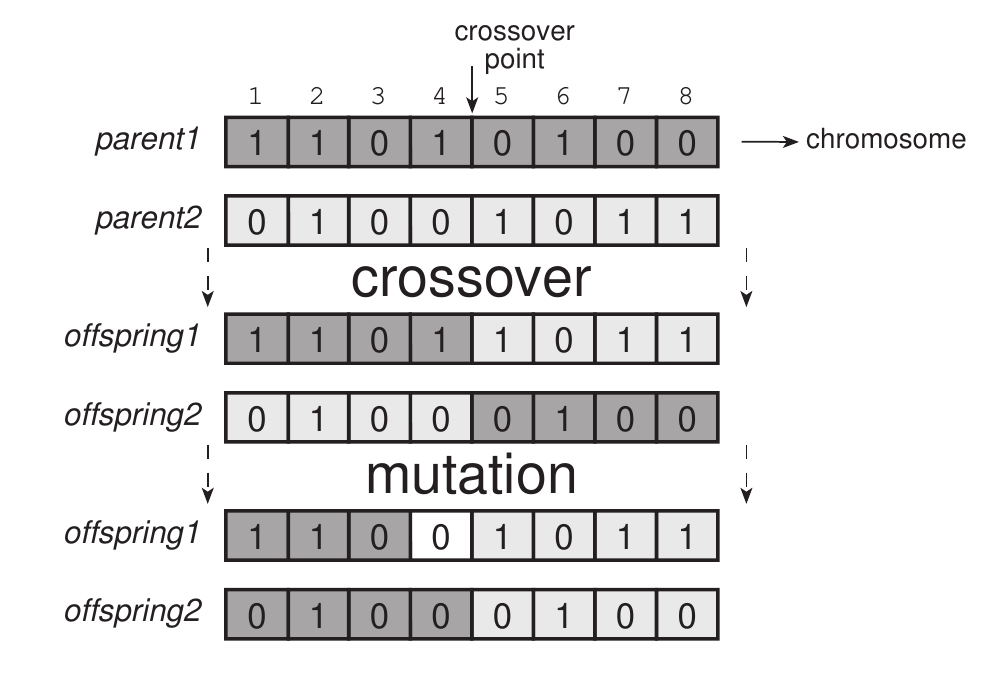}
\caption{Genetic variations \label{fig:GenVar}}
\end{figure}

\aclp{GA} are the most popular method among all \acp{EA}. It was
proposed by John Holland in 1962 \cite{Holland62} during the wake of
the quest for the general problem solver \cite{SimonNewell:1958,
newelshaw-1959, Gotlieb60}. \ac{GA} is a biologically-inspired
algorithm. It evolves a population of individuals encoded in
bitstrings (some variations of \ac{GA} use real numbers
representations). By analogy to biology, the encoded bitstring
structure of an individual in a population is known as a
\emph{chromosome}. Figure~\ref{fig:GenVar} shows an example of
chromosomes consisting of 8 bits each.

A \ac{GA} solver starts its procedure by creating a
\emph{population} of \emph{individuals} representing solutions to
the problem. The \emph{population size} (the number of individuals
in the population) is among the parameters of the algorithm itself.
A big population size helps exploring the solution space but is more
computationally expensive than a smaller population which may not
have the exploration power of the bigger population. After the
population is created, the \emph{fitness} of its individuals is
evaluated using the \emph{fitness function}. The fitness function is
a property of the problem being optimized and not the algorithm. It
reflects how good is a potential solution and how close it is to the
global optimum or the \ac{PF}. So, a good knowledge of the problem
is required to create a good fitness function that describes the
problem being optimized as accurately as possible. After evaluating
their fitness, some individuals are chosen to \emph{reproduce} and
are copied to the \emph{mating pool}. The number of individuals to
be chosen (the size of the mating pool) is a parameter of the
algorithm. Different mechanisms of selecting which individuals to
reproduce is explained in Subsection~\ref{ch:EA_sec:GA_subsec:sel},
but for now it is enough to say that most of these mechanisms favor
individuals with higher fitness values over those with lower fitness
values. The next step is to match those individuals for reproduction
which is done by random in most cases.

Two \emph{variation operators} are applied on the matched
individuals (\emph{parents}) to produce their \emph{offspring}. The
fist variation operator is the \emph{crossover} between the parents'
\emph{chromosomes}. As shown in Figure~\ref{fig:GenVar},
\emph{parent1} and \emph{parent2} are two matched individuals. The
chromosomes of those two parents are \emph{cut} at the
\emph{crossover point} (between bit~\#4 and bit~\#5), and the
resulting \emph{half chromosomes} are swapped to create two
offspring, \emph{offspring1} and \emph{offspring2}. It is to be
noted however that the crossover operator may not be applied on all
parents in the mating pool. The \emph{crossover ratio} defines the
percentage of parents in the mating pool which will be affected by
the crossover operator. This value is algorithm dependent but it
varies around 0.9 for most \ac{GA} implementations
\cite{spears:2000:MutaRecom}. After the crossover is done, the
offspring chromosomes are \emph{mutated}. Mutation of chromosomes in
binary representation is done by flipping one or more bits in a
chromosome. As shown in Figure~\ref{fig:GenVar}, \emph{offspring1}
was mutated by flipping its $4^{\text{th}}$ bit from $1\to 0$. As
the case with the crossover operator, the mutation operator may not
affect all individuals. The ratio of mutated \emph{bits} to the
total number of bits is known as the \emph{mutation ratio} and is
typically below one percent \cite{spears:2000:MutaRecom}. As the
mutation ratio increases, the algorithm becomes more a random search
algorithm.

The offspring join the population after evaluating their fitness to
fight for survival. This stage is crucial for all individuals; based
on their fitness, some of them will survive to the next generation,
while others will perish. Different mechanisms can be used to select
surviving individuals. Many of which are probabilistic techniques
that favor more fit individuals.

The surviving individuals make-up a new generation and restart the
cycle of fitness evaluation, mating selection, reproduction and
survival selection. This cycle is repeated until a stopping
condition is met, which could be the number of cycles, the solution
error of the best individual or any other condition or mix of
conditions set by the algorithm designer.

\acp{GA} are flexible problem solvers; they are less dependent on
the problem being solved than traditional techniques. Moreover they
provide high degree of flexibility for their designer. A \ac{GA}
designer can choose a suitable \emph{representation} scheme, a
\emph{mating selection} technique and the \emph{variation operators}
of his choice.

\subsection{Representations}
\label{ch:EA_sec:GA_subsec:rep}

How can a \ac{GA} be less dependent on the problem being optimized
than a traditional technique? The answer is because the problem is
being transformed to the algorithm domain before the \ac{GA} starts
solving the problem. Some people even argue that a \ac{GA} is
problem independent because this transformation is done before the
algorithm is invoked. The transformation to the problem domain is
done mainly by providing an encoding mechanism for possible
solutions to the problem, aka \emph{'representation'}, and by
creating a \emph{fitness function}.

The choice of which representation to use should be done within the
context of the problem. A good representation for a scheduling
problem may not be suitable for the Rastrigin problem given earlier
in Section~\ref{ch:opt_sec:SOP} and vice versa.

\subsubsection{Binary Representations}
\begin{figure}
\centering
\includegraphics{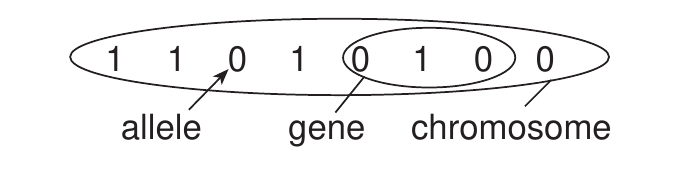}
\caption{Chromosome structure \label{fig:EA_chrom}}
\end{figure}

The oldest and most used scheme of representation is the binary
representation. As shown in Figure~\ref{fig:EA_chrom}, a chromosome
consists of a string of bits, each one of those bits is known as an
\emph{allele}. A \emph{gene} is a combination of one or more alleles
which determine a characteristic of the individual. For example, if
the chromosome shown in Figure~\ref{fig:EA_chrom} is for an
imaginary creature, then the three circled alleles $[0~1~0]$ may
represent its eye color gene, and by varying the values of those
three alleles its eye color will change.

To use binary representation for a combinatorial problem, each
possible solution to the problem can be represented by a unique
sequence of digits. For example, the solutions of the \ac{SAT}
problem given in Subsection~\ref{ch:opt_sec:dif_subsec:srch} can be
directly transformed to bitstrings. So the chromosome $[0111001101]$
will be one of $2^{10}$ different solutions to the 10-variables
\ac{SAT} problem.

However, using binary representation to encode real valued solutions
of the Rastrigin problem will work differently. First, the search
space or the solution space is continuous, which means there are
infinite number of possible solutions, so it must be sampled. A very
small sampling step leads to huge number of possible solutions which
fairly represents the solution space but it will be computationally
expensive to search this huge number of solutions. While a big
sampling step make it less expensive but on the expense of a bad
representation of the solution space. If the precision used to
sample the Rastrigin problem is four digits after the decimal point
it means that there are $6-(-6)\times 10000=120000$ possible
solutions, which requires 17 digits to represent them. So, $-6\to
00000000000000000$ and $6\to 11111111111111111$. A transformation of
the binary string $<b_0b_1\dotsm b_{17}>$ back to the decimal form
is done by transforming the binary number to a decimal one $x'$ then
using the following rule:
\begin{equation}
\label{eq:EA_rep_bin_trans}%
x=-6+\frac{12x'}{2^{17}-1}
\end{equation}
So the solution $01001011101101111$ is transformed to the decimal
number $x'=38767$ then by using~\eqref{eq:EA_rep_bin_trans}
\begin{equation}
x=-6+\frac{12\times 38767}{2^{17}-1}=-2.4507
\end{equation}

\subsubsection{Real-valued Representations}
\label{ch:EA_realrep} It could be more suitable for problems with
real valued solutions to be represented using a real-valued
representation. In this representation, each individual is a real
valued number that expresses the value of the solution. Using this
representation, there is no need to define a solution precision,
such as the one defined for the binary representation, because the
search or variation operators will be able to transform this
solution to any of the infinite number of possible solutions in the
solution space (if the floating point precision of the computer used
to solve the problem is unlimited). However this type of
representation requires another mechanism of crossover and mutation
as explained in Subsection~\ref{ch:EA_sec:GA_subsec:VarOp}.

\subsubsection{Integer Representations}
For permutation problems, it might be more intuitive to represent
solutions using integer values. This is illustrated using the
following \ac{TSP} example.

The \ac{TSP} is a combinatorial optimization problem that has been
extensively studied in the last 150 years \cite{TSPhistory} due to
its numerous applications such as in transportation, networking and
printed circuit manufacturing. The problem is defined as
\cite{TSPhistory}: given $n$ cities and their intermediate
distances, find a shortest route traversing each city exactly once.

\begin{figure}
\hfill
\begin{minipage}[b]{0.45\textwidth}
\centering
\includegraphics{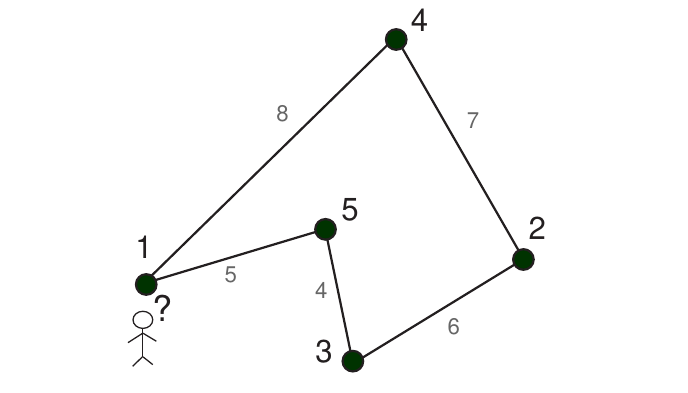}
\caption{The traveling-salesman problem \label{fig:EA_TSP}}
\end{minipage}
\hfill
\begin{minipage}[b]{0.45\textwidth}
\centering
\includegraphics{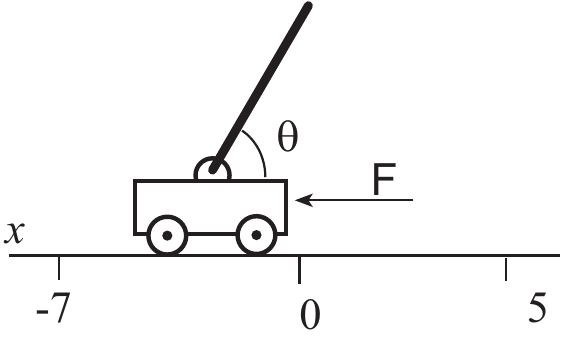}
\caption{The inverted pendulum problem \label{fig:EA_InvPen}}
\end{minipage}
\hfill
\end{figure}

As shown in Figure~\ref{fig:EA_TSP}, a possible route the traveling
salesman can take is $1\to 5\to 3\to 2\to 4\to 1$. This route can be
directly mapped to and represented by the string of real numbers
$[1~5~3~2~4~1]$, where the integers represent the cities to be
visited by the order they are stored in the string.

\subsubsection{Problem-Specific Representations}
Using only one of the previously mentioned representations to
represent all the variables of a problem may not be the best option
for some problems, which is the case in the following example; The
inverted pendulum problem shown in Figure~\ref{fig:EA_InvPen} is a
classic problem in control engineering. The objective is to
construct a function that maps the system states to a horizontal
force applied on the cart carrying the inverted pendulum to
stabilize it vertically. The states of this problem are the position
of the cart $x$, its speed $\dot{x}$, the angular position of the
pendulum $\theta$ and its angular velocity $\dot{\theta}$. For such
a problem the parameters $x, \dot{x}, \theta, \dot{\theta}$ will be
the input to the function and the output will be the applied
horizontal force. This problem can be represented using a parse tree
of symbolic expressions \footnote{although parse tree
representations are mainly used for \ac{GP}, it is was originally
proposed as a \ac{GA} representation \cite{Koza:1989:HGA}}
\cite{Koza:1989:HGA, mitchell:1996:iga, Michal00}. The parse tree
consists of operators, such as
($+,~-,~\cos,~\sin,~\times,~\exp,\dotsm$) and arguments, such as
($x,~\theta,~t,~r,\dotsm$). A possible tree may look like:
\begin{equation}
(\times(\cos(+(x)(3.14)(\theta)))(t))
\end{equation}
which is equivalent to:
\begin{equation}
t\cos(x+\theta+3.14)
\end{equation}

\subsubsection{More on Representations}
The three schemes of representations mentioned above are by no means
exhaustive. An algorithm designer may come up with his own
representation which may better suit his problem. However, the
representation scheme, as mentioned before, should never be
considered independent of the problem; If the variation operators
used do not suit the representation, they may produce offspring who
are totally different from their parents, which acts against the
learning process because the algorithm samples new trials from the
state space of possible solutions without regard for previous
samples. This procedure may perform worse than random search
\cite{Michal00}


Some representations incorporate the constraints of the problem and
provide feasible solutions for the problem all the way. Although it
may be desirable to evolve feasible solutions all the time and not
to worry about the constraints or define them explicitly in the
problem, it is useful to allow the algorithm to evolve few
infeasible solutions to explore the solution space for possible
scattered feasible regions that may contain better values.

\subsection{Mating Selection}
\label{ch:EA_sec:GA_subsec:sel}

After a decision about how solutions will be represented is made,
the next step is to decide about the mating selection procedure.
Mating selection focusses on the exploration of promising regions in
the search space \cite{ahn06}. It tends to select good individuals
to mate---hoping that their offspring will be as good or better than
them because they will inherit many of their parents'
characteristics. However, if the selection procedure strictly
selects the very few top of the population, the population will lose
its diversity because after few generations the population will only
contain a slightly different copies of few individuals who were the
best in their generations, and if this procedure continues, the
population will be made-up of almost identical individuals after few
more generations. The main idea behind a \ac{GA} is to evolve a
population of competing individuals, so if the individuals became
identical, there will be no competition and henceforth no evolution.
So, keeping a diverse population indeed helps the algorithm explore
the search space \cite{mitchell:1996:iga}. The degree to which the
algorithm favors and selects the best individuals in the population
for mating is measured by the \emph{selection pressure}. The
selection pressure is defined as ``\emph{the ratio of the
probability of selecting the best individual in the population to
that of an average individual}'' \cite{ahn06}. As the selection
pressure increases, the algorithms tends to choose the very best of
the population to mate and produce the next generation leading to a
population that converges to a local optima, this situation is known
as \emph{premature convergence}. On the other hand, if the selection
pressure decreases, the algorithm will converge slowly and wander in
the search space. It should be clear that the selection pressure is
not a parameter that the algorithm designer explicitly set its
value, instead, it is influenced by different aspects of the
algorithm, especially the selection mechanism used. It is to be
noted that the following selection mechanisms define the preference
of selecting an individual in the population, however the number of
offspring this individual produces for each mating process is
another issue.

\subsubsection{Fitness Proportionate Selection}
\ac{FPS} is one of the earliest selection mechanisms proposed by
Holland \cite{holland75:adapt} (Sometimes known as roulette wheel
selection). It selects individuals based on their absolute fitness.
That is, if the fitness of an individual $j$ in the population is
$f_j$, the probability of selecting this individual is
$z_j=\frac{f_j}{\sum_{i=1}^Nf_i}$, where $N$ is the population size.
This selection mechanisms allows the most fit individuals in
population to multiply very quickly early in the run and take over
the population, and after few more generations the diversity almost
vanishes and the selection pressure becomes very low leading to
stagnant population. In other words, leading to premature
convergence. Another handicap of this mechanism is that it behaves
differently on transposed versions of the same fitness function
\cite{eiben2003}. For example, given a population of two individuals
$a$ and $b$ with fitness values 1 and 2, respectively. The
probability of choosing the first individual is $\frac{1}{3}=0.33$
while that of the second is $\frac{2}{3}=0.67$, this is a $1:2$
ratio. However if a constant value of 10 is added to the fitness
values of the population the probabilities will become
$\frac{11}{23}=0.48$ and $\frac{12}{23}=0.52$ which is almost a
$1:1$ ratio.

A possible remedy to the weak selection pressure and the
inconsistent behavior of the algorithm along the run can be achieved
by subtracting the fitness of the worst individual in the population
of a \emph{window} containing the last $n$ generations. This
approach can be considered as a dynamic scaling of the population
fitness.

Another possible solutions is achieved by using Goldberg's sigma
scaling method \cite{Goldberg:book89}, which scales the fitness of
individuals using the mean $\bar{f}$ and standard deviation
$\sigma_f$ of fitness in the population.

\begin{equation}
f'(x)=\min(f(x)-(\bar{f}-c\cdot\sigma_f), 0)
\end{equation}

\subsubsection{Rank Selection}
Another way to overcome the deficiencies of the \ac{FPS} is to use
the rank selection. As its name implies, this procedure orders all
individuals in the population according to their fitness, and then
selects individuals based on their rank rather than their absolute
fitness such as in \ac{FPS}. This methods maintains a constant
selection pressure because no matter how big is the gap between the
most and least fit individuals in a population, the probability of
selecting each one of them will remain the same as long as the
population size remains the fixed.

After the individuals are ordered in the population, they are
assigned another fitness value inversely related to their rank. The
two most used fitness assignment functions result in \emph{linear
ranking} and \emph{exponential ranking}.

For linear ranking, the best individual is assigned a fitness of
$s\in[1,2]$, while the worst one is assigned a value of $2-s$. The
fitness values of the intermediate individuals are determined by
interpolation. This can be achieved using the following function:
\begin{equation}
f(i)=s - \frac{2(i-1)(s-1)}{N-1}
\end{equation}
Where $i$ is the individual rank $i\in [1,N]$ and $N$ is the number
of individuals. For linear ranking, the selection pressure is
proportional to $s-1$.

For nonlinear ranking, the best individual is assigned a fitness
value of 1, the second is assigned $s$, typically around 0.99
\cite{Hancock95}, the third $s^2$, and so on to lowest ranked
individual. The selection pressure for nonlinear ranking is
proportional to $1-s$.

\subsubsection{Tournament Selection}
Unlike previously mentioned selection procedures, the tournament
selection does not require a knowledge about the fitness of the
entire population, which can be time consuming in some application
with huge populations which requires fast execution. It is suitable
for for some situations with no universal fitness definition such as
in comparing two game playing strategies; it might be hard to set a
fitness function that evaluates the absolute fitness of each of
those two strategies, but it is possible to simulate a game played
by those two strategies as opponents, and the winner is considered
the fittest.

Tournament selection picks $k$ individuals by random from the
population and selects the best one of them for mating. This does
not require a full ordering of the sample nor an absolute knowledge
of their fitness. The selection pressure of this mechanism can be
varied by changing the sample size $k$, it increases by increasing
$k$ and reaches the maximum at $k=N$. This selection mechanism can
be used with or without replacement. Moreover a non-deterministic
selection can be used, which means that the probability of selecting
the best individual in the sample is less than one to give a chance
for the worst individual in the population to be selected,
otherwise, it will never be selected.

This selection mechanism is the most widely used because it is
simple, does not require knowledge of the global fitness, adds low
computation overhead and its selection pressure can be easily
changed by varying the sample size $k$. A common sample size is
$k=2$ \cite{baeck:1996:eatp}.

For more mating selection techniques the reader may refer to
\cite{Hancock95, baeck:2000:EC1}.

\subsection{Variation Operators}
\label{ch:EA_sec:GA_subsec:VarOp}

In real-world, although parents and their offspring have common
characteristics, they are never identical; Like father like son, but
the son is not a clone of the father. This \emph{variation} among
individual was noted by Charles Darwin in his controversial book
\cite{Darwin} where he emphasized that this variation is a major
force that drives the evolution process. To mimic this process in
\acp{GA}, researchers have developed \emph{variation operators} that
help the algorithm search the solution space, henceforth, they are
sometimes known as \emph{search operators}. These variation
operators have two goals: The first one is to produce offspring that
\emph{resemble} their parents, while the second one is to slightly
perturb their characteristics. The oldest and most widely used
variation operators are the \emph{crossover} and the \emph{mutation}
operators \cite{holland75:adapt}. They were proposed by John Holland
to operate on binary \acp{GA}, however many other variation
operators were proposed to operate on other forms of \ac{GA}
representations \cite{deb95simulated}. It is to be noted that all
syntactic manipulations by variation operators must yield
semantically valid results \cite{booker:EC1_recomb}.

\subsubsection{Crossover}

The fact that the offspring very often look much like their parents
intrigued Gregor Mendel (1822--1884) and lead him to do his famous
experiments on pea plants. By analyzing the outcomes of his
experiments on some 28000 samples he reached a conclusion about the
rules of inheritance and published these findings in a paper
\cite{Mendel1865} which is considered to be the basis of modern day
genetics rules. Researchers in \acp{GA} were inspired by these rules
and used a modified version of them in their algorithms
\cite{holland75:adapt}, and from there came the \ac{GA} crossover
operator.

A recombination operator is known to be an \emph{exploitation
operator} because it exploits the accumulated knowledge in the
current solution vectors by using parts of them as building blocks
of their offspring. It is widely believed among \ac{AI} community
that exploitation should be emphasized at later stages of the search
to prevent premature convergence \cite{eiben98evolutionary,
naudts99motivated}.

Different versions of the crossover operator are applied to binary,
real-valued, and parse tree representations but they essentially do
the same job; They simply create the genes of an offspring by
copying a combination of its parents genes.

\begin{figure}
\centering
\includegraphics{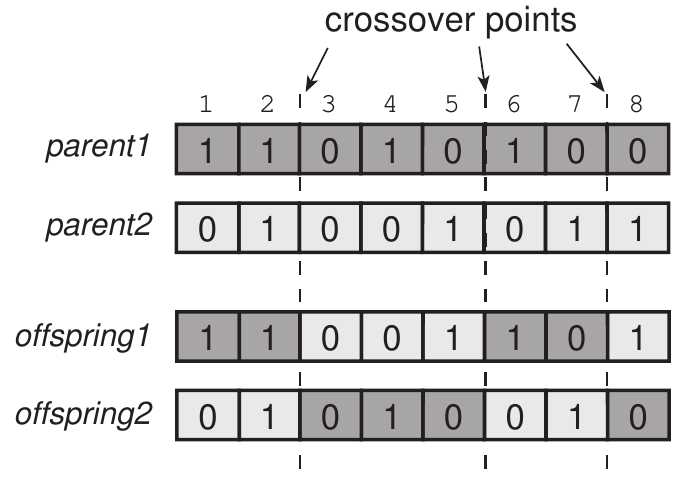}
\caption{Crossover in binary GA \label{fig:EA_xover}}
\end{figure}

\emph{n-point crossover} \cite{dejong:1975} is used mainly for
binary representations. The operator randomly picks $n$ points along
a copy of the two parents chromosomes, divide each one of them into
$n+1$ parts, and then swap some of these parts to create the
offspring chromosomes. Figure~\ref{fig:EA_xover} shows an example of
a \emph{3-point} crossover operation. First, the crossover points
are randomly assigned the positions shown in figure, dividing each
parent into 4 parts. Then, \emph{offspring1} is created by copying
the first and third parts of \emph{parent1}, and the second and
fourth parts of \emph{parent2}. While \emph{offspring2} is created
by copying the first and third parts of \emph{parent2}, and the
second and fourth parts of \emph{parent1}. The probability of
independently crossing over each individual in the mating pool is
known as the \emph{crossover ratio} $p_c$.

\emph{Uniform crossover} is another alternative to use with binary
\acp{GA}. The number of crossover points in this operator is not
fixed, instead, it creates each allele of the offspring by copying
one of the two corresponding alleles in its parents with a certain
probability. This procedure can be explained using the following
Matlab code:
\begin{lstlisting}[emph=z, emphstyle=\emph, title = {Uniform crossover procedure (Matlab code)}]
z = 0.5;        % z is the crossover probability
o = p1;         % o is an offspring, p1 is the first parent
for i = 1 : k   % k = chromosome length
  x = rand;     % random number in the range (0, 1)
  if (x>z)
    o(i)=p2(i); % copy the i allele of parent2 to offspring
  end
end
\end{lstlisting}
By increasing the value of $z$, the offspring alleles will be more
like those of p1. The previous code is repeated for each offspring.

A real-valued representation can be transformed into binary
representation to apply a binary crossover operator as shown
previously, and then get transformed back to real-valued
representation, but it is not recommended to follow this procedure
and add such a computational cost and sacrifice precision due to
sampling and rounding-off errors encountered in
decimal--binary---binary--decimal conversion. Instead, it is
recommended to apply some of the recombination operators specially
made for real-valued representations.

\emph{The blending methods} create the variables of the offspring by
weighting the corresponding variables in their parents chromosomes.
For example, to create the variables in the offspring chromosome,
the following rule can be used:
\begin{align}
o_{1,i}=\beta p_{1,i} + (1- \beta)p_{2,i}\\%
o_{2,i}=(1 - \beta) p_{1, i} + \beta p_{2,i}
\end{align}
Where $o_{1,i}$ is the $i^{\text{th}}$ variable of the \emph{first}
offspring, and $\beta$ is the weighting factor in the range [0,1].
The higher the value of $\beta$, the more the offspring will look
like the first parent ($p_1$), and the lower $\beta$ gets, the more
the offspring will resemble the second parent ($p_2$).

\emph{The \ac{SBX}} operator is a recombination alternative to
consider for real-valued representations. This operator was proposed
by Deb \cite{deb95simulated}, and he claims that it provides
self-adaptive search mechanism \cite{Deb99SBX}. This operator is
explained using the following rules:
\begin{align}
o_{1,i} = 0.5[(1 + {\beta}_i)p_{1,i} + (1 - {\beta}_i)p_{2,i}] \\%
o_{2,i} = 0.5[(1 - {\beta}_i)p_{1,i} + (1 + {\beta}_i)p_{2,i}]%
\end{align}
and ${\beta}$ is evaluated using the following rule:
\begin{equation}
\beta =  \left\{\begin{alignedat}{2}%
&(2u)^{\frac{1}{\eta + 1}},                            &\quad&\text{if} \quad u \le 0.5\\
&\left( \frac{1}{2(1-u)} \right)^{\frac{1}{\eta + 1}},
&&\text{otherwise}
\end{alignedat}\right.
\end{equation}
where $u$ is a uniformly distributed random number in the range $(0,
1)$, and $\eta$ is a distribution index; a low value of $\eta$ (less
than 2) gives high probability of producing offspring different from
their parents, while a high $\eta$ value (greater than 5) means that
the offspring will be very close to their parents in the solution
space.

For parse tree representations, a recombination operator do swap
subtrees of the solution vectors. For example, given the following
solution vector
\begin{equation}
(+(\cos(\exp (3)(7)))(\times(4)(\log(3))))
\end{equation}
by swapping the subtrees $\exp (3)(7)$ and $\times (4)(\log(3))$, it
becomes
\begin{equation}
(+(\cos(\times(4)(\log(3))))(\exp (3)(7)))
\end{equation}

However care must be taken to prevent trees from growing rapidly and
reaching an extremely long length, because it may halt the machines
executing the algorithm. This is known as \emph{bloating}, and it is
a major concern in \ac{GP} implementations.

\subsubsection{Mutation}
Unlike recombination operators, mutation operators do not make use
of the knowledge of the search space acquired through generations,
they do perturb the population by providing random genetic material
provided they result in semantically valid results.

\emph{For binary \acp{GA}}, the mutation operator first determines
the positions of the alleles that will undergo mutation. However,
the choice of these alleles is made at random with equal probability
for each one of them to be mutated (uniform distribution), and their
number is determined using the \emph{mutation ratio} $p_m$, which is
the probability of independently inverting one allele. Second, the
operator flips the selected alleles to produce the mutated
offspring. For example, given a pool with two solution vectors
$a_1=1001101011$, $a_2=1011011110$, and $p_m=0.1$. A mutation
operator being applied on them will mutate $0.1\times2\times10=2$
alleles, and may turn the solution vectors into
$a_3=10\boldsymbol{1}11010\boldsymbol{0}1$, $a_4=1011011110$,
respectively. Note that the two alleles to be mutated happened to be
at the first solution vector (shown in boldface), while the second
one remained intact.

Mutating \emph{real-valued \ac{GA}} pose some challenges. If the
mutation operator does not operator on specific values of the
parents \cite{MontanaD89}, it will allow the offspring escape local
optima and will help the algorithm explore new regions of the search
space. But it will break the link between the parents and their
offspring instead of causing causing small perturbation. On the
other hand, if the mutation operator do operate on specific values
of the parents to produce their offspring, it may not be very
helpful in escaping local optima, but will keep the link between the
parents and their offspring. The later approach will be further
explained here.

For a $n$-dimension solution vector $\v{x}\in \mathbb{R}^n$ the
mutation operator will be in the form \cite{back:EC1_mutation}
\begin{equation}
\v{x}' = m(\v{x})
\end{equation}
where $\v{x}$ is the parent solution vector, $m$ is the mutation
operator, and $\v{x}'$ is the mutated offspring solution vector. The
mutation operator $m$ may simply add a real random variable $\v{M}$
to the parent vector
\begin{equation}
x_i' = x_i + M_i
\end{equation}
where $x_i$ is the $i^{\text{th}}$ variable of the solution vector
$\v{x}$. It is recommended to create $\v{M}$ with a mean value of 0
to prevent a bias towards some parts of the search space and keep
the offspring uniformly distributed around their parents. If $\v{M}$
has a uniform distribution in the range $[-a,a]^n$, it will be
equally probably that $\v{x}'$ will take any value in the hyper-box
$[\v{x}-a,\v{x}+a]^n$ if mutated. Alternatively, $\v{M}$ can have a
normal (or Gaussian) distribution which can be represented by the
following equation:
\begin{equation}
M_i = \frac{1}{\sigma\sqrt{2\pi}} \, \exp \left( -\frac{(x_i-
\mu)^2}{2\sigma^2} \right)
\end{equation}
where $\sigma$ is the standard deviation, and $\mu$ is the mean. If
$\mu$ is set to 0, the value of $\sigma$ will determine the
probability of different mutation strengths; As $\sigma$ increases,
it becomes more probable that the offspring will lie away from its
parent, while decreasing $\sigma$ value increases the probability of
producing offspring that looks similar to its parent.

The \emph{Polynomial Mutation} is another mutation operator for
real-valued \acp{GA} \cite{deb95simulated, Raghuwanshi04:PolyMut}.
The offspring is mutated using the following rule
\begin{equation}
x_i'=x_i + (x_{iu}-x_{il})\delta_i
\end{equation}
where $x_{iu}$ and $x_{il}$ are the \emph{upper} and \emph{lower}
bounds of the variable $x_i$, respectively,  and $\delta_i$ is
calculated from a polynomial distribution by using
\begin{equation}
\delta_i = \left\{
\begin{alignedat}{2}
&(2r_i)^{\frac{1}{\eta_m+1}}-1       \qquad &\text{if}\; r_i<0.5\\%
& 1-[2(1-r_i)]^{\frac{1}{\eta_m+1}}  \qquad &\text{otherwise}%
\end{alignedat}
\right.
\end{equation}
where $r_i$ is a uniformly distributed random number in the range
$(0,1)$, and $\eta_m$ is a mutation distribution index.

\begin{figure}
\centering%
\subfloat[grow]{\includegraphics[height=5cm]{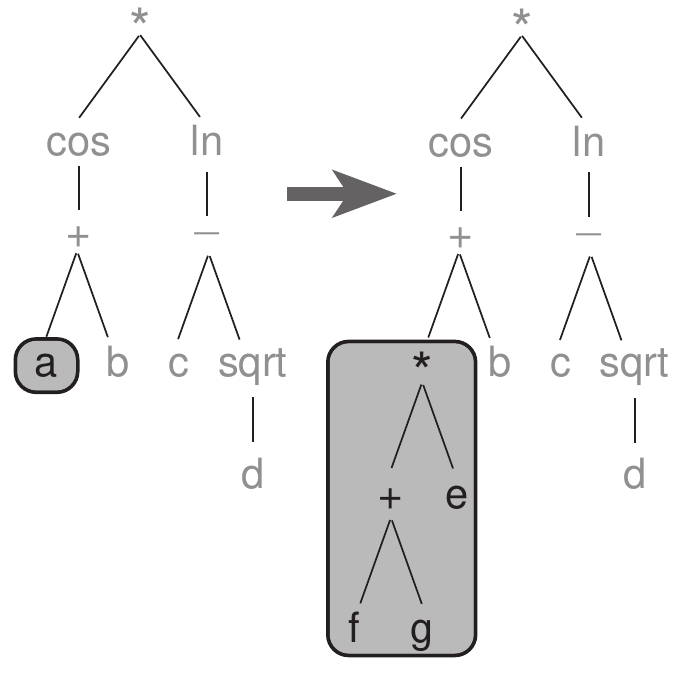}
\label{fig:EA_parse_grow}}
\qquad \qquad %
\subfloat[shrink]{\includegraphics[height=5cm]{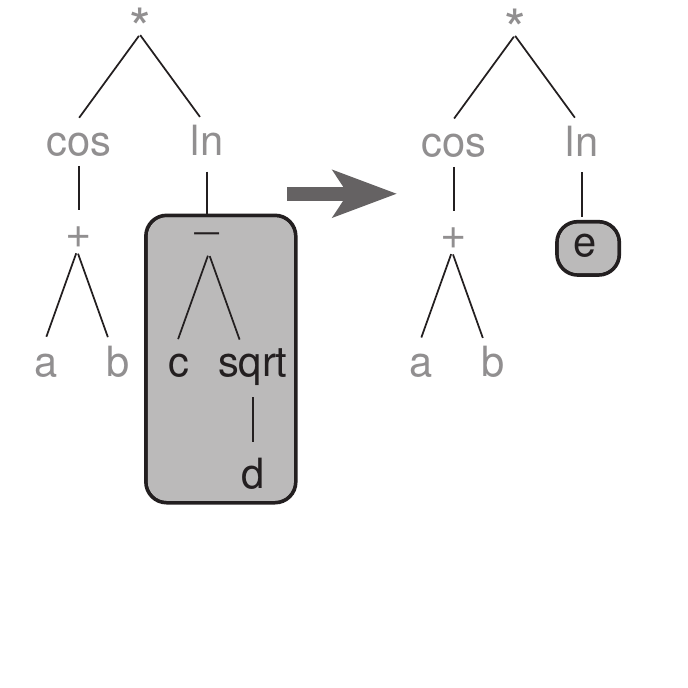} \label{fig:EA_parse_shrink}} \\
\subfloat[switch]{\includegraphics[height=5cm]{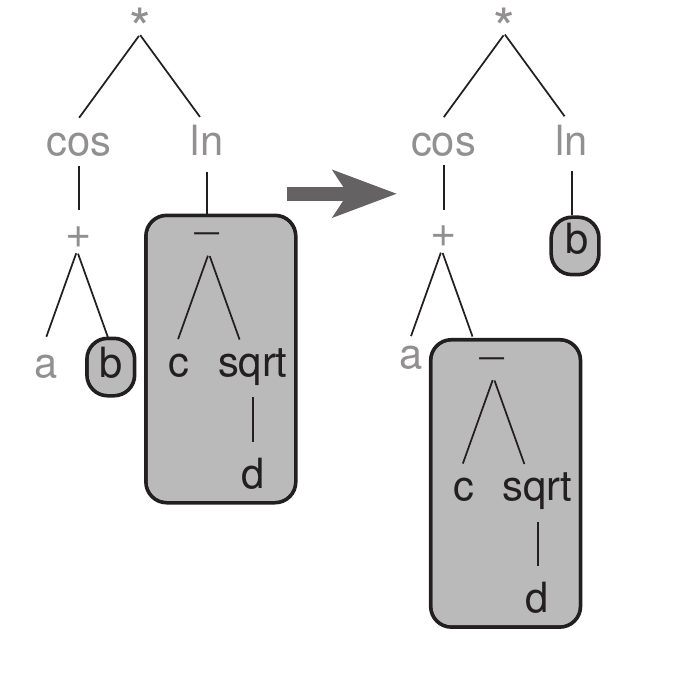}
\label{fig:EA_parse_switch}} \qquad \qquad%
\subfloat[cycle]{\includegraphics[height=5cm]{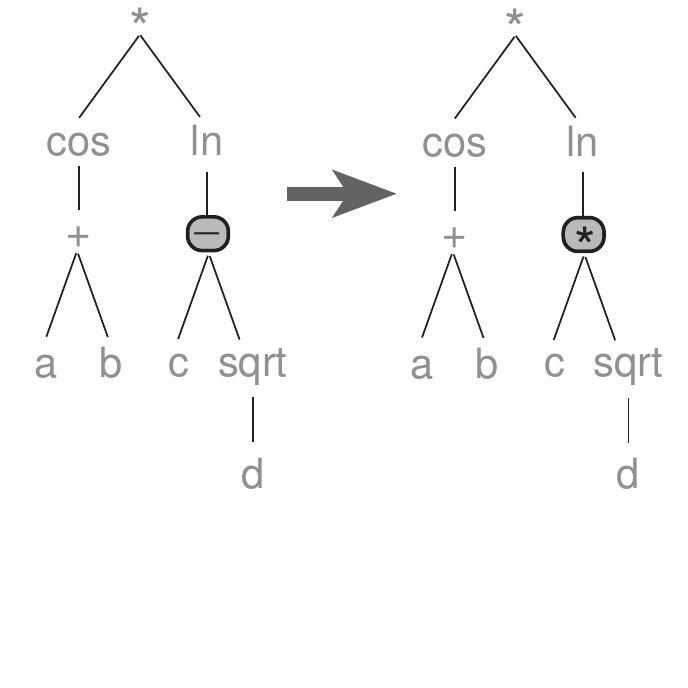} \label{fig:EA_parse_cycle}}%
\caption{Mutation methods for a parse
tree\label{fig:EA_parse_mutation}}
\end{figure}

The final mutation operators to be presented here is the \emph{parse
tree} mutation operators. Among many operators, Angeline defines
four mutation operators \cite{back:EC1_mutation}. The \emph{grow}
operator randomly selects a leaf from the tree and replace it with a
randomly generated subtree (Figure~\ref{fig:EA_parse_grow}). The
\emph{shrink} operator randomly selects an internal node from the
tree and replaces the subtree below it with a randomly generated
leaf (Figure~\ref{fig:EA_parse_shrink}). The \emph{switch} operator
randomly selects an internal node from the tree and rearrange its
subtrees (Figure~\ref{fig:EA_parse_switch}). While the \emph{cycle}
operator randomly selects a node from the tree and replaces it with
another node, provided that the resulting tree is a valid one
(Figure~\ref{fig:EA_parse_cycle}).

\subsection{Schema Theorem}
In his early work on \acp{GA}, John Holland presented the schema as
a building block for individuals in a population. While explicitly
evaluating the fitness of the individuals, a \ac{GA} implicitly
evaluates the fitness of many building blocks without any added
computation overhead (implicit parallelism). As the search for the
optimal solution progresses, the algorithm focusses on promising
regions of the search space defined by schemata having fitness above
the average of the population in their generations
(\emph{above-average schemata}); The more fit are the individuals a
schema produces, the more samples will be produced from this schema
\cite{mitchell:1996:iga}. The following discussion on schemata
assume a binary \ac{GA} representations.

A schema is a template for solution vectors. It defines \emph{fixed}
values for some positions, and assumes \emph{don't care} values for
the other positions. For example, given the following solution
vector $S_1$ of length $m=10$
\begin{equation}
\label{eq:EA_sch_order}%
S_1=(*~1~1~1~0~*~*~0~1~*)
\end{equation}
there are six fixed positions (taking a value of `1' or `0'), and
four \emph{don't care} positions (marked by `*'). The number of the
fixed positions in a schema is known as the \emph{order} or the
schema. So the order of the $S_1$ schema is $o(S_1)=6$. Another
property of the schema is its \emph{defining length}, which is the
distance between the first and last fixed positions in a schema. The
defining length of the $S_1$ schema is $\delta(S_1)=9-2=7$

Any schema may produce $2^d$ distinct individual, where $d$ is the
number of \emph{don't care} positions in this schema. On the other
hand, any solution vector can be matched by $2^n$ different
schemata. It is clear that the schema $(*~*~*~*~*~*~*~*~*~*)$ is the
most general schema; It matches $2^{10}$ solution vector (which
includes the entire population). While $(1~1~1~1~1~1~1~1~1~1)$ is
among the most specific schemata; It matches a single solution
vector, namely $(1111111111)$.

Another property of a schema is its \emph{fitness} at time $t$,
$f(S, t)$. The fitness of a schema is defined as the average fitness
of all individuals in the population matched by this schema
\cite{Micwcz:GA+DS+EP:96}.

Given a population of size $N$, with solution vectors $\v{x}_i$ of
length $m$ and a fitness function $f$. The schema $S$ which matches
$w$ individuals $(\v{x}_1,\dotsm ,\v{x}_w)$ in the population at
time $t$ will have an average fitness of
\begin{equation}
f(S,t)=\frac{1}{N} \, \sum_{i=1}^w f(\v{x}_i)
\end{equation}

If the number of individuals matched by schema $S$ is $\xi(S,t)$ and
their average fitness is $\frac{f(S,t)}{F(t)}$, where $F(t)=\sum_i^N
f(\v{x}_i(t))$, then it is expected that for the next generation the
number of individuals matched by $S$ will be
\begin{equation}
\xi(S, t+1)=\frac{\xi(S,t)\times N \times f(S,t)}{F(t)}
\end{equation}
if the average fitness of the population is
$\overline{F}(t)=\frac{F(t)}{N}$, the the previous equation can be
rewritten as
\begin{equation}
\label{eq:EA_schm_sel}%
\xi(S, t+1)=\frac{\xi(S,t)f(S,t)}{\overline{F}(t)}
\end{equation}

From the last equation, it is clear that if the schema is
above-average for the current generation ($f(S,t)>\overline{F}(t)$),
then the number of individuals matched by this schema will increase
in the next generation. But if the schema is below-average, the
number of individuals this schema matches will decrease in the next
generation.

If $\xi(S,0)$ is known, then $\xi(S,t)$ can be directly evaluated
from \eqref{eq:EA_schm_sel}
\begin{equation}
\label{eq:EA_schm_exp}%
\xi(S,t)=\xi(S,0)\,\left(\frac{f(S,t)}{\overline{F}(t)}\right)^t
\end{equation}
assuming that $S$ will maintain a fixed above-average fitness
through generations.

The \emph{reproductive schema growth}
equation~\eqref{eq:EA_schm_exp} only considers selection and ignores
the effect of crossover over the population. To understand how
crossover may disrupt a schema, the following example is presented

If the two solution vectors $\v{x}_1=(0111010010)$, which is counted
among $\xi(S_1,t)$, and     $\v{x}_2=(0110100101)$ do exist in a
population, and the crossover operator decided that those vectors
will be crossed-over right after the sixth allele. none of the two
offspring solution vectors produced ($\v{o}_1=(0111010101)$, and
$\v{o}_2=(0110100010)$) will be matched by the schema $S_1$.
However, if the crossover operation took place right after the third
allele, the offspring will look like $\v{o}_1=(0110100101)$, and
$\v{o}_2=(0111010010)$, and there will be one solution vector
($\v{o}_2$) matched by $S_1$. Which means that
equation~\eqref{eq:EA_schm_exp} is not accurate.

It is clear that a higher defining length for a schema increase the
probability of its destruction, because it will be more likely that
the crossover point will fall between the first and the last fixed
positions of the schema. Henceforth, given the crossover rate
($p_c$), the probability of schema survival will be $1 - p_c
\frac{\delta(S)}{m-1}$. But the crossover operation may not destroy
the schema even if it splits the vectors between two fixed
positions; It is possible to crossover two solution vectors matched
by the same schema, so regardless of the crossover position, the two
resulting offspring would still be matched by that schema. So, the
probability of schema survival will slightly increase, and the
\emph{reproduction schema growth equation} will look like;
\begin{equation}
\xi(S, t+1) \ge
\left(\frac{\xi(S,t)f(S,t)}{\overline{F}(t)}\right)\left(1- p_c \,
\frac{\delta(S)}{m-1}\right)
\end{equation}

To make the above equation more accurate, the mutation operator
should be considered as well.

The mutation operator will destroy a schema only if it inverts one
of its fixed positions. So, the higher the order of a schema is, the
more likely it will be destroyed by a mutation operator. Henceforth,
given the schema order ($o(S)$) and the mutation ratio ($p_m$). The
probability that a fixed position will survive mutation is
($1-p_m$), and the probability of schema survival is
$(1-p_m)^{o(S)}$, which for normal mutation ratios ($p_m \ll 1$) can
be approximated to $(1-o(S) \times p_m)$. The \emph{reproduction
schema growth equation} will now look like:

\begin{equation}
\label{eq:EA_schm_grth}%
\xi(S, t+1) \ge
\left(\frac{\xi(S,t)f(S,t)}{\overline{F}(t)}\right)\left(1- p_c \,
\frac{\delta(S)}{m-1} - o(S) p_m \right)
\end{equation}

This equation shows how fast a schema may influence the population;
The number of individuals created using this template
\emph{exponentially} increases over time if this schema is
above-average and has sufficiently short defining length and low
order. This equation shows also that a schema with relatively short
defining length and low order will be sampled more than another
schema with longer defining length and higher order.

\begin{thrm}[Schema Theorem] Short, low-order, above-average schemata receive exponentially
increasing rate of trials in subsequent generations of a genetic
algorithm. \end{thrm}

For a better understanding of the schema theorem and theoretical
data given above ,the following example is given.

A \ac{GA} is running a population of size $N=500$, with individuals
of length $m=20$. It has a crossover and mutation ratios of
$p_c=0.7$ and $p_m=0.01$ respectively. When the optimizer was
initialized there was 5 individuals matched by the schema $S_a$
after initialization, $\xi(S_a,0)=5$.

Using the reproduction schema growth equation, given
in~\eqref{eq:EA_schm_grth}, two tables are created;
Table~\ref{tb:EA_schm_fit} illustrates the effect of the
schema/population fitness ratio on the schema growth rate, while
$\delta(S_a)=11$ and $o(S_a)=6$, and Table~\ref{tb:EA_schm_dlo}
shows the effect of different defining length and order values for
schema $S_a$ on its growth rate, while
$\frac{f(S,t)}{\overline{F}(t)}=1.9$.

The values shown in the body of the two tables are the expected
number of individuals that schema $S_a$ will match through
generations (rounded to floor), and a ``--'' entry means that the
overwhelming number of schema instances in the population caused
violation of the assumed average population fitness.

\begin{table}
\begin{center}
\caption{Effect of schema/population fitness ratio on schema growth
rate \label{tb:EA_schm_fit}}
\begin{tabular}{r | c c c c c c c c c c c}
        &\multicolumn{11}{c}{Generations}\\%
        \cline{2-12}%
\small{$f(S)/\overline{F}$}        &1   &10    &20     &30     &40     &50     &60     &70     &80     &90     &100\\%
        \hline
1.8  &5   &4     &2      &2      &1      &1      &1      &0      &0      &0      &0\\%
1.9  &5   &6     &7      &8      &9      &11     &13     &15     &18     &21     &24\\%
2.0  &5   &9     &18     &35     &69     &134    &263    &--     &--     &--     &--\\%
2.1  &5   &14    &45     &144    &--     &--     &--     &--     &--     &--     &--\\%
2.2  &5   &22    &110    &--     &--     &--     &--     &--     &--     &--     &--\\%
\end{tabular}
\end{center}
\end{table}

\begin{table}
\begin{center}
\caption{Effect of schema defining length and order on schema growth
rate \label{tb:EA_schm_dlo}}
\begin{tabular}{c | c c c c c c c c c c c}
        &\multicolumn{11}{c}{Generations}\\%
        \cline{2-12}%
$(\delta~,~o)$      &1   &10    &20     &30     &40     &50     &60     &70     &80     &90     &100\\%
        \hline
(10, 6)  &5   &11    &24     &55     &125    &285    &--     &--     &--     &--     &--\\%
(10, 7)  &5   &9     &17     &33     &63     &120    &229    &--     &--     &--     &--\\%
(10, 8)  &5   &8     &12     &19     &31     &50     &79     &127    &203    &--     &--\\%
(11, 8)  &5   &4     &3      &3      &2      &2      &1      &1      &1      &1      &1\\%
(12, 8)  &5   &2     &1      &0      &0      &0      &0      &0      &0      &0      &0\\%
\end{tabular}
\end{center}
\end{table}

The figures presented in Table~\ref{tb:EA_schm_fit} show how
increasing the schema fitness by 10\% allows the schema to switch
form vanishing (for 1.8 ratio) to exponentially increase its samples
(for 1.9 ratio). By increasing the fitness ratio further, the schema
takes-over the population more and more faster, until at a fitness
ratio of 2.2, $S_a$ takes-over the population in less than 30
generations.

Table~\ref{tb:EA_schm_dlo} shows how a slight modification of the
defining length or the order of the schema can have a significant
effect on its growth rate; A unit increase in the defining length
(from 10 to 11) leads to $S_a$ extinction. A unit increase in the
schema order affects its growth rate indeed, but not as dramatically
as its defining length.

\section{Polyploidy}
\label{ch:EAs_sec:poly}

Due to the good results obtained using first \ac{GA} representations
\cite{holland75:adapt, dejong:1975}, researchers used them without
significant modifications. Although, many living organisms carry
redundant chromosomes in there cells (polyploid organisms), the
effect of polyploidy remains under-researched in \ac{EMO}. In
nature, redundant chromosomes help organisms adapt to their
environment. When a large asteroid or comet hit earth 65 million
years ago, some species adapted to their new environment and
survived, dinosaurs failed to adapt and perished. Moreover, these
redundant chromosomes help keeping a varied set of organisms to fill
different niches in the environment that these organisms live in. A
heavy and muscled animal can defend itself against attacks but is
slower than a less heavier and less muscled animal which can catch
its prey. A trade-off between those two traits (muscles and weight)
provides animals with advantages over one another.

Polyploid species have redundant chromosomes in their cells. The
alleles which result in an organism that well fits its environment
tend more to be expressed (\emph{dominant alleles}) than the other
alleles (\emph{recessive alleles}). Meanwhile, the other alleles are
held in abeyance and rarely expressed until the environment changes
to favor one or some them and makes them the new \emph{dominant}
alleles.

Though there are dominance schemes other than simple dominance, such
as partial dominance and co-dominance, their effects were not
investigated before. This is mainly because most research on
polyploidy was carried out on binary problems. In partial dominance,
an intermediate value between parents' alleles is expressed. It
provides even more population diversity than simple dominance in
which a distinct parent allele is expressed. A classical example of
partial dominance is the color of the carnation flower that take
variants of the red color due to the presence or absence of the red
pigment allele. In the co-dominance scheme, both alleles are
expressed. A well known example for co-dominance is the Landsteiner
blood types. In this example, both `A' and `B' blood type alleles
are expressed leading to an `AB' blood type which carries both
phenotypes.

\subsection{Current Representations}
Early work examining the effect of polyploidy in GA goes back to
1967 in Bagley's dissertation \cite{Bagley:1967:BAS} as he examined
the effect of diploid representation. In his work he used a variable
dominance map encoded in the chromosome. A drawback of his model was
the premature convergence of dominance values which led to an
arbitrary tie breaking mechanism \cite{Goldberg87:diploidy}. This
work was followed by a tri-allelic dominance scheme used by
Hollstien \cite{Hollstien71} and Holland \cite{holland75:adapt}. For
each allele, they added a dominance value associated and evolved
with it. It took values of 0, a recessive 1 or a dominant 1, though
they used different symbols.

Unlike previously mentioned work which was done on stationary
environments, Goldberg and Smith \cite{Goldberg87:diploidy} used a
non-stationary environment. They used a 0-1 knapsack problem with
two evolving limiting weights. They concluded that the power of
polyploidy is in non-stationary problems because of the abeyance of
recessive alleles that remember past experiments. However, they did
not show the performance of their algorithm in remembering more than
two oscillating objectives. This was a big shortcoming, because most
real world non-stationary problems are non-cycling problems, they
may come out of order, and sometimes never repeat. Ng and Wong
\cite{Ng:1995:Diploid} argued that the enhanced performance in
\cite{Goldberg87:diploidy} was due to the slow convergence
encountered in the diploid representation. They proposed a dominance
scheme that used dominant (0, 1) and recessive (0, 1) alleles and
inverted the dominance of alleles whenever the individual's fitness
fall below a 20\% threshold value. They reported enhanced
performance over tri-allelic representation. Some researchers
extended the application of polyploidy beyond GA. Polyploidy was
applied to Genetic Programming (GP) as well
\cite{Cavill05:multichromGP}.

All the work previously mentioned were conducted for single
objective optimization problems. Most of the researchers used
relatively low number of decision variables and binary problems such
as the 0-1 knapsack problem. At the same time many of these
investigations were concerned with manipulating the simple dominance
scheme and comparing its variants. The monoploid number (number of
chromosomes in each solution vector) remained constant in most of
these investigations. Scarce applications such as
\cite{massebeuf99multicriteria} were done on Multi-Objective
Optimization Problems (MOOP). In \cite{massebeuf99multicriteria},
diploid vectors were used to search in 2-dimensional space
optimizing 3 objectives for a food extrusion process. First, they
worked on each objective separately and produced offspring better
than the worst individual for this objective (otherwise the
offspring is killed) and then Pareto dominance was applied to the
combined populations of each objective.

What is common among all previous work is that they used simple
dominance where the allele either dominates or recesses, which is
not always the case in nature. Partial dominance and co-dominance
were not investigated before. Partial dominance produces new
phenotypes that help the population to adapt to totally new
environments and to remember them. It provides more phenotype
diversity than simple dominance. Henceforth, a new model is provided
here which differs from other polyploid models
\cite{holland75:adapt, Hollstien71, Goldberg87:diploidy,
Bagley:1967:BAS, Ng:1995:Diploid}. In this model;
\begin{inparaenum}[i)]
\item Uniform crossover of each parent chromosomes precedes
recombination.%
\item The offspring alleles are created from their parents' alleles
using partial dominance.%
\item The algorithm is applicable to non-binary problems.
\end{inparaenum}

\subsection{Proposed Representation and Procedure}
\begin{figure}
\centering
\includegraphics[width = \textwidth]{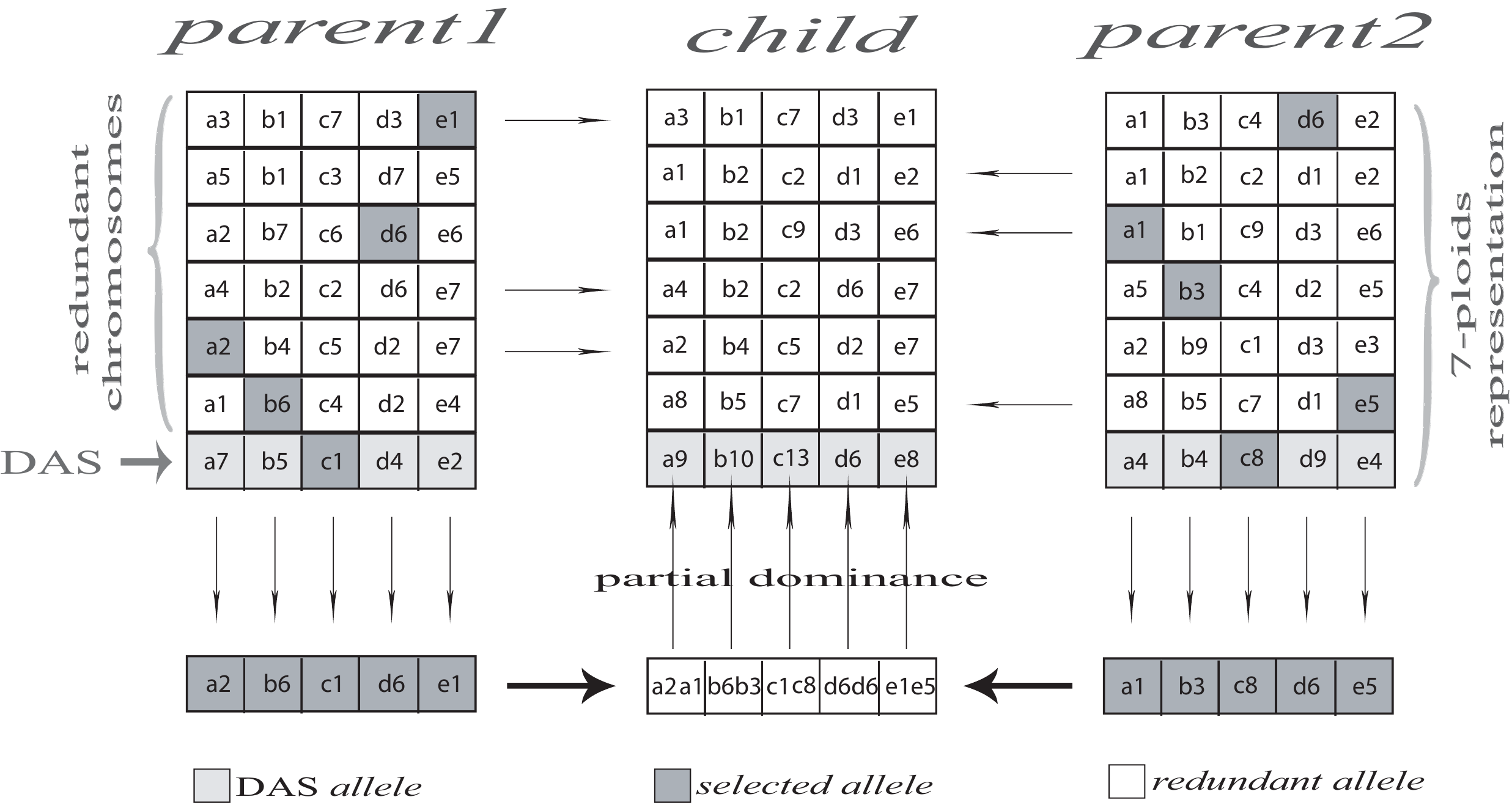}
\caption{The Polyploid mating procedure \label{fig:EA_poly_proc}}
\end{figure}

In biology, ploidy is the number of sets of chromosomes in a cell. A
cell that has one set of chromosomes is a monoploid cell, while the
one that has two sets is a diploid cell, three sets make it triploid
and so on. The term ``d-ploid'' is used to indicate the ploidy
number. So, 1-ploid representation is a monoploid one, 2-ploid
representation is a diploid one and so on. In this algorithm the
partial dominance scheme is used. The phenotype of each solution
vector is based on the partial dominant alleles or the \ac{DAS} as
shown in Figure~\ref{fig:EA_poly_proc}. The \ac{DAS} of any solution
vector determines its fitness value. The ploidy number $d$ of a
solution vector is the number of all chromosomes in that vector
including the \ac{DAS} chromosome. A detailed procedure of the
algorithm is given as follows;

\begin{enumerate}[i)]
\item \emph{Initialization}: For a $d$-ploid representation.
A population of size $N$ is initialized at random by filling each
\ac{DAS} and the $d-1$ chromosomes for each of the $N$ solution
vectors. Then, the fitness of each vector is evaluated according to
its \ac{DAS}. The domination of solution vectors is determined based
on Pareto dominance.

\item \emph{Mating Selection}: Only non-dominated solution vectors
are selected for mating. This will result in a varying mating pool
size and consequently a varying offspring size for each generation,
leading to high selection pressure.

\item \emph{Variation Operators}: After filling the mating pool,
two parents are selected at random from the mating pool without
replacement. Then, for each parent, an allele representing each
locus of a chromosome is randomly selected from all available
alleles at this locus to create two sets of alleles (one for each
parent). Any recombination operator may be applied to those two sets
to create the child's \ac{DAS}. Here, the \ac{SBX}
\cite{deb95simulated} operator is used. Then, the child's \ac{DAS}
is mutated. Any mutation operator may be used here. Polynomial
mutation is used here \cite{deb95simulated}. After the child's
\ac{DAS} is created, the remaining $d-1$ chromosomes of this child's
solution vector are selected and copied at random from both parents'
chromosomes. All chromosomes in both parents have the same
probability of being selected and copied to carry their genes to
later generations. No mutation is applied on these redundant $d-1$
chromosomes.

\item \emph{Survival Selection}: After evaluating the fitness of the offspring,
parents and offspring fight for survival as Pareto dominance is
applied to the combined population of parents and offspring. The
least dominated $N$ solution vectors (according to number of
solutions dominating them) survive to make the population of the
next generation. Ties are resolved at random.
\end{enumerate}

Figure~\ref{fig:EA_poly_proc} shows the structure of solution
vectors for the Polyploid algorithm. The solution vector
\texttt{parent1} contains the \ac{DAS} chromosome (the seventh row)
plus six more redundant chromosomes (rows 1--6), so this is $6 + 1
=$7-ploids representation. Each one of those seven chromosomes
contains five alleles. The \ac{DAS} chromosome of \texttt{parent1}
vector is \texttt{[a7 b5 c1 d4 e2]}, where \texttt{a7} is the first
allele, \texttt{b5} is the second allele, \texttt{c1}, \texttt{d4}
and \texttt{e2} are the third, fourth and fifth alleles.

After \texttt{parent1} and \texttt{parent2} are selected for mating,
for each one of them, an allele is selected to represent each locus
(column) of this solution vector as shown in
Figure~\ref{fig:EA_poly_proc}. For \texttt{parent1}, \texttt{a2} is
selected to represent the first locus (column), \texttt{b6} to
represent the second locus (column), \texttt{c1}, \texttt{d6} and
\texttt{e1} to represent the third, fourth and fifth loci (columns).
Henceforth, \texttt{paretnt1} is represented by the chromosome
\texttt{[a2 b6 c1 d6 e1]}. By following the same procedure, the
\texttt{[a1 b3 c8 d6 e5]} chromosome is produced for
\texttt{parent2} . Next, partial dominance is applied on these two
chromosomes to get the \ac{DAS} chromosome of the child solution
vector. \ac{SBX} \cite{deb95simulated} is used to simulate partial
dominance, then polynomial mutation is applied on the produced
\ac{DAS} chromosome. After the child's \ac{DAS} chromosome is
created and mutated, the remaining 6 chromosomes of the child's
solution vector are filled by copying 6 chromosomes from both
parents at random to complete the mating process.

\subsection{Experiments}
\begin{figure}
\centering
\includegraphics{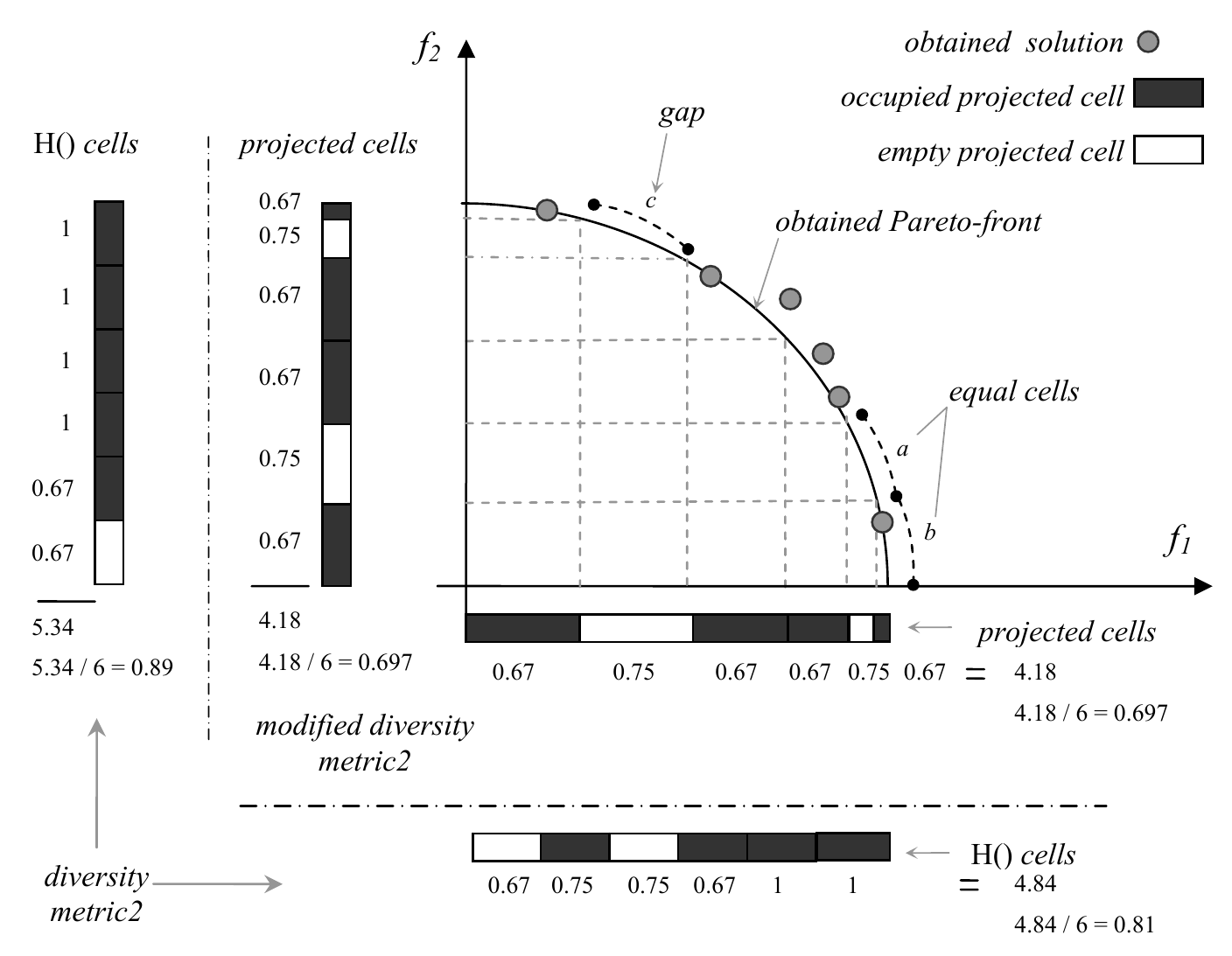}
\caption{Modified diversity metric2 for DTLZ2
\label{fig:EA_mod_divmet2}}
\end{figure}

In the following experiments the convergence and diversity of
solution vectors are measured using two running metrics to
understand the behavior of the algorithms. The average orthogonal
distance of solution vectors to the \ac{PF} is used to measure
convergence because the equations of the global front are known in
advance. While the \emph{diversity metric2} presented in
\cite{Khare03:PerfScalMOEA} is modified and used here to measure
diversity of solutions. The modified \emph{diversity metric2} is
explained as follows:
\begin{enumerate}[i)]
\item For a given objective, the obtained \ac{PF} using a population of size $N$ is
divided uniformly creating $N$ equal cells on the front surface,
such as cells $a$, $b$ and $c$ shown in
Figure~\ref{fig:EA_mod_divmet2}. The projection of these cells on
the current objective dimension gives $N$ projected cells.

\item The obtained solutions are projected as well on this objective dimension.

\item For every projected cell that contains one or more projected solutions an
occupation value of 1 is assigned to it, an occupation value of 0 is
assigned to the projected cell otherwise.

\item A diversity value is assigned to each non-boundary projected cell using a sliding
window according to the values shown in Table~\ref{tb:EA_nonb_div},
where the cell index is $n$. While diversity values for boundary
cells are assigned according to Table~\ref{tb:EA_b_div}, where the
boundary cell index is $k$ for the left boundary, and is $k+1$ for
the right boundary.

\item The cells' diversity values are added and divided by the number of cells to give the
current objective's diversity value in the range $(0, 1]$.

\item Previous steps are repeated for the remaining objectives.

\item The average diversity value of all objectives is the overall diversity value of the
population. The best population distribution yields a diversity
value of 1, while this value approaches 0 for the worst
distribution.
\end{enumerate}

\begin{table}
\centering
\begin{minipage}[t]{0.45\textwidth}
\centering \caption{Non-boundary cells' diversity values
\label{tb:EA_nonb_div}}
\begin{tabular}{ccc|c}
\hline%
\multicolumn{3}{c|}{occupation values}  &diversity\\%
\cline{1-3}%
cell($n-1$)    &cell($n$) &cell($n+1$)  &values\\%
\hline \hline%
0   &0  &0  &0\\%
0   &0  &1  &0.5\\%
0   &1  &0  &0.75\\%
0   &1  &1  &0.67\\%
1   &0  &0  &0.5\\%
1   &0  &1  &0.75\\%
1   &1  &0  &0.67\\%
1   &1  &1  &1\\%
\hline
\end{tabular}
\end{minipage}
\hfill
\begin{minipage}[t]{0.45\textwidth}
\caption{Boundary cells' diversity values \hspace{5ex}
\label{tb:EA_b_div}} \centering
\begin{tabular}{cc|c}
\hline%
\multicolumn{2}{c|}{occupation values}  &diversity\\%
\cline{1-2}%
cell($k$) &cell($k+1$)  &values\\%
\hline \hline%
0   &0  &0\\
0   &1  &0.67\\
1   &0  &0.67\\
1   &1  &1\\
\hline
\end{tabular}
\end{minipage}
\end{table}

Figure~\ref{fig:EA_mod_divmet2} compares diversity evaluation for
the modified and unmodified \emph{diversity metric2} for a
population of size 6 on a 2-dimensional DTLZ2 \cite{Deb02a:DTLZs}
minimization problem. To evaluate the diversity value for objective
$f_1$ using the modified \emph{diversity metric2}, the obtained
\ac{PF} is uniformly divided into 6 equal cells such as cells
\emph{a}, \emph{b} and \emph{c}. These equal cells are projected on
the $f_1$ axis giving 6 unequal projected cells. The obtained
solutions are projected on the $f_1$ axis as well, and for each
projected cell containing one or more projected solutions an
occupation value of 1 is assigned (occupied projected cell), an
occupation value of 0 is assigned otherwise (empty projected cell).
Figure~\ref{fig:EA_mod_divmet2} shows how the modified
\emph{diversity metric2} was able to detect the two gaps at cells
\emph{a} and \emph{c} on the \ac{PF} and detected all the other
occupied cells on the front. The occupation values vector produced
is [1 0 1 1 0 1]. Using a sliding window of size 3 with the values
presented in Table~\ref{tb:EA_nonb_div}, diversity values for each
non-boundary projected cell with index $n$ is assigned, while the
boundary projected cells are assigned diversity values according to
Table~\ref{tb:EA_b_div}. The diversity values vector produced is
[0.67 0.75 0.67 0.67 0.75 0.67]. These 6 values are summed and
divided by 6 (the maximum possible sum of the 6 diversity values
resulting from an ideal solutions distribution) to get a normalized
diversity value for the $f_1$ objective ($\frac{4.18}{6} = 0.697$).
This procedure is used to evaluate the diversity value for the $f_2$
objective as well. By taking the average of $f_1$ and $f_2$
diversity values, the overall diversity value of the population is
$\frac{0.697+0.697}{2} = 0.697$.

Figure~\ref{fig:EA_mod_divmet2} shows how the unmodified
\emph{diversity metric2} was unable to detect the gap at cell
\emph{a} in the obtained \ac{PF} when the solutions are projected on
$f_1$, and falsely detected a gap close to the boundary $f_1=0$.
Furthermore, the unmodified \emph{diversity metric2} was unable to
detect the gaps at cells \emph{a} and \emph{c} when the solutions
are projected on $f_2$, and again falsely detected a gap close the
boundary $f_2=0$. This shortcoming in the unmodified version is due
to its poor cell partitioning. It ignores the shape of the \ac{PF},
as a consequence, regions with higher slope are under represented in
$H()$ cells and gaps are overlooked in these regions, while regions
with lower slope get over represented in $H()$ cells leading to
false gap detection. The modification presented here exploits the
knowledge of the \ac{PF} in benchmark problems. The size of each
projected cell is inversely proportional to the slope of its
corresponding \ac{PF} cell.

The mapping of occupation values to cell diversity values is shown
in Table~\ref{tb:EA_nonb_div} for non boundary cells. This mapping
is made so that occupation vector [0 0 0] gets the lowest diversity
value (0) because it reflects an empty region (no solutions in three
consecutive cells), while occupation vectors [1 0 0] and [0 0 1] get
a higher diversity value (0.5) because only one occupied cell
appeared at the edge (non uniform distribution). Occupation vectors
[1 1 0] and [0 1 1] get a diversity value of 0.67 because they have
two occupied cells albeit not uniformly distributed, while the
occupation vectors [1 0 1] and  [0 1 0] get a higher diversity value
(0.75) due to the uniform distribution of their occupied cells. The
best diversity value (1) is assigned to occupation vector [1 1 1]
because all of its cells are occupied. A similar mapping is shown in
Table~\ref{tb:EA_b_div} for boundary cells. Occupation vectors [0 0]
and [1 1] get the lowest (0) and highest (1) diversity values
respectively, while occupation vectors [1 0] and [0 1] get an
in-between diversity value (0.67).

In the following experiments, some of the DTLZ benchmark problems
set \cite{Deb02a:DTLZs} are used to investigate the effect of
polyploidy on convergence and diversity. The first test problem
(DTLZ1) has a huge number of local fronts, but a simple linear
\ac{PF}, while the second test problem (DTLZ2) has a nonlinear
(spherical) \ac{PF}. The third test problem (DTLZ3) combines the
difficulties of DTLZ1 and DTLZ2. These three test problems test the
ability of the algorithm to escape local optima and converge to the
global front with varying degrees of problem difficulty. The fourth
test problem (DTLZ4) has a biased distribution of solutions along
the \ac{PF}. This problem tests the ability of the algorithm to
maintain a good distribution along the front. Using this varied set
of test problems a fair analysis of the behavior of the algorithms
regarding convergence and diversity is conducted.

The \ac{NSGA-II} and the Polyploid algorithm differ mainly in two
aspects. The first one is the absence of an explicit diversity
maintenance mechanism in the Polyploid algorithm unlike the
\ac{NSGA-II} algorithm which has a computationally expensive
diversity maintenance mechanism. The second one is the redundant
chromosomes in the Polyploid algorithm which have no counterpart in
the \ac{NSGA-II} algorithm. The following experiments tests whether
the redundant chromosomes in the Polyploid algorithm will compensate
for the absence of an explicit diversity maintenance mechanism and
help the algorithm maintain a good diversity or not. Moreover, the
effect of these redundant chromosomes on the speed of convergence to
the \ac{PF} is to be investigated.

All the test problems were run with 3, 4, 6, and 10 objectives and
with a population of size $N=100$. The following parameters were
empirically found to produce the best results for the \ac{NSGA-II}
and the Polyploid algorithm on the test problems used. For all test
problems, \ac{SBX} is used for recombination with $p_c = 1$ and
$\eta_c = 20$, and the polynomial mutation is used with $p_m =
\frac{1}{n}$ and $\eta_m= 15$, where $n$ is the number of decision
variables used.

\subsection{DTLZ1}

The first test problem to be used is DTLZ1 shown in
\eqref{eq:EA_poly_DTLZ1a} and \eqref{eq:EA_poly_DTLZ1b}.
\begin{gather}
\left.
\begin{aligned}
\text{Minimize}&    &&f_1(\v{x})=\frac{1}{2}(1 + g(\v{x}_M)) x_1 x_2
\dotsm x_{M-1},\\%
\text{Minimize}&    &&f_1(\v{x})=\frac{1}{2}(1 + g(\v{x}_M)) x_1 x_2
\dotsm (1 - x_{M-1}),\\%
\vdots&            &&\vdots\\%
\text{Minimize}&    &&f_{M-1}(\v{x})=\frac{1}{2}(1 + g(\v{x}_M)) x_1
(1 - x_2),\\%
\text{Minimize}&    &&f_{M}(\v{x})=\frac{1}{2}(1 + g(\v{x}_M))(1 -
x_1),\\%
\text{subject to}&  &&0 \leq x_i \leq 1, \quad \text{for} \; i =
1,2, \dotsc, n. \label{eq:EA_poly_DTLZ1a}
\end{aligned}
\right\}%
\\ g(\v{x}_M) = 100 \left[ |\v{x}_M| + \sum_{x_i \in \v{x}_M} (x_i -
0.5)^2 - \cos(20\pi(x_i - 0.5)) \right]. \label{eq:EA_poly_DTLZ1b}
\end{gather}
Where $\v{x}$ and $x_i$ are a decision vector and a decision
variable, respectively. The function $g(\v{x}_M)$ requires
$|\v{x}_M|=k$ variables, and the total number of variables is $n = k
+ M - 1$, where $M$ is the number of objectives.

This problem has a simple linear \ac{PF} but has a huge number of
local optima. There exists $11^{n - M + 1} - 1$ local fronts. The
high number of decision variables ($n = 40$) creates a huge number
of local optima to test the ability of the algorithm to escape them.
The Pareto-optimal solutions for this problem correspond to $x_i^* =
0.5$, ($x_i^* \in \v{x}_M$), while the objective functions lie on
the linear hyper-plane $\sum_{m=1}^M f_m^* = 0.5$. The algorithms
were run for 50,000 function evaluations to analyze the algorithms'
performance in long run.

\begin{figure}
\centering%
\subfloat[3 objectives
\label{fig:EA_poly_DTLZ1_conv_3}]{\includegraphics[width =
0.48\textwidth]{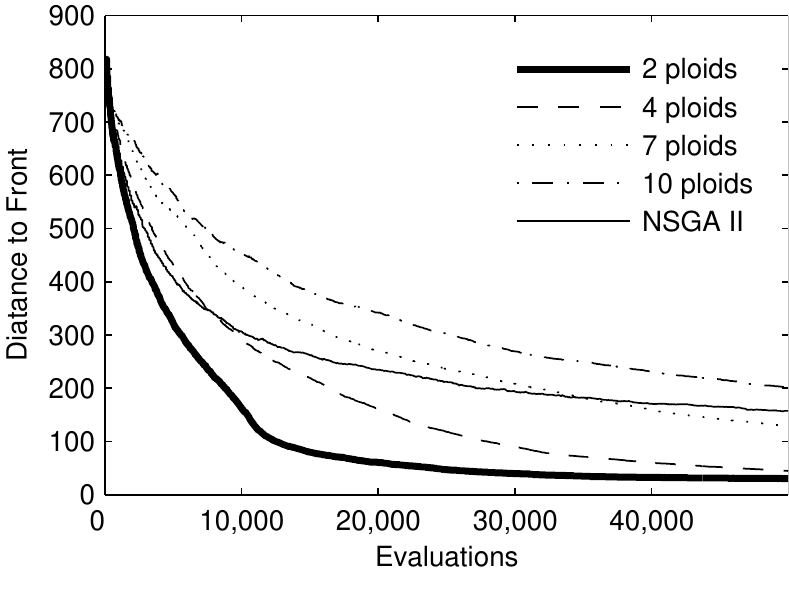}}%
\hfill%
\subfloat[4 objectives \label{fig:EA_poly_DTLZ1_conv_4}]{\includegraphics[width = 0.48\textwidth]{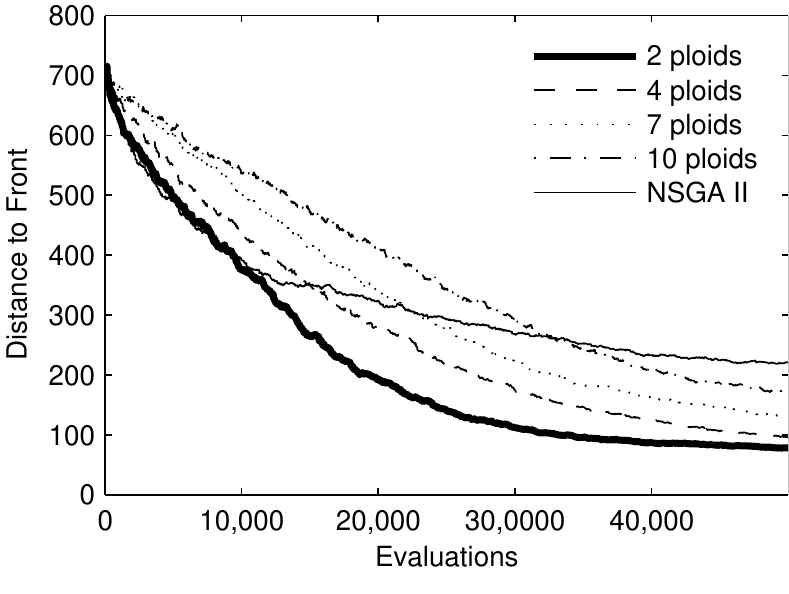}}%
\caption[Convergence speed for DTLZ1]{Convergence speed for DTLZ1 \label{fig:EA_poly_DTLZ1_conv}}%
\end{figure}

As shown in Figure~\ref{fig:EA_poly_DTLZ1_conv}, \ac{NSGA-II} has
the best convergence speed in early evaluations, but it get overcome
by the Polyploid algorithms one after the other, except for the
10-ploids algorithm in the 3 objectives case which does not catch it
in the scope of the 50,000 function evaluations
(Figure~\ref{fig:EA_poly_DTLZ1_conv_3}). However by increasing the
number of objectives, the Polyploid algorithms catch \ac{NSGA-II}
earlier in the run. In the 3 objectives case, the 7-ploids algorithm
catches \ac{NSGA-II} after around 35,000 evaluations, while the
10-ploids algorithm is unable to catch it
(Figure~\ref{fig:EA_poly_DTLZ1_conv_3}). But in the 4 objectives
case, the 7-ploids algorithm overcomes \ac{NSGA-II} after around
23,000 evaluations while the 10-ploids algorithm does so after
33,000 evaluations (Figure~\ref{fig:EA_poly_DTLZ1_conv_4}). This
reveals the ability of the Polyploid algorithms to handle many
objectives in rugged objective functions. This ability is magnified
when the problem gets more difficult.

Table~\ref{tb:EA_DTLZ1_div} shows the diversity of the algorithms
after 50,000 function evaluations. \ac{NSGA-II} has the highest
diversity values in the case of 4, 6 and 10 objectives, while the
7-ploids algorithm outperforms the other algorithms in the 3
objectives problem. The diversity of the Polyploid algorithms is
slightly increasing with increasing the number of objectives. A
diversity value of 0.65 for the 2-ploids algorithm in 3 objectives
reaches 0.79 in the 10 objectives case.

Table~\ref{tb:EA_DTLZ1_conv} shows the convergence of the algorithms
after 50,000 function evaluations. The 2- ploids algorithm
outperforms all the other algorithms in the 3, 4, and 6 objectives,
while in the 10 objectives case, the 4-ploids algorithm is the best
achieving a distance of 382 compared to 389 for the 2-ploids which
comes second.

By comparing the Polyploid algorithms performance regarding the two
performance criteria (convergence speed and diversity maintenance);
increasing the ploidy number slightly increases the diversity of
most Polyploid algorithms (maximum increase = 0.0618). On the other
hand, a higher ploidy number slows down the convergence speed.

\begin{table}
\centering%
\caption[Diversity values for DTLZ1]{Diversity after 50,000 function
evaluations for DTLZ1 \label{tb:EA_DTLZ1_div}}
\begin{tabular}{l|l||c|c|c|c}
\cline{3-6}%
\multicolumn{2}{l}{}    &\multicolumn{4}{|c}{objectives}\\%
\hhline{--||----}%
algorithm   &measure    &3      &4      &6      &10\\%
\hhline{==::====}%
\raisebox{-1.5ex}[0cm][0cm]{2-ploids}
            &Average    &0.6506 &0.6863 &0.7067 &0.7942\\%
            &Std. Dev.  &0.0002 &0.0003 &0.0002 &0.0003\\%
\hhline{--||----}%
\raisebox{-1.5ex}[0cm][0cm]{4-ploids}
            &Average    &0.7065 &0.7075 &0.7043 &0.7907\\%
            &Std. Dev.  &0.0004 &0.0001 &0.0005 &0.0001\\%
\hhline{--||----}%
\raisebox{-1.5ex}[0cm][0cm]{7-ploids}
            &Average    &{\bf 0.7124} &0.7141 &0.7480 &0.8036\\%
            &Std. Dev.  &0.0002 &0.0002 &0.0002 &0.0003\\%
\hhline{--||----}%
\raisebox{-1.5ex}[0cm][0cm]{10-ploids}
            &Average    &0.6993 &0.6953 &0.7201 &0.8198\\%
            &Std. Dev.  &0.0003 &0.0001 &0.0003 &0.0001\\%
\hhline{--||----}%
\raisebox{-1.5ex}[0cm][0cm]{NSGA-II}
            &Average    &0.6916 &{\bf 0.7945} &{\bf 0.9312} &{\bf 0.8722}\\%
            &Std. Dev.  &0.0914 &0.0312 &0.0195 &0.0099\\%
\hhline{--||----}%
\end{tabular}
\end{table}

\begin{table}
\centering%
\caption[Convergence values for DTLZ1]{Convergence after 50,000
function evaluations for DTLZ1 \label{tb:EA_DTLZ1_conv}}
\begin{tabular}{l|l||c|c|c|c}
\cline{3-6}%
\multicolumn{2}{l}{}    &\multicolumn{4}{|c}{objectives}\\%
\hhline{--||----}%
algorithm   &measure    &3      &4      &6      &10\\%
\hhline{==::====}%
\raisebox{-1.5ex}[0cm][0cm]{2-ploids}%
            &Average    &{\bf 30.370} &{\bf 78.380} &{\bf 351.08} &389.06\\%
            &Std. Dev.  &0.0785 &0.0782 &0.2550 &0.3853\\%
\hhline{--||----}%
\raisebox{-1.5ex}[0cm][0cm]{4-ploids}%
            &Average    &44.550 &95.390 &418.80 &{\bf 382.07}\\%
            &Std. Dev.  &0.1756 &0.0828 &0.3139 &0.1801\\%
\hhline{--||----}%
\raisebox{-1.5ex}[0cm][0cm]{7-ploids}%
            &Average    &127.96 &129.97 &447.77 &426.67\\%
            &Std. Dev.  &0.1571 &0.1657 &0.1966 &0.1564\\%
\hhline{--||----}%
\raisebox{-1.5ex}[0cm][0cm]{10-ploids}%
            &Average    &201.45 &167.09 &466.41 &414.77\\%
            &Std. Dev.  &0.3229 &0.0905 &0.2295 &0.0647\\%
\hhline{--||----}%
\raisebox{-1.5ex}[0cm][0cm]{NSGA-II}%
             &Average   &157.00 &219.09 &442.94 &485.17\\%
             &Std. Dev. &12.090 &9.4720 &25.340 &3.6930\\%
\hhline{--||----}%
\end{tabular}
\end{table}

\subsection{DTLZ2}
The second benchmark problem to be used is DTLZ2, which is shown in
\eqref{eq:EA_poly_DTLZ2a} and \eqref{eq:EA_poly_DTLZ2b}.
\begin{gather}
\left.
\begin{aligned}
\text{Minimize}&    &&f_1(\v{x})=(1 + g(\v{x}_M)) \cos(x_1\frac{\pi}{2}) \cos(x_2\frac{\pi}{2}) \dotsm \cos(x_{M-2}\frac{\pi}{2}) \cos(x_{M - 1}\frac{\pi}{2}),\\%
\text{Minimize}&    &&f_2(\v{x})=(1 + g(\v{x}_M)) \cos(x_1\frac{\pi}{2}) \cos(x_2\frac{\pi}{2}) \dotsm \cos(x_{M-2}\frac{\pi}{2}) \sin(x_{M - 1}\frac{\pi}{2}),\\%
\text{Minimize}&    &&f_3(\v{x})=(1 + g(\v{x}_M)) \cos(x_1\frac{\pi}{2}) \cos(x_2\frac{\pi}{2}) \dotsm \sin(x_{M-2}\frac{\pi}{2}),\\%
\vdots&             &&\vdots\\%
\text{Minimize}&    &&f_{M-1}(\v{x})=(1 + g(\v{x}_M)) \cos(x_1\frac{\pi}{2}) \sin(x_2\frac{\pi}{2}),\\%
\text{Minimize}&    &&f_{M}(\v{x})=(1 + g(\v{x}_M)) \sin(x_1\frac{\pi}{2}),\\%
\text{subject to}&  &&0 \leq x_i \leq 1, \quad \text{for} \; i =
1,2, \dotsc, n. \label{eq:EA_poly_DTLZ2a}
\end{aligned}
\right\}%
\\ g(\v{x}_M) = \sum_{x_i \in \v{x}_M} (x_i - 0.5)^2. \label{eq:EA_poly_DTLZ2b}
\end{gather}
Where $\v{x}$ and $x_i$ are a decision vector and a decision
variable, respectively. The function $g(\v{x}_M)$ requires
$|\v{x}_M|=k$ variables, and the total number of variables is $n = k
+ M - 1$, where $M$ is the number of objectives.

The \ac{PF} of this problem corresponds to $x_i^* = 0.5$, ($x_i \in
\v{x}_M$) which leads to $g(\v{x}_M)=0$ and $\sum_{m=1}^M
(f_m^*)^2=1$. For this problem, the number of decision variables is
set to $n=30$.

\begin{figure}
\centering%
\subfloat[3 objectives \label{fig:EA_poly_DTLZ2_conv_3}]{\includegraphics[width = 0.48\textwidth]{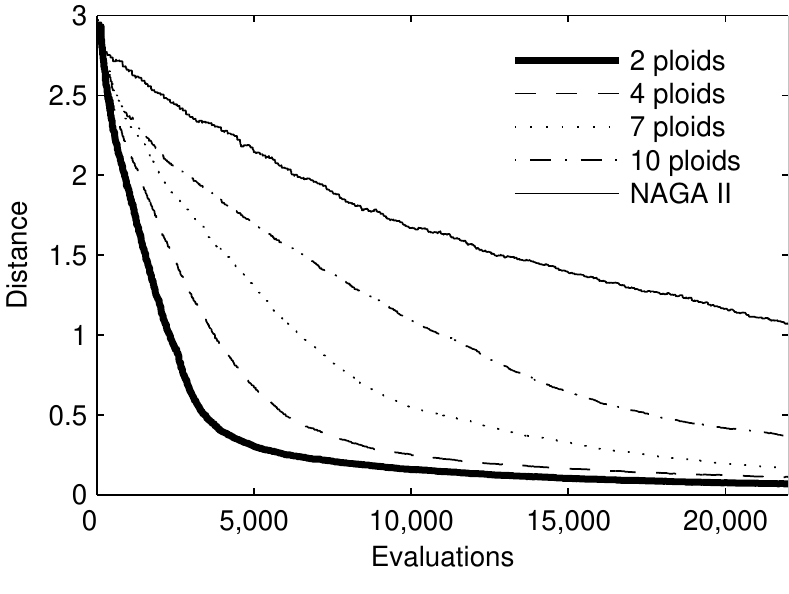}}%
\hfill%
\subfloat[4 objectives \label{fig:EA_poly_DTLZ2_conv_4}]{\includegraphics[width = 0.48\textwidth]{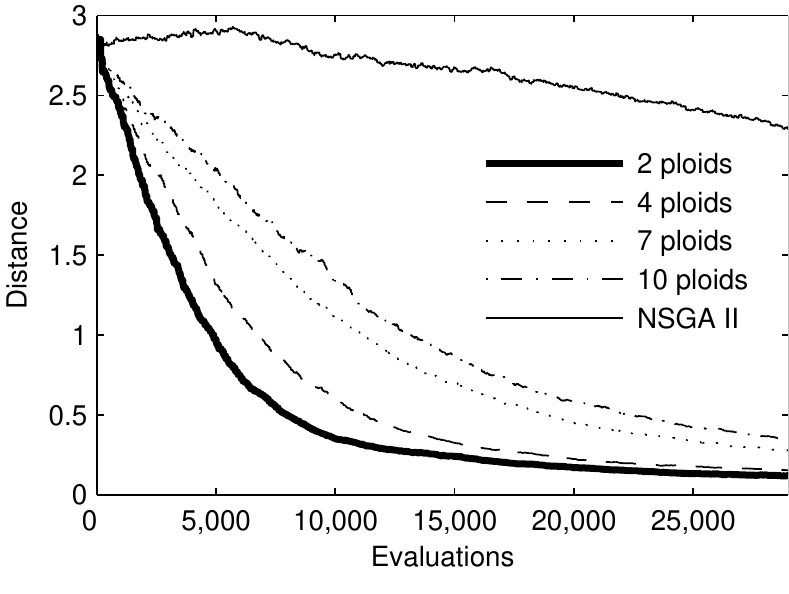}}\\%
\subfloat[10 objectives \label{fig:EA_poly_DTLZ2_conv_10}]{\includegraphics[width = 0.48\textwidth]{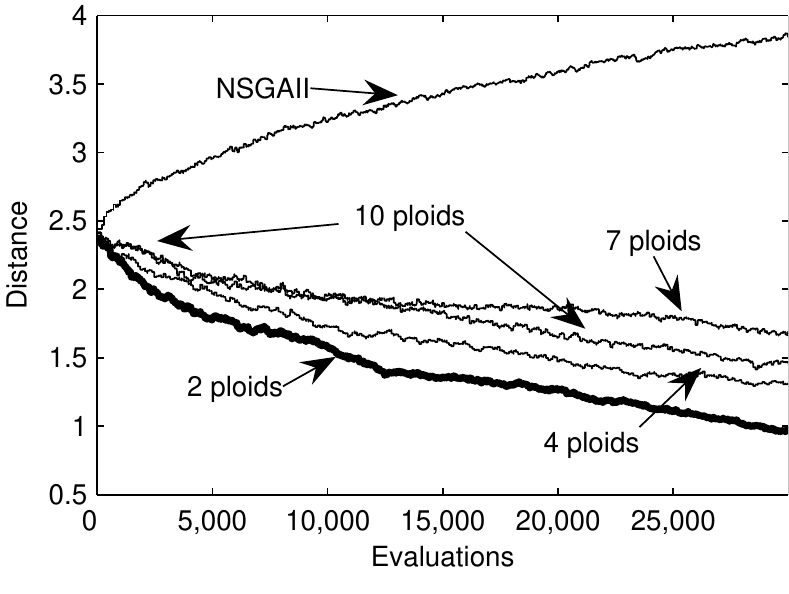}}%
\caption[Convergence speed for DTLZ2]{Convergence speed for DTLZ2
\label{fig:EA_poly_DTLZ2_conv}}
\end{figure}

Figure~\ref{fig:EA_poly_DTLZ2_conv} shows the distance to \ac{PF}
achieved by the algorithms against function evaluations. As with the
case of DTLZ1, the 2-ploids algorithm is the best performer. It is
converging faster than the other algorithms used. However, the
performance of the optimizers get closer as the number of objectives
increase. In the case of 10 objectives
(Figure~\ref{fig:EA_poly_DTLZ2_conv_10}) the 10-ploids algorithm
overcomes the 7-ploids algorithm after 10,000 function evaluations.
Figure~\ref{fig:EA_poly_DTLZ2_conv_3}--\subref{fig:EA_poly_DTLZ2_conv_10}
shows how the performance of \ac{NSGA-II} is deteriorating by
increasing the number of objectives, that in the case of 10
objectives the algorithm is diverging
(Figure~\ref{fig:EA_poly_DTLZ2_conv_10}).

Table~\ref{tb:EA_DTLZ2_div} shows the diversity of the different
algorithms. Although \ac{NSGA-II} achieved the best diversity values
in the 3, 4, 6 objective problems, its performance is declining with
increasing the number of objectives. On the other hand, the
performance of the Polyploid algorithms is either steady or
improving with increasing the number of objectives. In the 3
objectives case, \ac{NSGA-II} has a superior diversity value of
0.799, while the 7-ploids algorithm achieves a lower diversity value
of 0.6452. But for the 10 objectives problem, the 7-ploids algorithm
overcomes \ac{NSGA-II} by achieving a value of 0.7398 while
\ac{NSGA-II} achieves 0.7194.

\begin{table}
\centering%
\caption{Effect of varying ploidy number on diversity for DTLZ2
\label{tb:EA_DTLZ2_div}}%
\begin{tabular}{l|l||c|c|c|c}
\cline{3-6}%
\multicolumn{2}{l}{}    &\multicolumn{4}{|c}{objectives}\\%
\hhline{--||----}%
algorithm   &measure    &3      &4      &6      &10\\%
\hhline{==::====}%
\raisebox{-1.5ex}[0cm][0cm]{2-ploids}%
            &Average    &0.6621 &0.6594 &0.6263 &0.6872\\%
            &Std. Dev.  &0.0208 &0.0744 &0.0543 &0.0721\\%
\hhline{--||----}%
\raisebox{-1.5ex}[0cm][0cm]{4-ploids}%
            &Average    &0.6622 &0.6539 &0.6610 &0.7198\\%
            &Std. Dev.  &0.0114 &0.0212 &0.0416 &0.0492\\%
\hhline{--||----}%
\raisebox{-1.5ex}[0cm][0cm]{7-ploids}%
            &Average    &0.6452 &0.6807 &0.6773 &{\bf 0.7398}\\%
            &Std. Dev.  &0.0432 &0.0229 &0.0175 &0.0115\\%
\hhline{--||----}%
\raisebox{-1.5ex}[0cm][0cm]{10-ploids}%
            &Average    &0.5865 &0.6459 &0.6881 &0.7153\\%
            &Std. Dev.  &0.0449 &0.0229 &0.0232 &0.0338\\%
\hhline{--||----}%
\raisebox{-1.5ex}[0cm][0cm]{NSGA-II}%
            &Average    &{\bf 0.7997} &{\bf 0.7750} &{\bf 0.7479} &0.7194\\%
            &Std. Dev.  &0.0148 &0.0159 &0.0109 &0.0082\\%
\hhline{--||----}%
\end{tabular}
\end{table}

\begin{figure}
\centering%
\subfloat[2-ploids \label{fig:EA_poly_DTLZ2_PF_2front}]{\includegraphics[width = 0.48\textwidth]{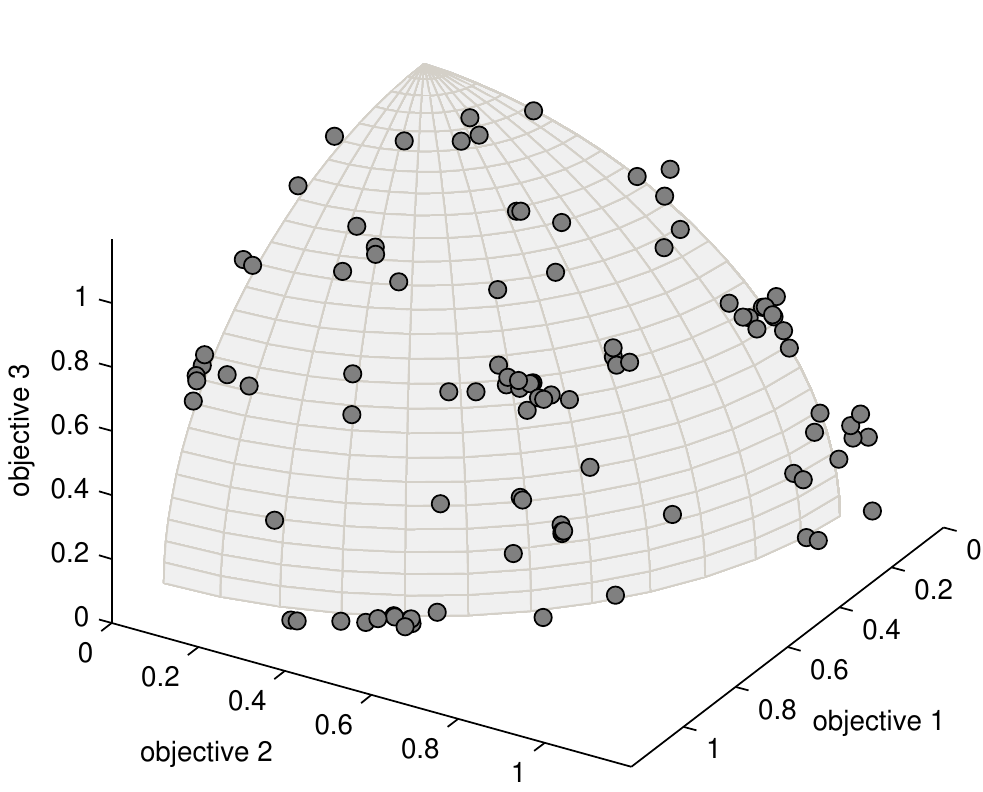}}%
\subfloat[2-ploids (side view)\label{fig:EA_poly_DTLZ2_PF_2side}]{\includegraphics[width = 0.48\textwidth]{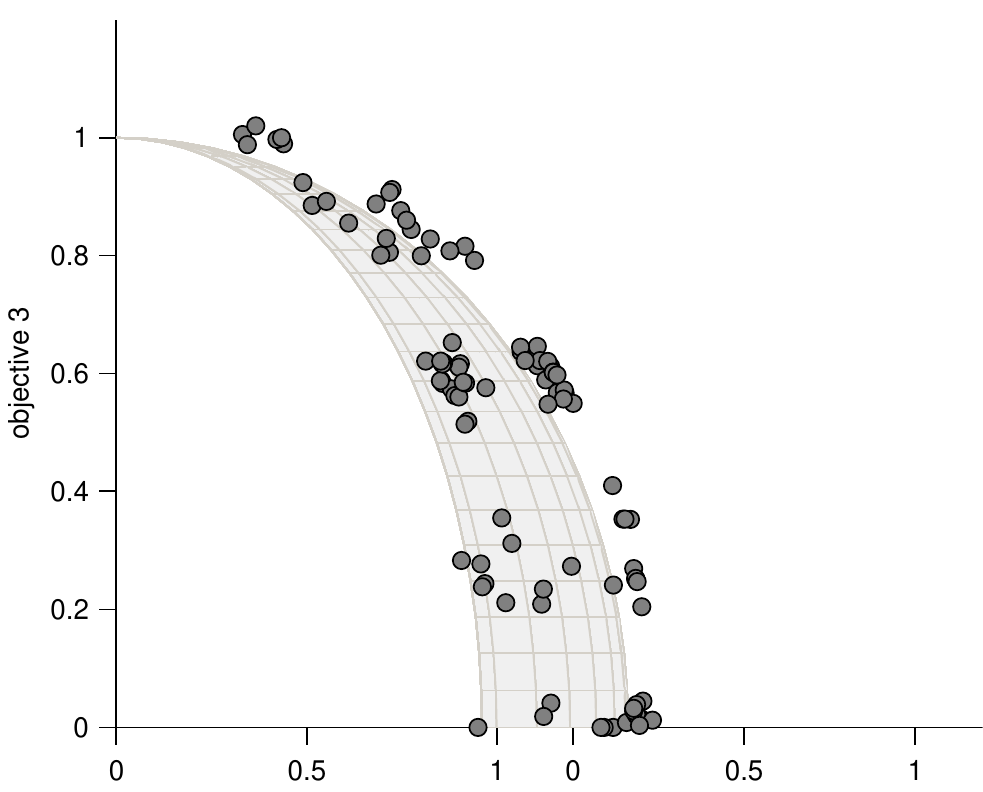}}\\%
\subfloat[NSGA-II \label{fig:EA_poly_DTLZ2_PF_NSGAII}]{\includegraphics[width = 0.48\textwidth]{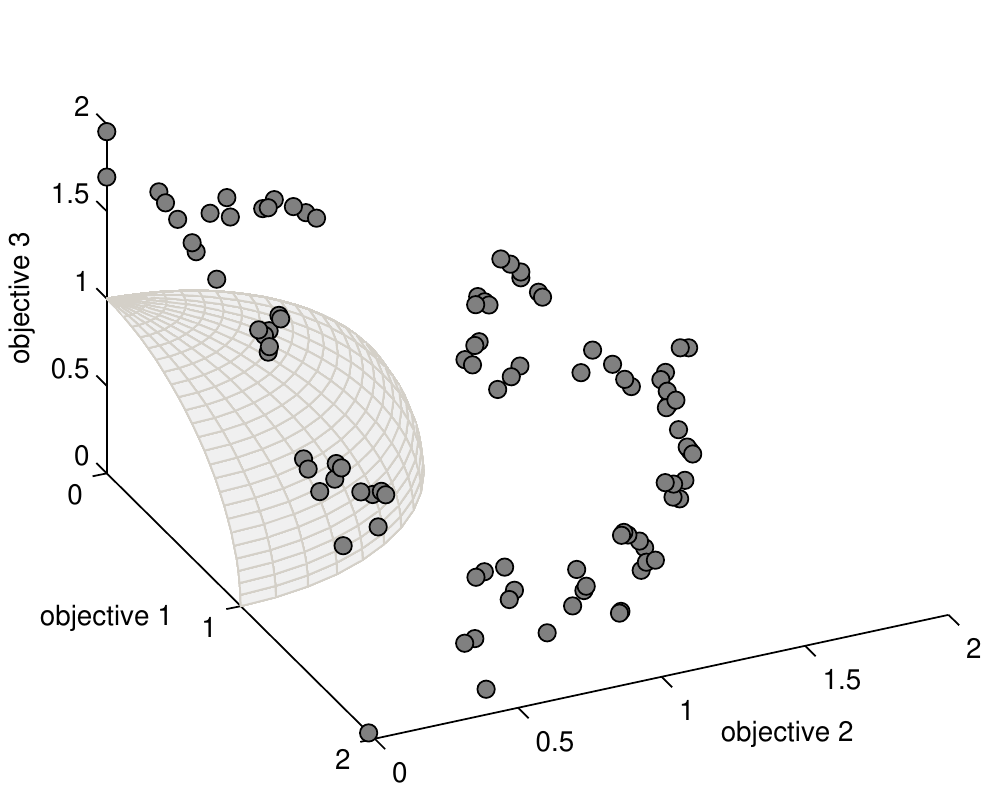}}%
\caption[Obtained PF for DTLZ2]{Obtained PF for DTLZ2 with $n=40$
\label{fig:EA_poly_DTLZ2_PF}}
\end{figure}

As shown in
Figure~\ref{fig:EA_poly_DTLZ2_PF_2front},\subref*{fig:EA_poly_DTLZ2_PF_2side},
the 2-ploids algorithm converged well to the \ac{PF} of the 3
objectives problem, while \ac{NSGA-II} stood at a relatively larger
distance from that front as shown in
Figure~\ref{fig:EA_poly_DTLZ2_PF_NSGAII}.

The relative performance of the Polyploid algorithms in the DTLZ2
test problem is similar to that of the DTLZ1 test problem. Regarding
diversity, increasing the ploidy number slightly affects the
diversity of the Polyploid algorithms. The maximum change in
diversity values is a decrease in the diversity of the 2-ploids
algorithm by 0.0756, while the other Polyploid algorithms almost
maintained a constant diversity value. Regarding convergence,
increasing the ploidy number always slows down convergence, except
for the 7 and 10-ploids algorithms in the 10 objectives problem as
pointed out earlier.

To analyze the performance of the redundant chromosomes the
following experiments are conducted. A new population is created by
extracting all chromosomes in each solution vector in the Polyploid
algorithms. This new population has a size of $N \times d$ , where
$N$ is the original population size and $d$ is the ploidy number.
Then the average distance of the new population to the \ac{PF} is
calculated, and the percentage of the dominated solutions in the new
population is evaluated. For each of the Polyploid algorithms in
Table~\ref{tb:EA_poly_ext_perf}, the first row shows the average
distance of the original population to the \ac{PF}. The second row
shows the average distance of the new population to the \ac{PF}. The
third row shows the percentage of dominated solutions in the new
population (using Pareto dominance).

\begin{table}
\centering%
\caption{Performance of the extracted population
\label{tb:EA_poly_ext_perf}}%
\begin{tabular}{l|l||c|c|c|c}
\hhline{--||----}%
            &distance       &\multicolumn{4}{c}{objectives}\\%
\hhline{~|~||----}%
algorithm   &to front       &3      &4      &6      &10\\%
\hhline{=:=::====}%
            &original pop.  &0.0515 &0.1139 &0.4381 &0.9738\\%
2-ploids    &new pop.       &0.0518 &0.1145 &0.4428 &0.9929\\%
            &\%dominated    &7.52   &6.25   &4.89   &2.11\\%
\hhline{-|-||----}%
            &original pop.  &0.0871 &0.1469 &0.7348 &1.3115\\%
4-ploids    &new pop.       &0.0890 &0.1499 &0.7528 &1.3463\\%
            &\%dominated   &19.64  &13.57  &10.31  &3.105\\%
\hhline{-|-||----}%
            &original pop.  &0.1402 &0.2692 &0.9319 &1.6890\\%
7-ploids    &new pop.       &0.1484 &0.2785 &0.9919 &1.7293\\%
            &\%dominated    &40.17  &20.55  &13.43  &3.26\\%
\hhline{-|-||----}%
            &original pop.  &0.3402 &0.3358 &1.2231 &1.4619\\%
10-ploids   &new pop.       &0.5006 &0.3685 &1.3160 &1.4983\\%
            &\%dominated    &70.36  &36.67  &18.52  &5.08\\%
\hhline{-|-||----}%
\end{tabular}
\end{table}

As shown Table~\ref{tb:EA_poly_ext_perf}, the average distance of
the new population is slightly worse than that of the original
population. In the case of 2-ploids algorithm with 3 objectives, the
average distance is 0.0515 for the original population and it is
0.0518 for the new population. But in the case of 10-ploids
algorithm with 3 objectives, the average distance deteriorates from
0.34 to 0.5. Note that the percentage of dominated solutions is low
in the case of 2-ploids algorithm with a maximum value of 7.52\% for
the 3 objectives case. This value is steadily increasing with
increasing the ploids number in all objectives cases reaching
70.36\% dominated solutions for the 10- ploids algorithm with 3
objectives.

The new population which offers $N(d -1)$ more solutions may be used
instead of the original population for the 2-ploids algorithm giving
the decision maker more choices. It can be used in problems with
high cost of function evaluations, as the $N(d -1)$ extra solutions
are produced without any extra computational cost.

\subsection{DTLZ3}
The DTLZ3 problem is given as follows:
\begin{gather}
\left.
\begin{aligned}
\text{Minimize}&    &&f_1(\v{x})=(1 + g(\v{x}_M)) \cos(x_1\frac{\pi}{2}) \cos(x_2\frac{\pi}{2}) \dotsm \cos(x_{M-2}\frac{\pi}{2}) \cos(x_{M - 1}\frac{\pi}{2}),\\%
\text{Minimize}&    &&f_2(\v{x})=(1 + g(\v{x}_M)) \cos(x_1\frac{\pi}{2}) \cos(x_2\frac{\pi}{2}) \dotsm \cos(x_{M-2}\frac{\pi}{2}) \sin(x_{M - 1}\frac{\pi}{2}),\\%
\text{Minimize}&    &&f_3(\v{x})=(1 + g(\v{x}_M)) \cos(x_1\frac{\pi}{2}) \cos(x_2\frac{\pi}{2}) \dotsm \sin(x_{M-2}\frac{\pi}{2}),\\%
\vdots&             &&\vdots\\%
\text{Minimize}&    &&f_{M-1}(\v{x})=(1 + g(\v{x}_M)) \cos(x_1\frac{\pi}{2}) \sin(x_2\frac{\pi}{2}),\\%
\text{Minimize}&    &&f_{M}(\v{x})=(1 + g(\v{x}_M)) \sin(x_1\frac{\pi}{2}),\\%
\text{subject to}&  &&0 \leq x_i \leq 1, \quad \text{for} \; i =
1,2, \dotsc, n. \label{eq:EA_poly_DTLZ3a}
\end{aligned}
\right\}%
\\ g(\v{x}_M) = 100 \left[ |\v{x}_M| + \sum_{x_i \in \v{x}_M} (x_i -
0.5)^2 - \cos(20\pi(x_i - 0.5)) \right]. \label{eq:EA_poly_DTLZ3b}
\end{gather}
Where $\v{x}$ and $x_i$ are a decision vector and a decision
variable, respectively. The function $g(\v{x}_M)$ requires
$|\v{x}_M|=k$ variables, and the total number of variables is $n = k
+ M - 1$, where $M$ is the number of objectives.

This problem combines some of the properties of DTLZ1 and DTLZ2. It
has a spherical \ac{PF} like DTLZ2, and a huge number of local
optima like DTLZ1. The \ac{PF} is achieved at $x_i^*=0.5$ leading to
$g(\v{x}_M^*)=0$. For this problem, the number of decision variables
is set to $n=30$ and the optimizers were allowed to go for 50,000
function evaluations.

\begin{figure}
\centering%
\subfloat[3 objectives
\label{fig:EA_poly_DTLZ3_conv_3}]{\includegraphics[width =
0.48\textwidth]{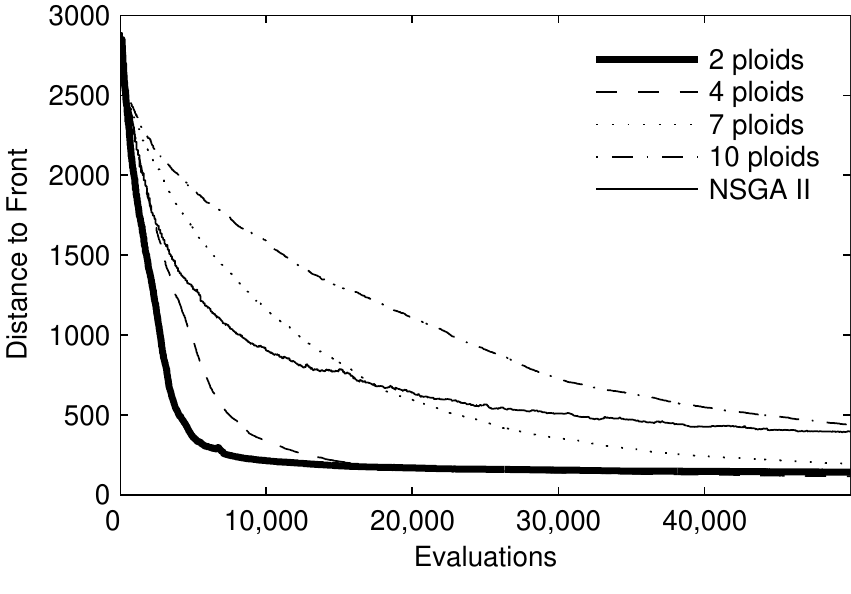}}%
\hfill%
\subfloat[4 objectives
\label{fig:EA_poly_DTLZ3_conv_4}]{\includegraphics[width =
0.48\textwidth]{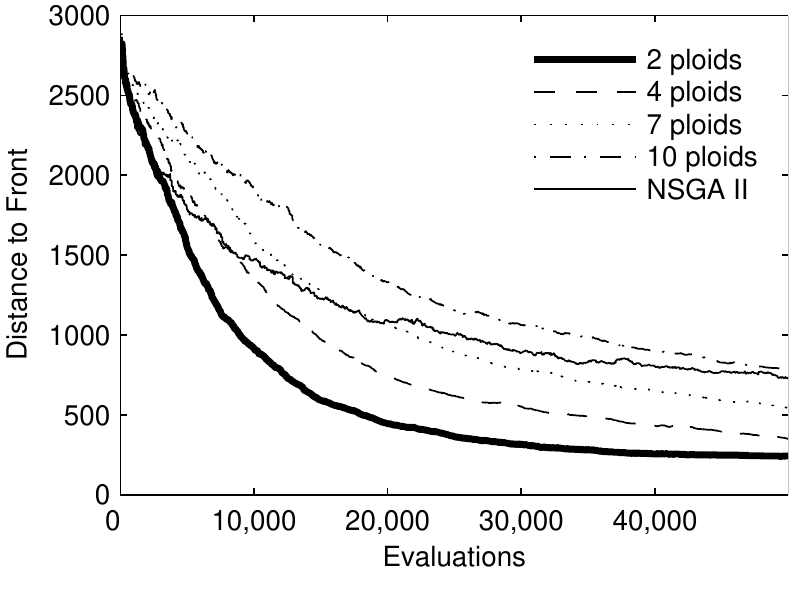}}%
\caption[Convergence speed for DTLZ3]{Convergence speed for DTLZ3
\label{fig:EA_poly_DTLZ3_conv}}
\end{figure}

As shown in Figure~\ref{fig:EA_poly_DTLZ3_conv}, \ac{NSGA-II} is
converging well in early evaluations, and is overcome by the
Polyploid algorithms one after the other, except the 10-ploids
algorithm that gets very close to it after 50,000 evaluations. The
2-ploids algorithm is the fastest converging algorithm in all
problems except for the 3 objectives problem where the 4-ploids
algorithm catches it after around 20,000 evaluations.

Regarding diversity, as shown in Table~\ref{tb:EA_DTLZ3_div},
\ac{NSGA-II} achieves the best values for the 4, 6, and 10
objectives problems. It has a value of 0.7399 in the 6 objectives
problem followed by the 10-ploids algorithm with a value of 0.5994.
However \ac{NSGA-II} has the worst diversity for the 3 objectives
problem with a value of 0.4665, and the second worst is the
10-ploids algorithm with 0.5144 diversity value.

\begin{table}
\centering%
\caption[Diversity values for DTLZ3]{Diversity after 50,000 function
evaluations for DTLZ3
\label{tb:EA_DTLZ3_div}}%
\begin{tabular}{l|l||c|c|c|c}
\cline{3-6}%
\multicolumn{2}{l}{}    &\multicolumn{4}{|c}{objectives}\\%
\hhline{--||----}%
algorithm   &measure    &3      &4      &6      &10\\%
\hhline{==::====}%
\raisebox{-1.5ex}[0cm][0cm]{2-ploids}%
            &Average    &0.6158 &0.5944 &0.4768 &0.6279\\%
            &Std. Dev.  &0.0004 &0.0005 &0.0004 &0.0010\\%
\hhline{--||----}%
\raisebox{-1.5ex}[0cm][0cm]{4-ploids}%
            &Average    &{\bf 0.6380} &0.5504 &0.5312 &0.6736\\%
            &Std. Dev.  &0.0005 &0.0004 &0.0005 &0.0008\\%
\hhline{--||----}%
\raisebox{-1.5ex}[0cm][0cm]{7-ploids}%
            &Average    &0.5753 &0.5419 &0.5554 &0.7059\\%
            &Std. Dev.  &0.0008 &0.0004 &0.0006 &0.0004\\%
\hhline{--||----}%
\raisebox{-1.5ex}[0cm][0cm]{10-ploids}%
            &Average    &0.5144 &0.5940 &0.5994 &0.6784\\%
            &Std. Dev.  &0.0003 &0.0003 &0.0006 &0.0007\\%
\hhline{--||----}%
\raisebox{-1.5ex}[0cm][0cm]{NSGA-II}%
            &Average    &0.4665 &{\bf 0.6217} &{\bf 0.7399} &{\bf 0.7271}\\%
            &Std. Dev.  &0.0022 &0.0002 &0.0004 &0.0002\\%
\hhline{--||----}%
\end{tabular}
\end{table}

As shown in Table~\ref{tb:EA_DTLZ3_conv}, the 2-ploids algorithm,
again, achieves the best convergence values in the 4, 6 and 10
objectives problems. The 4-ploids algorithm is the best for the 3
objectives problem as it reaches a distance to the \ac{PF} of 115.5
followed by the 2-ploids algorithm with a distance of 141.1. The
effect of increasing the ploidy number in this test problem is
little higher than DTLZ1 and DTLZ2. Regarding diversity, the
performance of the 3 objectives problem decreases by 0.1014 when the
ploidy number increases from 2 to 10, while the performance of the 4
objectives problem remained the same and the 6 and 10 objectives
problems have an increase of 0.1226 and 0.0505, respectively, for
the same increase in ploidy number. Regarding convergence,
increasing the ploidy number slows down convergence in all problems
except for the 2 and 4-ploids algorithms in the 3 objectives problem
as pointed out earlier.

\begin{table}
\centering%
\caption[Convergence values for DTLZ3]{Convergence after 50,000
function evaluations for DTLZ3
\label{tb:EA_DTLZ3_conv}}%
\begin{tabular}{l|l||c|c|c|c}
\cline{3-6}%
\multicolumn{2}{l}{}    &\multicolumn{4}{|c}{objectives}\\%
\hhline{--||----}%
algorithm   &measure    &3      &4      &6      &10\\%
\hhline{==::====}%
\raisebox{-1.5ex}[0cm][0cm]{2-ploids}%
            &Average    &141.10 &{\bf 242.48} &{\bf 944.94} &{\bf 2363.3}\\%
            &Std. Dev.  &0.3452 &0.2461 &1.4432 &1.7410\\%
\hhline{--||----}%
\raisebox{-1.5ex}[0cm][0cm]{4-ploids}%
            &Average    &{\bf 115.50} &350.82 &1137.7 &2426.3\\%
            &Std. Dev.  &0.3571 &0.4869 &2.4923 &1.0542\\%
\hhline{--||----}%
\raisebox{-1.5ex}[0cm][0cm]{7-ploids}%
            &Average    &191.24 &546.10 &1499.5 &2393.6\\%
            &Std. Dev.  &0.5813 &0.6889 &1.7420 &1.1733\\%
\hhline{--||----}%
\raisebox{-1.5ex}[0cm][0cm]{10-ploids}%
            &Average    &436.78 &784.23 &1669.1 &2490.1\\%
            &Std. Dev.  &0.8684 &0.9927 &0.8052 &1.4455\\%
\hhline{--||----}%
\raisebox{-1.5ex}[0cm][0cm]{NSGA-II}%
            &Average    &396.20 &712.24 &2328.5 &3124.5\\%
            &Std. Dev.  &0.6275 &0.5085 &1.4124 &0.9903\\%
\hhline{--||----}%
\end{tabular}
\end{table}

\subsection{DTLZ4}
The DTLZ4 test problem is defined as follows:
\begin{gather}
\left.
\begin{aligned}
\text{Minimize}&    &&f_1(\v{x})=(1 + g(\v{x}_M)) \cos(x_1^\alpha\frac{\pi}{2}) \cos(x_2^\alpha\frac{\pi}{2}) \dotsm \cos(x_{M-2}^\alpha\frac{\pi}{2}) \cos(x_{M - 1}^\alpha\frac{\pi}{2}),\\%
\text{Minimize}&    &&f_2(\v{x})=(1 + g(\v{x}_M)) \cos(x_1^\alpha\frac{\pi}{2}) \cos(x_2^\alpha\frac{\pi}{2}) \dotsm \cos(x_{M-2}^\alpha\frac{\pi}{2}) \sin(x_{M - 1}^\alpha\frac{\pi}{2}),\\%
\text{Minimize}&    &&f_3(\v{x})=(1 + g(\v{x}_M)) \cos(x_1^\alpha\frac{\pi}{2}) \cos(x_2^\alpha\frac{\pi}{2}) \dotsm \sin(x_{M-2}^\alpha\frac{\pi}{2}),\\%
\vdots&             &&\vdots\\%
\text{Minimize}&    &&f_{M-1}(\v{x})=(1 + g(\v{x}_M)) \cos(x_1^\alpha\frac{\pi}{2}) \sin(x_2^\alpha\frac{\pi}{2}),\\%
\text{Minimize}&    &&f_{M}(\v{x})=(1 + g(\v{x}_M)) \sin(x_1^\alpha\frac{\pi}{2}),\\%
\text{subject to}&  &&0 \leq x_i \leq 1, \quad \text{for} \; i =
1,2, \dotsc, n.
\end{aligned}
\right\}%
\\ g(\v{x}_M) = \sum_{x_i \in \v{x}_M} (x_i - 0.5)^2.
\end{gather}
Where $\v{x}$ and $x_i$ are a decision vector and a decision
variable, respectively. The function $g(\v{x}_M)$ requires
$|\v{x}_M|=k$ variables, and the total number of variables is $n = k
+ M - 1$, where $M$ is the number of objectives.

The \ac{PF} for this problem is attained when $x_i^*=0.5$, leading
to $g(\v{x}_M)=0$. This problem is a modified version of DTLZ2 with
a different meta-variable mapping ($x \to x^\alpha$). This mapping
allows a dense set of solutions near the $f_M - f_1$ planes
\cite{Deb02a:DTLZs}. This biased distribution of solutions attracts
the algorithms to produce more solutions in the $f_M - f_1$ planes
and makes it difficult for them to maintain a good distribution. For
this problem, then number of decision variables is set to $n=30$,
while $\alpha=100$ as suggested in \cite{Deb02a:DTLZs}.

\begin{figure}
\centering%
\subfloat[6 objectives
\label{fig:EA_poly_DTLZ4_conv_6}]{\includegraphics[width =
0.48\textwidth]{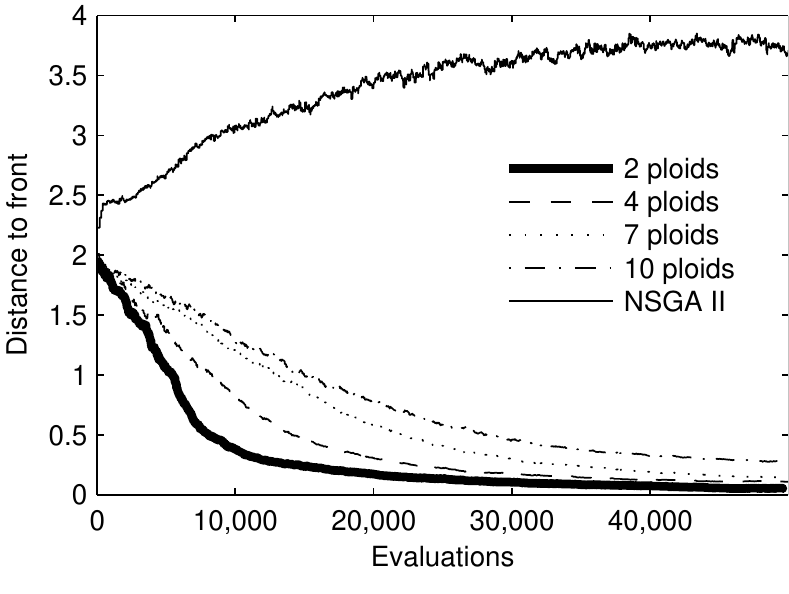}}%
\hfill%
\subfloat[10 objectives
\label{fig:EA_poly_DTLZ4_conv_10}]{\includegraphics[width =
0.48\textwidth]{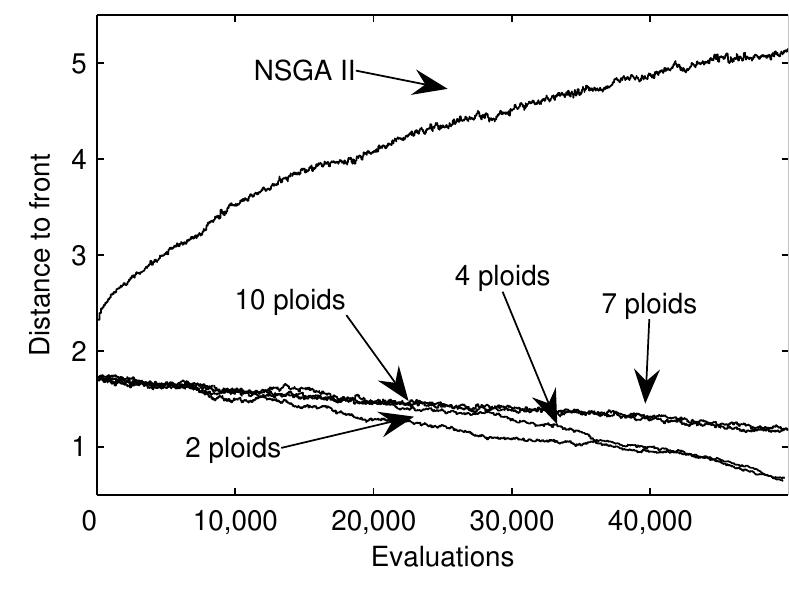}}%
\caption[Convergence speed for DTLZ4]{Convergence speed for DTLZ4
\label{fig:EA_poly_DTLZ4_conv}}
\end{figure}

Figure~\ref{fig:EA_poly_DTLZ4_conv} shows the convergence of the
algorithms in the 6 and 10 objectives problems. The 2-ploids
algorithm is the fastest converging algorithm, again, and is
followed by the 4-ploids algorithm, while the \ac{NSGA-II} is the
slowest converging algorithm and diverges in the 6 and 10 objectives
problems. As shown in Table~\ref{tb:EA_DTLZ4_div}, \ac{NSGA-II}
achieves the best diversity values for the 3, 6 and 10 objectives
problems, and comes third in the 4 objectives problems after the 10
and 7-ploids algorithms, respectively. \ac{NSGA-II} achieves much
better diversity value in the 10 objectives problem than any of the
Polyploid algorithms. It achieves a diversity value of 0.7668
followed by a diversity value of 0.2763 for the 10-ploids algorithm.

\begin{table}
\centering%
\caption[Diversity values for DTLZ4]{Diversity after 50,000 function
evaluations for DTLZ4
\label{tb:EA_DTLZ4_div}}%
\begin{tabular}{l|l||c|c|c|c}
\cline{3-6}%
\multicolumn{2}{l}{}    &\multicolumn{4}{|c}{objectives}\\%
\hhline{--||----}%
algorithm   &measure    &3      &4      &6      &10\\%
\hhline{==::====}%
\raisebox{-1.5ex}[0cm][0cm]{2-ploids}%
            &Average    &0.5956 &0.6101 &0.4946 &0.1754\\%
            &Std. Dev.  &0.0010 &0.0003 &0.0005 &0.0005\\%
\hhline{--||----}%
\raisebox{-1.5ex}[0cm][0cm]{4-ploids}%
            &Average    &0.6517 &0.6487 &0.5634 &0.1866\\%
            &Std. Dev.  &0.0002 &0.0008 &0.0002 &0.0005\\%
\hhline{--||----}%
\raisebox{-1.5ex}[0cm][0cm]{7-ploids}%
            &Average    &0.6634 &0.6926 &0.5746 &0.2618\\%
            &Std. Dev.  &0.0002 &0.0002 &0.0002 &0.0004\\%
\hhline{--||----}%
\raisebox{-1.5ex}[0cm][0cm]{10-ploids}%
            &Average    &0.6711 &{\bf 0.6986} &0.6087 &0.2763\\%
            &Std. Dev.  &0.0003 &0.0002 &0.0002 &0.0004\\%
\hhline{--||----}%
\raisebox{-1.5ex}[0cm][0cm]{NSGA-II}%
            &Average    &{\bf 0.7320} &0.6870 &{\bf 0.6970} &{\bf 0.7668}\\%
            &Std. Dev.  &0.0003 &0.0003 &0.0002 &0.0002\\%
\hhline{--||----}%
\end{tabular}
\end{table}

The low diversity values for the Polyploid algorithms in the 10
objectives problem compared to that of \ac{NSGA-II} reflects its
failure to maintain a good diversity of solutions in problems with
non-uniform distribution of solutions along the \ac{PF}, coupled
with a high number of objectives (10 objectives).

Table~\ref{tb:EA_DTLZ4_conv} shows that all Polyploid algorithms
have better convergence values than \ac{NSGA-II}, except for the
case of 10-ploids with 3 objectives case. In this case the 10-ploids
reaches a distance of 0.2239 to the \ac{PF} compared to a value of
0.1759 for \ac{NSGA-II}. The 2-ploids algorithm achieves the best
convergence values at the end of the 50,000 function evaluations
except for the 4 objectives problem where it achieves a value of
0.0587 and comes second to the 4-ploids algorithm which reaches a
distance of 0.0547 to the \ac{PF}.

\begin{table}
\centering%
\caption[Convergence values for DTLZ4]{Convergence after 50,000
function evaluations for DTLZ4
\label{tb:EA_DTLZ4_conv}}%
\begin{tabular}{l|l||c|c|c|c}
\cline{3-6}%
\multicolumn{2}{l}{}    &\multicolumn{4}{|c}{objectives}\\%
\hhline{--||----}%
algorithm   &measure    &3      &4      &6      &10\\%
\hhline{==::====}%
\raisebox{-1.5ex}[0cm][0cm]{2-ploids}%
            &Average    &{\bf 0.0249} &0.0587 &{\bf 0.0529} &{\bf 0.6421}\\%
            &Std. Dev.  &0.0001 &0.0002 &0.0001 &0.0032\\%
\hhline{--||----}%
\raisebox{-1.5ex}[0cm][0cm]{4-ploids}%
            &Average    &0.0359 &{\bf 0.0547} &0.1069 &0.6778\\%
            &Std. Dev.  &0.0001 &0.0001 &0.0003 &0.0027\\%
\hhline{--||----}%
\raisebox{-1.5ex}[0cm][0cm]{7-ploids}%
            &Average    &0.0748 &0.0967 &0.1431 &1.1748\\%
            &Std. Dev.  &0.0002 &0.0003 &0.0003 &0.0012\\%
\hhline{--||----}%
\raisebox{-1.5ex}[0cm][0cm]{10-ploids}%
            &Average    &0.2239 &0.1241 &0.2654 &1.1822\\%
            &Std. Dev.  &0.0008 &0.0003 &0.0008 &0.0017\\%
\hhline{--||----}%
\raisebox{-1.5ex}[0cm][0cm]{NSGA-II}%
            &Average    &0.1759 &0.7957 &3.7163 &5.1274\\%
            &Std. Dev.  &0.0011 &0.0032 &0.0029 &0.0029\\%
\hhline{--||----}%
\end{tabular}
\end{table}

Diversity values are steadily, though slightly, increasing in all
problems with increasing the ploidy number. The convergence
performance on the other hand is negatively affected by increasing
the ploidy number. The average distance to the \ac{PF} is steadily
increasing with increasing the ploidy number except for the 2 and
4-ploids cases in the 4 objectives problem as pointed out earlier.

\subsection{Conclusion}
\label{ch:EA_sec:Poly_subsec:conc} The benchmark problems used
tested the performance of the Polyploid algorithms and the
\ac{NSGA-II} algorithm regarding convergence to the \ac{PF} and the
diversity of the obtained solutions. The first three test problems
(DTLZ1--3), which emphasize convergence, showed the ability of the
Polyploid algorithms to converge faster than \ac{NSGA-II} to the
\ac{PF}, however, the convergence speed decreased as the ploidy
number increased. On the other hand, an increase in the ploidy
number resulted in a slight improvement regarding diversity of
solutions in problems with higher number of objectives (6, 10
objectives). This slight improvement almost vanished for problems
with lower number of objectives (3, 4 objectives). The fourth test
problem (DTLZ4) tested the ability of the algorithms to maintain a
good distribution of solutions across the \ac{PF}. The Polyploid
algorithms maintained a reasonable degree of diversity but lower
than that of the \ac{NSGA-II} algorithm in the 3, 4, and 6
objectives problems. However, they failed to maintain a satisfactory
degree of diversity in the 10 objectives problem compared to the
diversity obtained by the \ac{NSGA-II} algorithm.

Although an increase in the ploidy number generally resulted in a
slight improvement in diversity values, this marginal benefit was
accompanied by a decrease in convergence speed overshadowing the
diversity enhancement. Based on the obtained results, a ploidy
number between 2 and 4 is recommended for obtaining a good
convergence speed while not sacrificing diversity.

%% file: SI/SI.tex
\chapter{Swarm Intelligence Methods}
\label{ch:SI}

\section{Introduction}
Not long time ago, man dreamed of a computing machine that could do
his time consuming and tiresome mathematical calculations. Although
man did not know how exactly this machine will look like or operate,
he expected that this machine will be like \emph{human brain},
capable of reasoning and solving problems just like humans do, or
even better.

When the first models of this computing machine started to appear,
researchers and philosophers started arguing about the effects that
these new computing machines or computers will have on mankind. Some
of them were doubtful about their widespread\footnote{``\emph{I
think there is a world market for maybe five computers},'' Thomas
Watson, President of International Business Machines (IBM), 1943.},
and usability\footnote{``\emph{There is no reason anyone would want
a computer in their home},'' Ken Olson, President, Chairman and
Founder of Digital Equipment Co., 1977.}, while others anticipated
that these infallible machines will control the human race. It did
not take much time until scientists and researchers realized the big
differences between the human brain and these computers. Those
computers cannot recognize a face or understand spoken language,
although these are easy tasks for a four years old child.

Some researchers believed that the best way to resemble human
ability of solving problems is to create a model of his brain and
use it in solving problems. They created \acp{ANN}, which are loose
models of the cortical structures of the brain. \acp{ANN} were
relatively successful in some applications such as pattern
recognition compared to other techniques preceded it, but yet failed
to achieve the feats that human brain can do, because they are not
\emph{intelligent}. But what is \emph{intelligence}?

There is no common definition for intelligence or what are the
properties of an intelligent entity. Sometimes it is defined using a
set of qualities such as the verbal, analytical, problem-solving and
reasoning abilities, among others, while according to Edwin G.
Boring: intelligence is whatever it is that an intelligence test
measures. Alan Turing defined intelligence as the ability to pass a
test he proposed \cite{Turing50}. In this test, a judge sits in a
room and makes a conversation in writing with a man and a machine in
another room without seeing or hearing them. If the judge cannot
identify or wrongfully identifies the man or the machine, then this
machine is considered \emph{intelligent}.

The Turing definition of intelligence is unique in the sense that it
includes a \emph{social} aspect; An intelligent machine should be
able to \emph{understand} and \emph{distinguish} different meanings
of a word according to the context of the conversation. It can
\emph{feel} the tone of the language and \emph{distinguish} a joke
from a serious speech.

Many researchers accepted Turing's definition of intelligence and
started investigating social interactions of different species.
Their views and findings were quite interesting.

Some researchers investigating the social behaviors of bees
concluded that the beehive is a single living creature, just as a
man, and a bee is only a part or an organ of this creature, just as
a nail or an eye to a man. The simple brain of a single bee does not
allow it to build the complex structured beehive, find food, and
protect the hive. But the collective behavior of the entire swarm
manages to do all this. It is to be noted that the collective
behavior of the swarm is not a sum of the parts; rather it is a
behavior that \emph{emerges} due to social interaction between parts
of the system.

The behavior of many insects, birds and fish was fascinating and
inspiring. After noticing that a swarm of ants can find the shortest
route from the nest to a piece of food, researchers created a model
that copies the behavior of this swarm to find the shortest route
for the \ac{TSP} \cite{Dorigo91:ACO}. In another observation,
researchers managed to build a computer paradigm which mimics the
behavior of a flock of birds searching for food
\cite{kennedy:1995:PartSwarmOpt}. This paradigm has strong ability
to find the optimum value of a function just like the ability of a
bird flock in finding food.

\acf{SI} methods are optimization techniques which are based on the
collective behavior in decentralized, self-organized systems,
comprising relatively simple agents equipped with limited
communication, computations and sensing abilities \cite{Abraham06,
Alshuler06, Fleischer:05:SIfoundations}.

James Kennedy and Russell Eberhart proposed extending the
conventional definition of a swarm to include any such loosely
structured collection of agents interacting in a space (not
necessarily a physical space) which may allow the existence of more
than one agent at the same position (such as the cognitive space
where collision is not a concern) \cite{KennedyEberhart:SIbook:01}.

The most widely known and used \ac{SI} methods are \ac{ACO} and
\ac{PSO}, though there are other techniques that fall under the
\ac{SI} umbrella such as the \ac{SDS} method \cite{Bishop:89}.

\section{How and Why They Work?}
\label{ch:SI_sec:HW?}

\ac{SI} methods work differently, so there is no global answer to
the question ``How \ac{SI} methods work?''.

\ac{ACO} works by simulating a colony of ants searching for food.
Each ant leaves a pheromone trail whenever it walks, and this trail
evaporates gradually over time. However, an ant passing over an old
trail will leave its pheromone over that trail, leading to
accumulation of pheromone and consequently a stronger pheromone
trail. If an ant ran into a crossroad, it will choose the one which
has the strongest pheromone trail. This mechanism leads to the
emergent behavior which helps the ant colony find the shortest route
to food.

\begin{figure}
\centering
\includegraphics{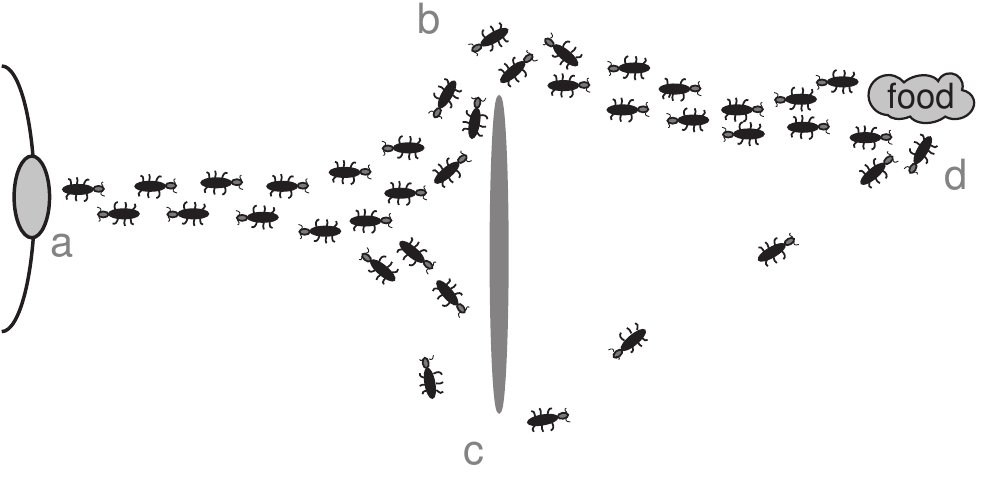}
\caption{Swarm of ants find the shortest route from nest to food
\label{fig:SI_ACO}}
\end{figure}

As shown in Figure~\ref{fig:SI_ACO}, `a-b-d' and `a-c-d' are two
possible routes between the nest and a food source. Ants initially
choose a random route, so it will be equally probably that an ant
chooses any of the two routes. However, the ants taking the first
route `a-b-d' will make more trips between the nest and the food
source in a given time, than those taking the second route `a-c-d'.
Henceforth, the pheromone trail on the first route will gradually
become stronger than that of the second route, and because ants tend
to choose the route with the strongest pheromone trail, the
probability that an ant chooses the first route will increase over
time. As more ants choose the first route, the stronger its
pheromone trail becomes and the more ants will prefer it. On the
other hand, the pheromone trail on the second route gradually
evaporates, and as long as the ants keep switching to the first
route and reinforcing it on the expense of the second one, the
pheromone trail on second route will get weaker and weaker.
Eventually, almost all the ants will go through the first route
which \emph{optimizes} their travel distance.

The \ac{ACO} algorithm can be used to solve the \ac{TSP}
\cite{dorigo97ant}, where the cost of traveling between two cities
is equivalent to part of the tour length the ants take from the nest
to the food source. Analogously, it can be used for scheduling
problems and many combinatorial optimization problems as well
\cite{maniezzo99ant, schoonderwoerd96antbased}.

The way that the \ac{PSO} algorithm works will be explained in
detail in Section~\ref{ch:SI_sec:PSO}.

Despite few theoretical analysis of some \ac{SI} techniques
\cite{Toner98, Tanner03a, Tanner03b, KennedyEberhart:SIbook:01}, the
answer to ``Why \ac{SI} methods work?'' is still unclear. The
complexity of the behavior that emerges from simple social
interactions among swarm members makes the mathematical analysis of
such models quite hard. It was noted that \ac{SI} methods are simple
to implement but are hard to understand \cite{Kennedy05}. This may
explain the tendency of most researchers to conduct empirical rather
than theoretical studies \cite{Sutton06}.

\section{Particle Swarm Optimization}
\label{ch:SI_sec:PSO}

The first models of a flying flock of birds were created for
animation purposes, so it was mainly about high aesthetic animation
rather than solving a problem. Craig Reynolds created a powerful
simulation of flocking birds. His swarm of Boids (artificial birds)
was driven by three simple rules so that each swarm member would
avoid collision, match its velocity with other swarm members, and
move to the swarm center as it perceive it \cite{reynolds87flocks}.
The realistic animation this algorithm produced, which was driven by
three simple rules, encouraged other researchers to follow Reynolds.
Frank Heppner and Ulf Grenander analyzed the films they recorded for
flocking birds and created a powerful computer simulation of
artificial birds. Their rules were similar to those set by Reynolds
though there were some differences.

Two researchers, James Kennedy, a social psychologist, and Russell
Eberhart, an electrical engineer, were inspired by the work of
Reynolds and Heppner \cite{KennedyEberhart:SIbook:01} though they
perceived it differently---influenced by their areas of research.
They created a computer paradigm which simulates a flock of flying
birds. Their paradigm shares the same theme of Reynolds' and
Heppner's work in the sense that there was no central control over
the swarm, however their paradigm was more simple and had different
set of rules. Unlike their predecessors, Kennedy and Eberhart
proposed applications outside computer graphics field. They proposed
using their algorithm in simulating and studying social interactions
of different societies, be it human or animal societies. In the same
paper \cite{kennedy:1995:PartSwarmOpt}, they used their paradigm in
solving an engineering problem. It was used in training an \ac{ANN}.
Because their computer simulations of flocking birds looked more
like \emph{particles} on computer screen than real birds, the
collision avoidance rules were removed, the flock turned into a
\emph{swarm}, and due to the applicability of their algorithm in
\emph{optimization} problems, Kennedy and Eberhart called their
paradigm ``\acl{PSO}''.

After many experiments and modifications, the first \ac{PSO} model
\cite{kennedy:1995:PartSwarmOpt} was built on a population of agents
or particles. Each particle occupies a point in the $n$-dimensional
solution space, which makes it a potential solution vector. The
algorithm initializes by assigning each particle a random position
and velocity vectors. As the particles fly in the solution space,
three forces act upon them. The first one is an inertia which helps
each particle maintain its current direction and velocity. The
second force pushes each particle towards the best position it found
in the past (personal best \emph{pbest}), while the third one pushes
each particle towards the best position found by all the particles
of the swarm (global best \emph{gbest}). It is to be noted that the
second and third forces are directly proportional to the distance
between current position of the particle from one side, and pbest
and gbest positions from the other side, respectively. The effect of
these forces on the velocity of the particles can be described by
the following equation:

\begin{equation}
\label{eq:SI_PSObasic}
\begin{split}
v_{id}(t+1)=&v_{id}(t)+\\
            &2 \times rand_1 \times (p_{id}(t)-x_{id}(t))+\\
            &2 \times rand_2 \times (p_{gd}(t)-x_{id}(t))
\end{split}
\end{equation}
Where
\begin{itemize}
\item[-] $v_{id}(t+1)$ is the $d$ velocity component of particle $i$ at
time $t+1$,

\item[-] $rand_1$ and $rand_2$ are two independent random numbers in the
range $(0,1)$,

\item[-] $p_{id}$ is the $d$ component of the best position found by particle $i$ (pbest position),

\item[-] $p_{gd}$ is the $d$ component of the best position found by all swarm members (gbest
position), and

\item[-] $x_{id}(t)$ is the $d$ position component of particle $i$ at
time $t$.
\end{itemize}

At each time step the position of each particle is updated according
to the following rule:

\begin{equation}
x_{id}(t+1)=x_{id}(t)+v_{id}(t+1)
\end{equation}

As the particles fly in the solution space, they are pushed by their
inertia to explore new regions, while the second and third
components of \eqref{eq:SI_PSObasic} work together to help the
particles converge to the global, or a good local optimum solution.
A weight value of 2 multiplied by a random number with a mean value
of 0.5 means that the particles will overfly the target about half
the time \cite{kennedy:1995:PartSwarmOpt}, and will eventually
converge and land over an optimum solution.

A Matlab code describing a simple \ac{PSO} algorithm is presented
below for a minimization problem:

\begin{lstlisting}[title = A simple \ac{PSO} algorithm (Matlab code)]
initialize(x, v, t);                % initialize parameters;
                                    % x => position,
                                    % v => velocity.
while (termination == 0)
  gbest = inf;                      % gbest => global best.
  for i = 1 : swarm_size
    fitness(i) = evaluate(x(i, :)); % evaluate fitness.
    if (fitness(i) < pbest(i))      % if better than pbest,
      pbest(i) = fitness(i);        % store the new value,
      pl(i, :) = x(i, :);           % and update pbest.
      if (fitness(i) < gbest)       % if better than gbest,
        gbest = fitness(i);         % store the new value,
        pg = i;                     % and update gbest.
      end
    end
  end
  v = v + 2*rand*(x - pl) + 2*rand*(x - pl(pg, :));
                                    % update velocity,
  x = x + v;                        % and position.
  t = t + 1;                        % increment counter.
  termination = term_check(x, v, t);% termination check.
end
\end{lstlisting}

\subsection{Spaces of the Algorithm}
Kennedy has worked out an analysis of the \ac{PSO} algorithm which
was greatly influenced by his social psychological experience. He
used a swarm of humans instead of the systematic use of birds. This
replacement of characters by itself added strength to his argument
for three reasons:
\begin{inparaenum}[i)]
\item it overcomes the conventional depiction of \ac{PSO} models as
a swarm of birds and presents a new example to evoke the imagination
of other researchers.
\item the choice of humans allows him to apply the results obtained
by decades of social psychology research in \ac{AI} context.
\item the examples used were clear because they touch the social
life experience of humans.
\end{inparaenum}
In his analysis, a man is subjected to various stimuli by his
environment from which he learns affected by other individuals
trying to reach a point that achieves their maximum satisfaction.
This analysis was derived based on a three dimensional space; The
parameters space, the sociometric space, and the evaluative space
\cite{Kennedy05}.

\subsubsection{The Parameters Space}
The parameters space simply is the solution space or the decision
space. It is made-up of the problem parameters which when correctly
tuned, the global optimum solution is attained. These parameters can
be perceived as the variables of mind. They determine how the mind
processes information and reacts to different stimuli in its
environment. They are complex and interwoven in the sense that a
change in the value of one variable affects many objectives in most
practical problems. This complexity is analogous to epistasis in
\acp{GA}.

The variables of mind, such as beliefs and norms, may change as man
learns and acquires experience in the course of his life. A man is
influenced by his environment, past experience, and the experience
of his neighbors with whom he interacts, among other influences.

The environment may influence a man by limiting his choices, so he
may not tune his mind variables to his satisfaction, such as an
oppressive society that persecutes people holding certain beliefs.
The environment in this case adds \emph{constraints} on the
\emph{variables}.

Personal experience is very influential in people's life. Man tends
to remember the most successful experiences in his life, and when a
situation repeats, he recalls the action which lead to the most
satisfactory result (according to his standards) for that situation
in the past, and tends to repeat this action or try a similar one.
The action which caused the most satisfactory result is represented
by $p_{id}$ in \eqref{eq:SI_PSObasic}.

Another major source of influence for humans is the experience of
their neighbors. The definition of neighborhood will be explained in
detail in Subsection~\ref{ch:SI_sec:Var_subsec:SocNet}, but for now
it is enough to say that the neighbors of an individual are those
individuals who are at certain degree of closeness to him. A man
tends to imitate or resemble his most successful neighbor, believing
that by taking the same action, he will get a similar successful
result. This imitation may come in handy in many situations;
Scientific research in fact is heavily based on this kind of
imitation, a researcher reads about the work of other researchers
and have discussions with them to learn from their experience. The
action which caused the most successful result among all individuals
in a man's neighborhood is represented by $p_{gd}$ in
\eqref{eq:SI_PSObasic}.

It is to be noted that learning occurs at a slower pace than
perceiving information. Although there is an overwhelming volume of
information that pours into man's mind every day, the states of his
mind do not change in reaction to each piece of this information.
The rate at which he is affected by these stimuli is known as the
\emph{learning rate}, which will be explained in
Subsection~\ref{ch:SI_sec:Var_subsec:LR}.

\subsubsection{The Sociometric Space}
The sociometric space is among the properties which distinguish
\ac{PSO} from \acp{EA}. In this space, the particles' neighborhood
is defined and directly affects the interactions among different
particles and how they will learn from each other. Every man gets
influenced by his neighbors, but practically, this influence depends
on many factors. Kennedy defined three factors affecting this
influence \cite{Kennedy05}:
\begin{enumerate}[i)]
\item \emph{Strength}: is a relative measure of how much a man is
attracted to the neighbor in a certain situation. The strength of a
neighbor could be a measure of his persuasiveness, social status, or
a personal experience with that neighbor. In the \ac{PSO} example
given earlier, strength was measured by the fitness of the particle;
a particle is affected by the most fit particle in its neighborhood.

\item \emph{Immediacy}: is a measure of the degree of closeness to the
neighbor. Closeness should not be associated exclusively with the
physical space or the Euclidian sense. It could be the distance in a
cognitive space: a man is affected by neighbors sharing his beliefs,
or blood bond: a man is affected by his father who lives in another
continent. The degree of closeness could be crisp (neighbor or not a
neighbor), or fuzzy (varying gradually as distance changes). In the
previous \ac{PSO} example, immediacy was crisp and was limited by
the neighborhood topology.

\item \emph{Number}: is the number of neighbors sharing the same
belief. This factor is not applicable in some situations; in real
life, the beliefs and standards of people are spread along wide
spectrum, it is really hard to find two people share the exact
beliefs. Furthermore, if the fuzzy neighborhood definition is
adopted, the classification of people as neighbors and not-neighbors
will not be possible. The `number' factor was not considered in the
previous \ac{PSO} example.
\end{enumerate}

In real-world, influence is not symmetric. An idol has a far
reaching influence on millions, but there is no reciprocal influence
by those millions on that idol. A less drastic example is the mutual
influence between a father and a son. So, the influence has to be
defined in both directions. Moreover, things may get more
complicated when realizing that some influences depend on others.
The influence of a parent on his son could be drastically reduced if
the two parents get separated, noting that the son was directly
connected to both of them. Influence can be mutually interactive;
$a$ trusts $b$ because $b$ trusts him.

The social networks greatly depend on people and environment. A
solitary man is connected to much fewer people than a gregarious
one, and a man living in a city hub will meet more people than a one
living on the fringe of the city.

Neglecting the effect of time is not a wise choice. The social
networks are highly dynamic, even if one tries hard to keep a
connection, it could be broken from the other side, so a shrewd man
may adjust his social network from time to time to his benefit.

The premise that an individual is affected by one neighbor is in
fact not judicious. Although a neighbor could be more influential on
a man than other neighbors, he does not block their influence on
that man. In many situations, a man is affected by the norms and
beliefs of his society which do not stem from a single individual or
a bunch of people, and do not emerge a fortnight.

Although two people may not be directly connected, they could still
influence each other through the neighbors they share, or the
neighbors that their neighbors share. Obviously, this chain can go
on and on, and its shape determines what is known as the \emph{flow
of influence}. The flow of influence determines, among other things,
how fast the influence of a superior man will spread through people
and affects them, and the direction or path of this spread. A good
example is the spread of an epidemic disease, in this case, the
spread of the virus depends on our social networks among other
factors. The spread of influence is a mixed blessing, a rapid spread
of influence means fast convergence, but to a local optimum in most
cases. While a slow spread helps exploring the search space looking
for the global optimum, and meandering in the meanwhile.

\subsubsection{The Evaluative Space}
The evaluative space is the space where all possible reactions to
different stimuli is defined. This reaction is mainly based on the
states of mind and is affected by noise. In the evaluation process,
the input parameters of a $n$-dimensional space are mapped to a
relative evaluation values in a one dimensional space, and based on
their relative position along this dimension a decision is made.
However this mapping could be misleading; such as the drag force of
an object approaching the speed of sound. The drag force acting on
this abject increases as its speed gets closer to the speed of
sound, but once it reaches it, the drag force suddenly collapses.
The mapping could be flat; like searching for the correct numbers
combination to open a lock, there is no indication whatsoever to
guide the search process towards the optimum solution.

Different stimuli are compared relative to each other on the one
dimensional evaluative space. The absolute evaluation value of a
stimulus is worthless, where does a 1000\$ salary stands? It is high
when compared to a 500\$ pay, but a low one when compared to a
5000\$ stipend. Does that make a 5000\$ allowance a good one? Not
when compared to a 7000\$ one.

As time passes by and situations change, the comparison level
changes as well. The comparison level, adaptation level, or anchor
\cite{Kennedy05} is the reference level, which compared to it, the
values of different stimuli are classified as satisfactory and
non-satisfactory. Henceforth, a man who gets the highest salary in
his firm may not be satisfied with it because he compares it to the
higher salary he had at his previous position. But once his salary
surpasses that old one, the adaptation level rises with it. The new
adaptation level is the value of the higher salary, and a lower
salary which was acceptable in the past will no longer become
acceptable.

In the \ac{PSO} example given in Section~\ref{ch:SI_sec:PSO}, the
$n$-dimensional position of each particle (in the parameters space)
was mapped to a one dimensional fitness value (in the evaluative
space). The fitness value of each particle assigns it a position
along the fitness scale to be relatively compared to other
particles. While $p_{id}$ and $p_{gd}$ are the local and global
adaptation levels , respectively, that a particle maintains. In this
\ac{PSO} algorithm, the order of a particle is all that matters; if
a particle has the best local/global fitness value, this value
becomes the local/global best with no regard to its absolute value,
and with no regard to differences between this value and the values
of other particles.

\section{Variations}
\label{ch:SI_sec:var}

The basic \ac{PSO} algorithm which was developed by Kennedy and
Eberhart and was presented in their 1995 paper became obsolete.
Different variations has been made to the algorithm either by those
two researchers or by others when the algorithm was widely accepted
few years after its embarkation. Different rates of learning were
tested and proposed \cite{Ratnaweera04}. The inertia that help the
particles maintain their direction were constricted
\cite{ShiEberhart:PSOpar98}. Different social networks were tested
and suggested \cite{MohaisMWP05, Pasupuleti06, JansonM05,
kennedy:1999:swment}, and extension of the \ac{PSO} algorithm was
made to binary and discrete problems as well \cite{Correa06,
Kennedy05, KennedyEberhart:SIbook:01}.

\subsection{Learning Rates}
\label{ch:SI_sec:Var_subsec:LR}

The learning rates, or acceleration coefficients, are the parameters
that determine the degree to which a particle is affected by its
past experience and the experience of its neighbors. In the basic
\ac{PSO} algorithm, there was two such parameters, however their
were set to a constant value of 2. In modern \ac{PSO}
implementations, these parameters ($\varphi_1$, $\varphi_2$) are
introduced to the velocity update equation
\begin{equation}
\begin{split}
v_{id}(t+1)=&v_{id}(t)+\\
            &\varphi_1 \times rand_1 \times (p_{id}(t)-x_{id}(t))+\\
            &\varphi_2 \times rand_2 \times (p_{gd}(t)-x_{id}(t))
\end{split}
\end{equation}

When $\varphi_1 > \varphi_2$, the particle tends to rely on its past
experience than the experience of its neighbors, and when $\varphi_1
< \varphi_2$, the particle trusts the experience of its neighbors
more than its own experience. Most \ac{PSO} implementations use
equal values for those two parameters ($\varphi_1 = \varphi_2$).
Maurice Clerc and Kennedy suggested setting those two parameters
such that $\varphi_1 + \varphi_2 = 4.1$ \cite{ClercKennedy:2002},
however other researchers reported better results over a wide set of
test function when those two parameters were set at other values
\cite{JansonM05, Elshamy07}.

The summation of $\varphi_1$ and $\varphi_2$ affects the performance
of the algorithm as well. As the value of $\varphi_1 + \varphi_2$
increases, the particles increase their steering degree and become
relatively less affected by the inertia force. The situation is like
driving a moving car, $v_{id}$ is the velocity vector of the car,
and $\varphi_1$, $\varphi_2$ are the degrees to which the driver
steers the wheel to hit his target. As $\varphi_1$ and $\varphi_2$
increases, the driver steers his wheel more sharply towards pbest
and gbest points, respectively, whenever he sees them, and
consequently explores smaller portion of the landscape.

\subsection{Constriction}
It was found during early experiments on \ac{PSO} that the velocity
of the particles explode and approach infinity \cite{Kennedy05}. So,
Kennedy and Eberhart set a maximum velocity limit on the movements
of the particles $v_{max}$ to prevent this behavior, without
understanding its causes. Eberhart and Shi
\cite{Eberhart98:comparison} introduced an inertia weight to the
algorithm ($w$). The velocity update equation became:
\begin{equation}
\begin{split}
v_{id}(t+1)=&w\times v_{id}(t)+\\
            &\varphi_1 \times rand_1 \times (p_{id}(t)-x_{id}(t))+\\
            &\varphi_2 \times rand_2 \times (p_{gd}(t)-x_{id}(t))
\end{split}
\end{equation}

The weight $w$ is typically below the unity value, to act like
damper for the velocity of the particle. When $w$ is close to a
value of 1, the particles swing and meander in the search space, and
have a high exploration power. While decreasing the value of $w$,
inhibits the velocity of the particles and allows them to converge
faster. They suggested starting the algorithm by a weight value of
0.9 to scout out the landscape, then reducing this value to exploit
the obtained knowledge as the algorithm proceeds, till it reaches
0.4 at the end of the run.

Clerc and Kennedy worked out a mathematical analysis of the \ac{PSO}
algorithm in their award winning paper \cite{ClercKennedy:2002}.
They explained the reasons which lead to explosion of the velocity
of the particles and analyzed the particle's trajectories as it
moves in discrete time and developed a generalized model in a
five-dimensional complex space with a set of coefficients to control
the convergence of the algorithm. They suggested a constriction
coefficient ($\chi$) to wight the entire right-hand side of the
velocity update equation. It looked like
\begin{equation}
\begin{split}
v_{id}(t+1)=&\chi \times (v_{id}(t)+\\
            &\varphi_1 \times rand_1 \times (p_{id}(t)-x_{id}(t))+\\
            &\varphi_2 \times rand_2 \times (p_{gd}(t)-x_{id}(t)))
\end{split}
\end{equation}

The value of $\chi$ is suggested to be approximately 0.729, and
$\varphi_1 + \varphi_2 = 4.1$.

\subsection{Social Networks}
\label{ch:SI_sec:Var_subsec:SocNet}

The social network of the \ac{PSO} algorithm is what mainly
distinguishes it from other \ac{SI} techniques, and more generally,
from other \ac{CI} methods. The basic \ac{PSO} algorithm which was
presented in Section~\ref{ch:SI_sec:PSO} used a fully connected
social network; every particle is connected to all other particles
as shown in Figure~\ref{fig:SI_PSO-g}, which means it is aware of
their best fitness value and the position that resulted in this
value. This social network is known as the \emph{gbest} topology,
where `g' stands for `global' because each particle is affected by
the global best position. The gbest topology is known for its rapid
convergence, and susceptibility to local optima as well. As soon as
one particle finds a global best value early in the run, the other
particles hurtle to it, and as long as those particles encounter
improvements in their fitness values while approaching that global
best, they get more strongly sucked towards it, and eventually, they
all converge to this global best which happens to be a local optima
in most practical problems. Henceforth, many other social networks
were proposed to alleviate this shortcoming and to add more strength
to the algorithm.

\subsubsection{lbest}
\begin{figure}
\hfill
\begin{minipage}[b]{0.45\textwidth}%
\centering
\includegraphics[width=5cm]{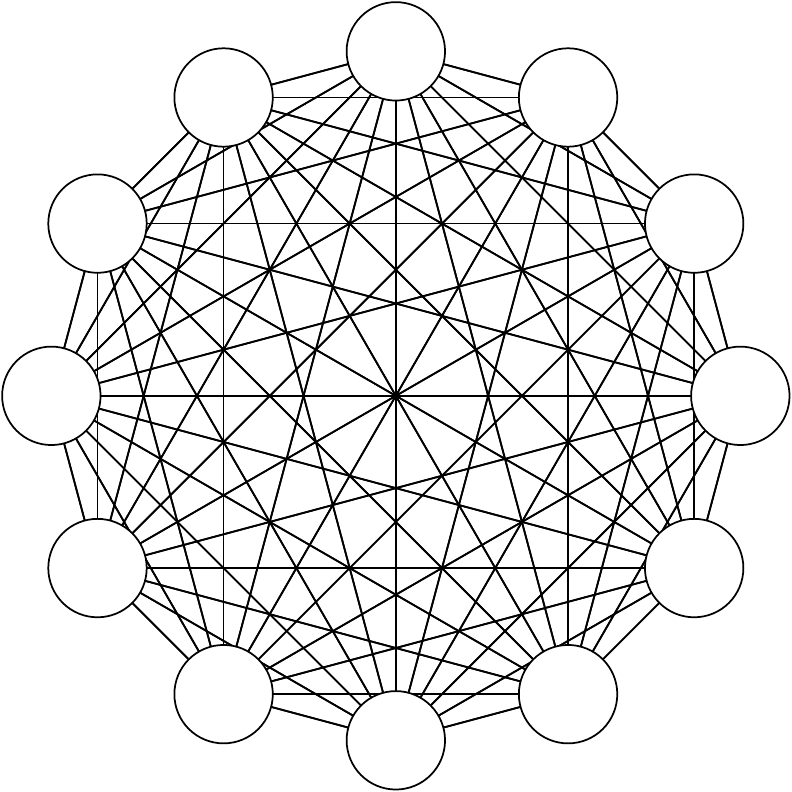}\caption{gbest topology \label{fig:SI_PSO-g}}
\end{minipage}
\hfill
\begin{minipage}[b]{0.45\textwidth}%
\centering
\includegraphics[width=5cm]{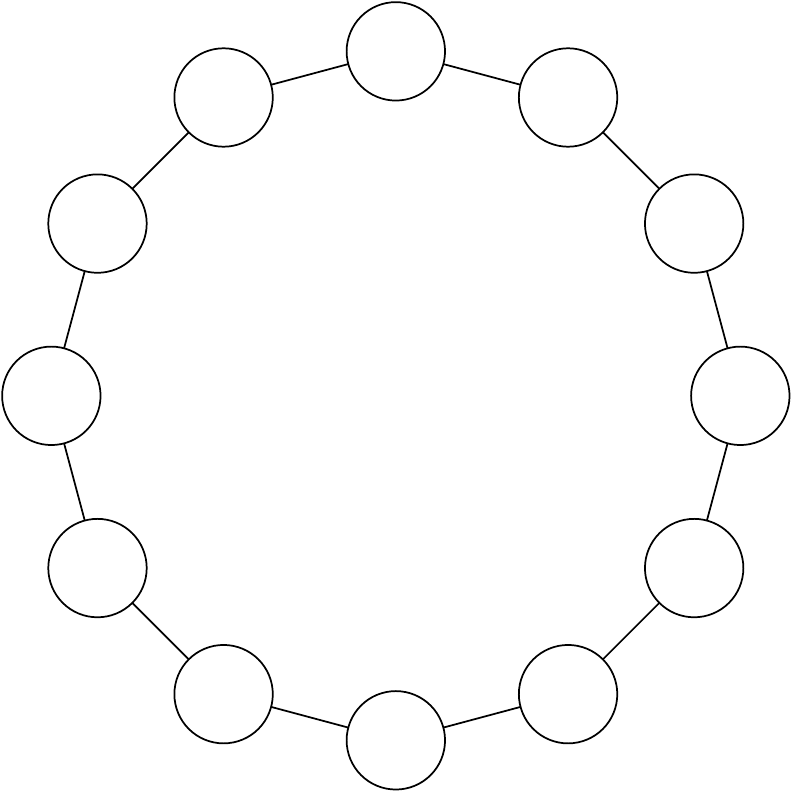}\caption{lbest topology \label{fig:SI_PSO-l}}
\end{minipage}
\hfill
\end{figure}

The lbest topology is one of the earliest topologies used. The `l'
in its name refers to `local', because each particle is only
connected to its `locals'. By other words, if the particles were
arranged in a circle, each one will be connected to the particle
that precedes it, and the one which succeeds it along the circle
path as shown in Figure~\ref{fig:SI_PSO-l}. Note that the
neighborhood in the lbest topology is based on an arbitrary, but
fixed, index values for the particles, and it has nothing to do with
their inter-distances in the decision space or the objective space.

Due to lack of unique position attracting all swarm members, such as
in gbest topology, a swarm connected using an lbest topology will be
affected by different points of attraction. If a particle, or a
group of particles get trapped in a local optima, their detrimental
influence will spread slower than the case of the gbest topology,
and it is likely that their neighbors will find a better value and
pull them out of the trap.

\subsubsection{Hierarchical PSO}
\begin{figure}
\centering
\includegraphics[width=10cm]{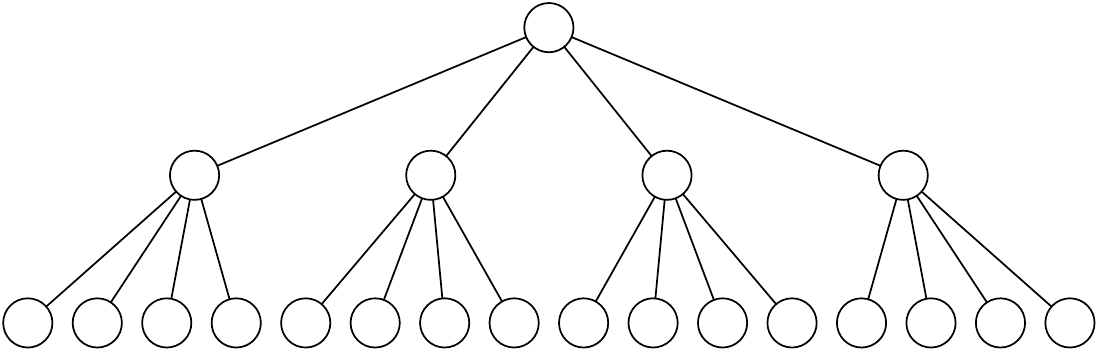}
\caption{Hierarchical PSO \label{fig:SI_H-PSO}}
\end{figure}

\ac{H-PSO} is another \ac{PSO} topology proposed by Stefan Janson
and Martin Middendorf \cite{JansonM05}. By using this topology, the
particles are arranged in a hierarchy structure, and the
neighborhood of each particle is made of the particle itself and its
parent node in the hierarchy. The hierarchy is defined by its
\emph{height `h'}, and its \emph{branching degree `d'}. For example,
Figure~\ref{fig:SI_H-PSO} shows a swarm of 21 particles arranged in
a hierarchy of height $h=3$, and a branching degree of $d=4$.
However, it will not be possible to construct a regular tree for the
particles with a uniform branching degree at all nodes. Henceforth,
any inconsistency will be pushed to the inner nodes at the deepest
level of the tree, and the maximum difference of branching between
any of those irregular nodes will be at most one.

The fitness of each swarm member is evaluated as usual, then
starting from the top of the tree and proceeding to the bottom, the
local best fitness value of each swarm member represented by a
parent node in the tree is compared to that of its child nodes in
the hierarchy. If the parent is less fit than the best one of its
child nodes, they swap their positions in the hierarchy, and if not,
they stay at the same hierarchy position. This procedure pushes the
more fit particles up in the tree, and drags the less fit down
towards the bottom of the tree. As a result, the procedure will
arrange the particles in the tree according to their fitness, with
the most fit at the top of the tree, and the least fit at the
bottom. Since the neighborhood of each particle consists of itself
and its parent node, the more close a particle gets to the top of
the tree, the larger its influence will become. Traversing the tree
using a breadth-first procedure starting from the top of the tree
allows a particle to move down the tree up to $h-1$ levels in a
single iteration of the algorithm if it is worse enough, but limits
the ascending speed to 1 level per iteration. This tree update
procedure is repeated with every iteration of the \ac{PSO}
algorithm.

The \ac{PSO} algorithm will continue as usual by updating the
velocity of the particles using their local best and their
neighbors' best, and then updating their position in the decision
space. The algorithm repeats by evaluating the fitness of these new
positions, then traversing and updating the tree, following by a
velocity and position update\ldots and so on.

The philosophy of the \ac{H-PSO} is almost the opposite of the
\ac{C-PSO}. \ac{H-PSO} promotes the best particles and increases
their influence in the swarm hopefully to speed up convergence,
while the worst particles are dragged to the bottom of the tree to
let them learn directly from particles which are the second worst
after them. On the other hand, \ac{C-PSO} topology reduces the
influence of the best particles by inhibiting their social
interaction with other particles, while allowing the worst particles
to socialize more and increases their chance of learning directly
from the best particles. \ac{H-PSO} and \ac{C-PSO} are similar in
other facets, they both employ a dynamic neighborhood which adapts
to the fitness or the particles, however they adapt differently.

\subsubsection{Fitness Distance Ratio PSO}
\ac{FDR-PSO} is variant of the \ac{PSO} which is based on a social
network unique from those presented earlier. This topology which was
proposed by Kalyan Veeramachaneni et al. \cite{VeeramachaneniPMO03},
was inspired from the observation of animal behavior. An animal is
most likely to be influenced by its close neighbors, and the more
successful are those neighbors, the higher their influence on that
animal becomes. However, Veeramachaneni et al. proposed considering
the influence of one neighbor only to avoid the canceling out of
different forces acting in different directions.

They proposed updating the $n$-components of the particle velocity
vector independently. For the $d^{\text{th}}$ dimension, the
neighbor particle $j$ which maximizes the ratio of the fitness
difference between those two particles to the distance between them
along this dimension is chosen as the influencing neighbor. The
Fitness-Distance-Ratio along the $d^{\text{th}}$ dimension between a
particle $i$ and its neighbor with local best $p_j$ for a
maximization problem can be expressed by:
\begin{equation}
FDR(j,i,d) = \frac{f(p_j) - f(x_i)}{|p_{jd}-x_{id}|}
\end{equation}
where $f$ is the fitness function, and $|\ldots|$ means the absolute
value.

After finding the neighbors who maximize the FDR of particle $i$
along the $d$-dimensions, they are used to update its velocity
components according to the following rule:
\begin{equation}
\label{eq:SI_FDR-PSO}
\begin{split}
v_{id}(t+1)=&w \times (v_{id}(t)+\\
            &\psi_1 \times (p_{id}(t)-x_{id}(t))+\\
            &\psi_2 \times (p_{gd}(t)-x_{id}(t))+\\
            &\psi_3 \times (p_{nd}(t)-x_{id}(t)))
\end{split}
\end{equation}
where $v_{id},w,p_{id},p_{gd}$ are the same as in the basic
\ac{PSO}, $p_{nd}$ is the position of the particle which maximizes
the FDR along the $d^{\text{th}}$ dimension, and $\psi_i$ is a
weighting factor used to change the influence effect of the last
three terms of~\ref{eq:SI_FDR-PSO}. When
$(\psi_1,\psi_2,\psi_3)=(1,1,0)$ the algorithm resembles the basic
\ac{PSO}, and when $\phi_3\neq 0$ the effect of the proposed
procedure starts to appear.

This velocity update mechanism was tested on various test functions
and it outperformed the basic \ac{PSO} even when the original
\ac{PSO} terms where disabled;
($(\psi_1,\psi_2,\psi_3)=(0,0,\psi3)$) \cite{VeeramachaneniPMO03}.

\subsubsection{The Fully Informed Particle Swarm}
Rui Mendes et al. proposed an influence scheme where each particle
is not only affected by the best particle in its neighborhood, it is
affected by all its neighbors \cite{MendesKN04}. This proposition,
however, does not imply any social topology, it can be used with any
one of the previously mentioned social networks and others. The
\ac{FIPS} can be described using the following equations.
\begin{gather}
v_{id}(t+1)=\chi (v_{id}(t) + \varphi (p_{id}(t) - x_{id}(t)))\\
p_{id}=\frac{\sum_{k \in \mathcal{N}}f(k) \, \varphi_{k} \, p_{kd}}{\sum_{k \in \mathcal{N}} f(k) \, \varphi_{k}}\\
\varphi_k = \mathrm{U}\left[0,
\frac{\varphi_{max}}{|\mathcal{N}|}\right] \forall k \in \mathcal{N}
\end{gather}
where $\varphi = \sum_{k \in \mathcal{N}} \varphi_k$,
$\varphi_{max}=4.1$, $\mathcal{N}$ is the set of particle $i$
neighbors, $f(k)$ the fitness of particle $k$, $p_{kd}$ is the $d$
component of the position that resulted in the best fitness value
found by particle $k$, $\mathrm{U}[min,max]$ returns a random number
in the range $[min,max]$ following a uniform distribution.

\subsection{Representations}
First \ac{PSO} models was designed to work on continuous time
problems \cite{kennedy:1995:PartSwarmOpt}. Not all problems can be
solved in a continuous domain. The \ac{SAT} problem and many others
work in a binary space, while the \ac{TSP} is a combinatorial
problem which works on a discrete space. So there is a need for
\ac{PSO} versions which can handle these problems.

\subsubsection{Binary PSO}
Kennedy and Eberhart modified their simple \ac{PSO} to produce its
binary version. The velocity of the particle in the $d^{\text{th}}$
dimension is transformed to a probability threshold in the range
$(0,1)$, and by producing a random number in the same range and
comparing it to the threshold, the bit at this dimension will either
be set to `1' or `0'. The algorithm works according to the following
rules:
\begin{gather}
\begin{split}
v_{id}(t+1)=&w \times v_{id}(t)+\\
            &\varphi_1 \times rand_1 \times (p_{id}(t)-x_{id}(t))+\\
            &\varphi_2 \times rand_2 \times (p_{gd}(t)-x_{id}(t))
\end{split}\\
S(v_{id}) = \frac{1}{1 + \exp(-v_{id})}\\
x_{id}= \left\{
\begin{aligned}
1 \qquad &\text{if} \; S(v_{id}) > rand,\\
0 \qquad &\text{otherwise}.
\end{aligned}\right.
\end{gather}
where $S$ is the function which produces the probability threshold
value of $v$.

\subsubsection{Discrete PSO}
The discrete version of the \ac{PSO} algorithm operate in discrete
space. Unlike their continuous cousins, particles in a discrete
\ac{PSO} move by discrete steps in their $d$-dimensional space. For
example, the particles exploring the solution space of a
combinatorial optimization problem defined in the four discrete
dimensions $d_1 = (a, b, c, d, e, f)$, $d_2 = (1, 2, 3)$, $d_3 =
(cyan, magenta, yellow, black)$, and $d_4 = (north, \- south,\-
east, west)$ may take positions such as $(a, 3, magnets, south)$.
But how the velocity can be defined in this discrete space? Given
three particles $x_1=(c,\- 1,\- cyan,\- east)$, $x_2=(e, 1, magneta,
north)$, and $x_3=(a, 3, yellow, west)$, which particle is closer to
$x_1$,($x_2$ or $x_3$)? Is $x_2$ closer because it shares the second
dimension position $(d_2 = 1)$ with $x_1$? The definition of
distance and velocity in this discrete space must be defined before
answering these questions. According to Maurice Clerc
\cite{Clerc00:discPSO}, the velocity operator is a function which
when applied to a position during one step, gives another position.

\section{Clubs-based PSO}
\label{ch:SI_sec:C-PSO}

First \ac{PSO} models were confined to perceive the swarm as a flock
of birds that fly in the search space. The picture of fly-ing birds
has limited the imagination of researchers somehow for sometime.
Recently, a more broad perception of the swarm as a group of
particles, whether birds, humans, or any socializing group of
particles began to emerge. In the \ac{C-PSO} algorithm, there are
\emph{clubs} for particles analogous to social clubs where people
meet and socialize. In this model, every particle can join more than
one club, and each club can accommodate any number of particles.
Vacant clubs are allowed \cite{Elshamy07}.

After randomly initializing the particles position and velocity in
the initialization range, each particle joins a predefined number of
clubs, which is known as its \emph{default membership level}, and
the choice of these clubs is made random. Then, current values of
particles are evaluated and the best local position for each
particle is updated accordingly. While updating the particles'
velocity, each particle is influenced by its best found position and
the best found position by all its neighbors, where its neighborhood
is the set of all clubs it is a member of. After velocity and
position update, the particles' new positions are evaluated and the
cycle is repeated.

While searching for the global optimum, if a particle shows superior
performance compared to other particles in its neighborhood, the
spread of the strong influence by this particle is reduced by
reducing its membership level and forcing it to leave one club at
random to avoid premature convergence of the swarm. On the other
hand, if a particle shows poor performance, that it was the worst
performing particle in its neighborhood, it joins one more club
selected at random to widen its social network and increase the
chance of learning from better particles. The cycle of joining and
leaving clubs is repeated every time step, so if a particle
continues to show the worst performance in its neighborhood, it will
join more clubs one after the other until it reaches the maximum
allowed membership level. While the one that continues to show
superior performance in every club it is a member of will shrink its
membership level and leave clubs one by one till it reaches the
minimum allowed membership level.

During this cycle of joining and leaving clubs, particles which no
longer show extreme performance in its neighborhood, either by being
the best or the worst, go back gradually to default membership
level. The speed of going back to default membership level is made
slower than that of diverting from it due to extreme performance.
The slower speed of regaining default membership level allows the
particle to linger, and adds some stability and smoothness to the
performance of the algorithm. A check is made every $rr$
(\emph{r}etention \emph{r}ation) iterations to find the particles
that have membership levels above or below the default level, and
take them back one step towards the default membership level if they
do not show extreme performance. The static inertia weight which
controls the inertia of the particle is replaced by a uniformly
distributed random number in the range $(0,w)$. A Matlab code
explaining the algorithm is shown below.
\begin{lstlisting}[title=Clubs-based PSO (Matlab code), mathescape]
[prt,clb,p,rr,w,v,phi1,phi2,min_memb,...   % initialize:
  max_memb,def_memb,iter] =init();         % prt=>particles,
while (term_cond == 0)                     % clb=>clubs.
  f = eval(prt);                           % evaluate fitness.
  p = lbest(prt,f,p);                      % p=>local best
  for (i = 1 : swarm_size)
    g = best(neighbors(i,clb),p);
                            % find particle `i' best neighbor.
    for d = 1 : n           % n=>number of dimensions.
      v(i,d) = w*rand*v(i,d) + ...       % w=>velocity weight.
        phi1*rand*(p(i,d) - prt(i,d)) + ...
        phi2*rand*(p(g,d) - prt(i,d));   % update velocity,
      prt(i,d) = prt(i,d) + v(i,d);      % and position.
    end
  end
  for j = 1 : swarm_size
    if (best(neighbors(j,clb),p) == j)&&... % if best in
      (membership(j,clb) > min_memb)        % neighborhood,
      clb = leave_club(j,clb);              % leave rand club.
    end
    if (worst(neighbors(j,clb),p) == j)&&...% if worst in
      membership(j,clb) < max_memb)         % neighborhood,
      clb = join_club(j,clb);               % join rand club.
    end
    if (mod(iter,rr) == 0)&&...             % check every
      (membership(j,clb) $\sim$= def_memb)  % rr iterations
        clb = memb_updt(j,clb);             % and update
    end                                     % membership
  end                                       % levels.
  iter = iter + 1;             % increment iterations counter.
  term_cond = updt(term_cond); % update termination condition.
end
\end{lstlisting}

Where \texttt{min\_memb}, \texttt{max\_memb}, and \texttt{def\_memb}
are the minimum, maximum, and default membership levels,
respectively.

Figure~\ref{fig:SI_C-PSO} shows a snapshot of the clubs during an
execution of the C-\ac{PSO} algorithm. In this example, the swarm
consists of 8 particles, and there are 6 clubs available for them to
join. Given the previous code, and that the minimum, default and
maximum membership levels are 2, 3 and 5 respectively. The following
changes in membership will happen to particles in
Figure~\ref{fig:SI_C-PSO} for the next iteration which is a multiple
of rr:

\begin{enumerate}
\item Particle$_3$ will leave club$_{1,2}$ or $_3$ because it is the best
particle in its neighborhood.

\item Particle$_5$ will join club$_{1,2}$ or $_4$ because it is the worst particle in its neighborhood.

\item Particle$_2$ will leave club$_{1,2,3}$ or ${_4}$, while particle$_4$ will join
club$_{2,3,4}$, or $_6$ to go one step towards default membership
level because they do not show extreme performance in their
neighborhood.
\end{enumerate}

\subsection{Flow of influence}

The flow of influence or how the effect of the best performing
particles spreads and affects other particles in the swarm is
critical to the performance of all \ac{PSO} algorithms. If the
influence spreads quickly through the swarm, they get strongly
attracted to the first optimum they find, which is a local optimum
in most cases. On the other hand, if the influence spreads slowly,
the particles will go wandering in the search space and will
converge very slowly to the global or a local optimum.

In order to study the effect of different default membership levels
on the flow of influence the following experiment were conducted. A
swarm of 20 particles is created. Clubs membership is assigned
randomly but every particle joins exactly $m$ of total 100 clubs.
The membership level $m$ is kept fixed for every single run, so best
and worst performing particles do not leave or join clubs.  All the
particles are initialized to random initial positions in the range
$[1000 2000]^n$ except for one particle which is initialized to
$[0]^n$. The value of each particle to be minimized is simply the
sum of its coordinate position values.

\begin{figure}
\centering%
\begin{minipage}[b]{0.45\textwidth}
\centering%
\includegraphics[width=\textwidth]{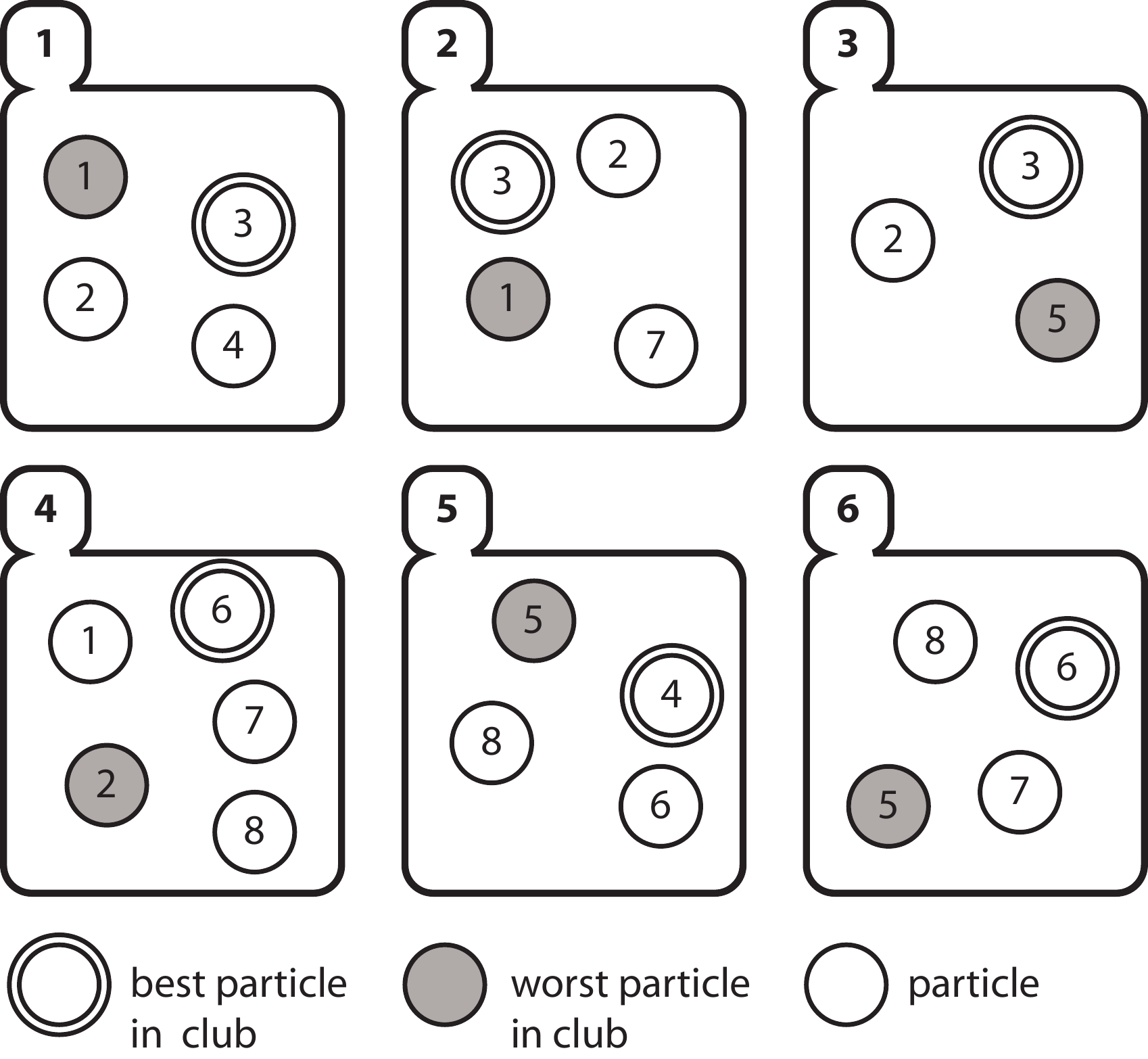}%
\end{minipage}
\hfill%
\begin{minipage}[b]{0.45\textwidth}
\centering%
\includegraphics[width = \textwidth]{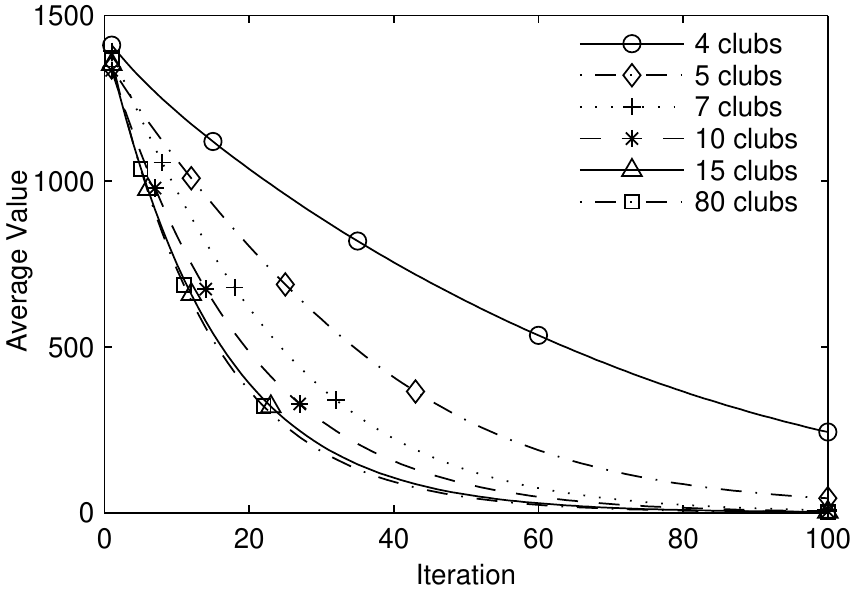}%
\end{minipage}\\

\begin{minipage}[t]{0.45\textwidth}
\caption[A snapshot of clubs' membership]{A snapshot of clubs' membership during an iteration of the C-PSO algorithm  \label{fig:SI_C-PSO}}%
\end{minipage}
\hfill
\begin{minipage}[t]{0.45\textwidth}
\caption[Effect of membership level on the flow of influence]{Effect
of different default membership levels on the speed
of the flow of influence \label{fig:SI_influence}}%
\end{minipage}
\end{figure}

The flow of influence for some default membership levels is shown in
Figure~\ref{fig:SI_influence}. The average value of all particles is
shown against search progress. The average value of particles
decreases because they are influenced by the best performing
particle which has a value of `0'. So, rapid decrease of the average
value indicates faster flow of influence speed. It is clear that the
flow of influence speed monotonically decreases with decreasing
default membership levels.

\subsection{Experiments}
The goal of the following experiments is to test and analyze the
effect of the dynamic social network employed in the proposed
\ac{C-PSO} algorithm on its performance and compare it with the
performance of other \ac{PSO} algorithms which have static social
networks. Five well known benchmark problems were used and presented
in Table~\ref{tb:SI_benchmark}. The first two functions are simple
unimodal functions. They test the ability of the optimizers to deal
with smooth landscapes. The next three functions are multimodal
functions containing a considerable number of local minima where the
algorithm may fall into, so these functions test the ability of the
algorithm to escape these traps. The performance of the different
optimizers is compared using two criteria which were used in
\cite{JansonM05}. The first one is the ability to escape local
minima, and is measured by the degree of closeness to the global
optimum the optimizer achieves after a long number of iterations.
The second one is the convergence speed, which is measured by the
required number of iterations to achieve a certain degree of
closeness to the global optimum in the evaluation space. Using these
metrics on the five benchmark functions, three versions of the
\ac{C-PSO} with different default membership levels of 10, 15 and 20
from a total number of 100 clubs are compared with gbest and lbest
\ac{PSO} algorithms. The three default membership levels are chosen
based on initial empirical results. It was found that lower
membership levels decrease the speed of flow of influence, as shown
previously, which was reflected on slow convergence. While higher
membership levels cause premature convergence. For all simulation
runs the following parameters were used. $\varphi_1 = 1.494$,
$\varphi_2 = 1.494$, which were used in \cite{JansonM05} and
suggested in \cite{Trelea03}. For gbest and lbest $w = 0.729$ as in
\cite{JansonM05} and \cite{Trelea03}, while the value of $w$ for
\ac{C-PSO} which reflects the range of the random inertia weight is
presented in Table~\ref{tb:SI_benchpara} for each problem. It was
found that when the values of these inertia weights increase, the
particles start wandering in the search space and the convergence
speed decreases, and when decreased, the particles converge
prematurely. The minimum and maximum allowed membership levels are 5
and 33 respectively, while $rr = 2$. A smaller value for $rr$ caused
the particles to get back quickly to their default membership level
and leads to fast oscillation between two membership levels in some
situations which causes excessive wandering. While a higher value
for $rr$ than the one chosen here causes the inferior particles to
stay longer than needed at the extra clubs they joined and leads to
premature convergence. A swarm of 20 particles is used for all
simulation runs. The particles position and speed are randomly
initialized in the ranges shown in Table~\ref{tb:SI_benchpara}
depending on the benchmark problem used. The absolute speed values
for particles are kept within the $V_{max}$ limit for all dimensions
during simulation. On the other hand, particles movements are not
restricted by boundaries, so particles may go beyond the
initialization range and take any value. Every simulation run was
allowed to go for 10000 iterations, and each simulation has been
repeated 50 times. All simulation runs were executed using
MATLAB$^\circledR$\ R2006a.

\begin{table}
\caption{Benchmark functions \label{tb:SI_benchmark}} \centering
\begin{tabular}{@{}p{2cm}l @{}l}
\hline%
Sphere \mbox{(unimodal)}&       $\displaystyle f_1(x) = \sum_{i=1}^{n}x_i^2$\\
\hline%
Rosenbrock \mbox{(unimodal)}&  $\displaystyle f_2(x) =
\sum_{i=1}^{n-1} \left[
100(x_{i+1} - x_i^2)^2 + (x_i - 1)^2 \right]$\\
\hline%
Rastrigin \mbox{(multimodal)}& $\displaystyle f_2(x) =
\sum_{i=1}^{n} \left[ x_i^2 -
10\cos(2 \pi x_i) + 10 \right]$\\
\hline%
\mbox{Schaffer's $f_6$} \mbox{(multimodal)}&    $\displaystyle
f_4(x) = 0.5 +
\frac{\sin^2(\sqrt{x^2 + y^2}) - 0.5}{(1 + 0.001(x^2 + y^2))^2}$\\
\hline%
Ackley \mbox{(multimodal)} &$\displaystyle f_5(x) = -20
\exp\left(-0.2\sqrt{\frac{1}{n}\sum_{i=1}^n x_i^2} \right) - \exp
\left( \frac{1}{n}\sum_{i=1}^n \cos(2 \pi x_i) \right) + 20 +
e$\\
\hline%
\end{tabular}
\end{table}

\begin{table}
\caption{Parameters for benchmark functions \label{tb:SI_benchpara}}
\centering
\begin{tabular}{l|c|c|c|c}
\hline%
Function &Dim.  &Init. range &$V_{max}$ &$w$ (C-PSO)\\
\hline \hline%
Sphere          &30     &$[100; 100]^n$     &100    &1.2\\
Rosenbrock      &30     &$[-30;30]^n$       &30     &1.2\\
Rastrigin       &30     &$[-5.12; 5.12]^n$  &5.12   &1.4\\
Schaffer's $f_6$&2      &$[-100; 100]^n$    &100    &1.65\\
Ackley          &30     &$[-32; 32]^n$      &32     &1.36\\
\hline
\end{tabular}
\end{table}

\subsection{Results}
Each graph presented in this section represents the average of the
50 independent simulation runs for all optimizers unless otherwise
stated.

\subsubsection{Escaping Local Minima}
Regarding the first criterion which is the ability of the algorithm
to escape local minima. The Sphere and Rosenbrock problems have the
lowest number of local minima. Their unique minimum makes them the
easiest of the five benchmark problems in finding the global
minimum.

\begin{figure}
\begin{minipage}[b]{0.45\textwidth}
\centering%
\includegraphics[width = \textwidth]{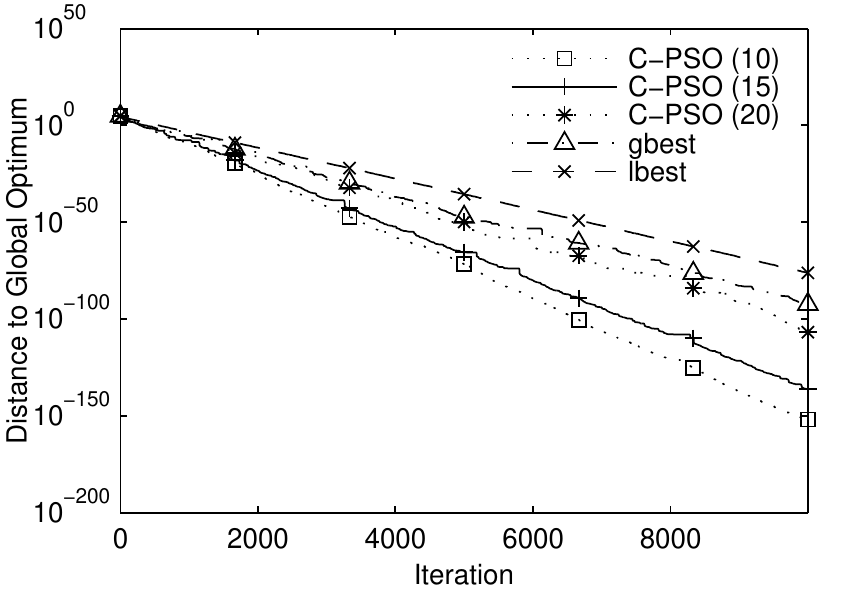}
\caption[Closeness to global optimum for the Sphere
problem]{Sphere--closeness to global optimum for C-PSO(10, 15, 20),
gbest, and lbest \label{fig:SI_C-PSO_sphere}}
\end{minipage}
\hfill
\begin{minipage}[b]{0.45\textwidth}
\centering%
\includegraphics[width = \textwidth]{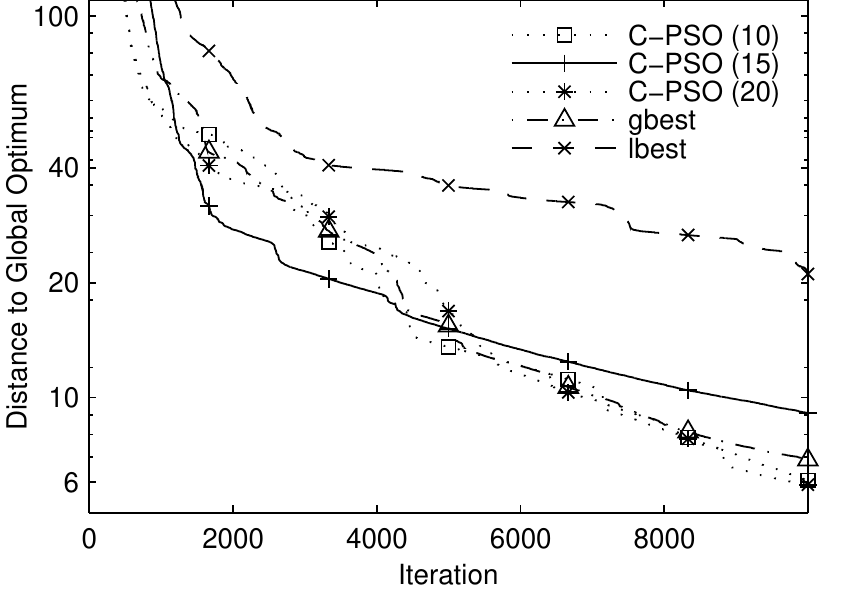}
\caption[Closeness to global optimum for the Rosenbrock
problem]{Rosenbrock--closeness to global optimum for C-PSO(10, 15,
20), gbest, and lbest \label{fig:SI_C-PSO_rosenbrock}}
\end{minipage}
\end{figure}

For the Sphere problem as shown in Figure~\ref{fig:SI_C-PSO_sphere},
all \ac{C-PSO} versions managed to finish closer to the unique
minimum than gbest and lbest, and the lower the membership level the
faster the algorithm converges. The lbest algorithm was the worst
performer followed by gbest. For the Rosenbrock problem presented in
Figure~\ref{fig:SI_C-PSO_rosenbrock}, \ac{C-PSO} (10, 20) and gbest
show very close performance, though gbest is little behind them.
\ac{C-PSO} (15) follows them by a short distance, while lbest is the
worst of all, lagging behind by a relatively long distance.

\begin{figure}
\begin{minipage}[b]{0.45\textwidth}
\centering%
\includegraphics[width = \textwidth]{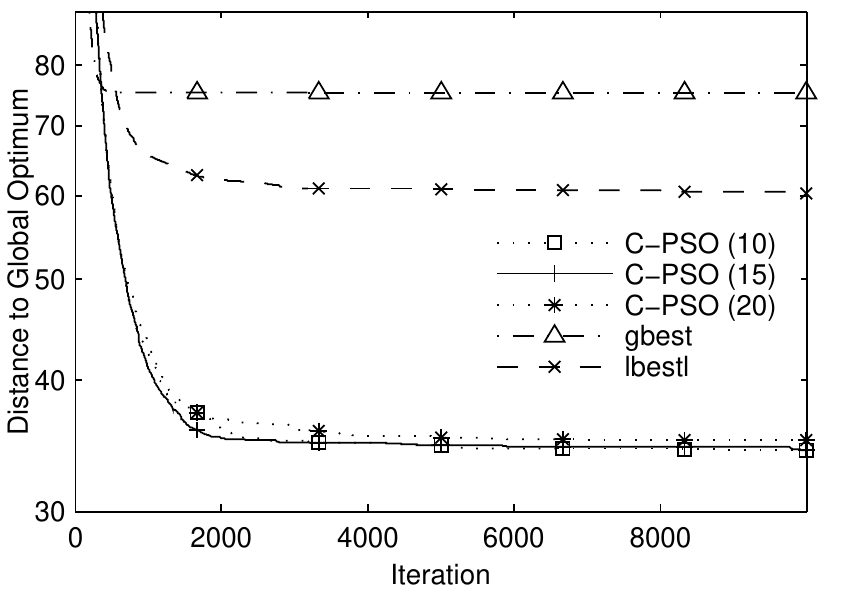}
\caption[Closeness to global optimum for the Rastrigin
problem]{Rastrigin--closeness to global optimum for C-PSO(10, 15,
20), gbest, and lbest \label{fig:SI_C-PSO_rastrigin}}
\end{minipage}
\hfill
\begin{minipage}[b]{0.45\textwidth}
\centering
\includegraphics[width = \textwidth]{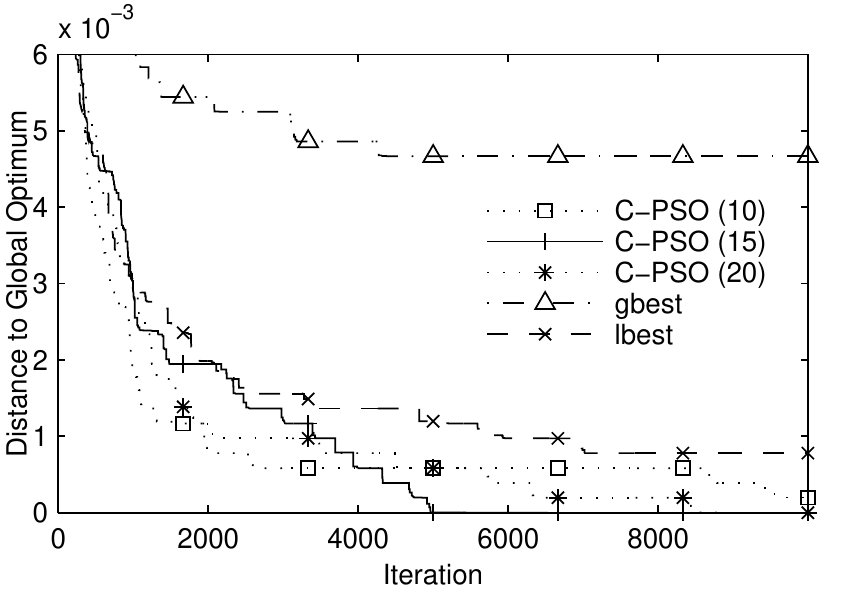}
\caption[Closeness to global optimum for the Schaffer's $f_6$
problem]{Schaffer's $f_6$--closeness to global optimum for C-PSO(10,
15, 20), gbest, and lbest \label{fig:SI_C-PSO_schaffer}}
\end{minipage}
\end{figure}

\begin{figure}
\centering
\includegraphics[width = 0.45\textwidth]{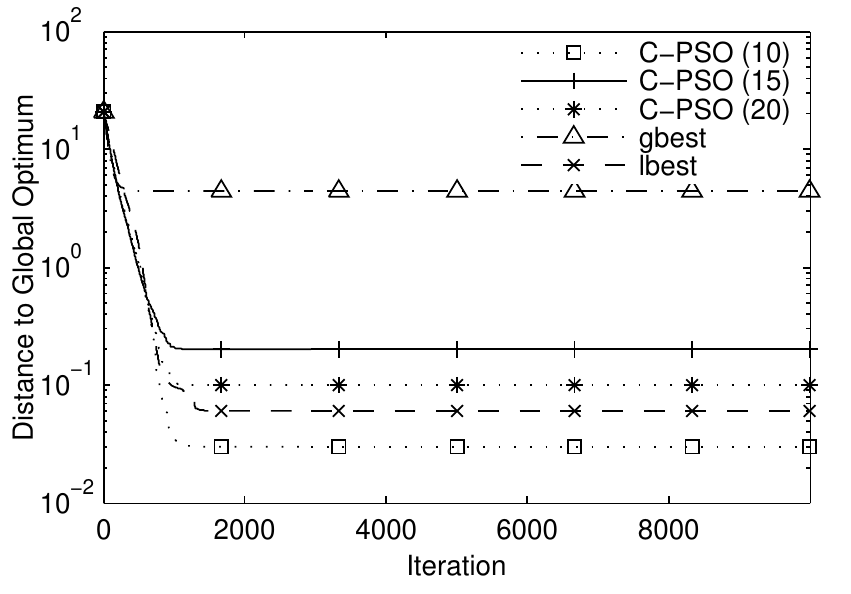}
\caption[Closeness to global optimum for the Ackley
problem]{Ackley--closeness to global optimum for C-PSO(10, 15, 20),
gbest, and lbest \label{fig:SI_C-PSO_ackley}}
\end{figure}

As shown in Figure~\ref{fig:SI_C-PSO_rastrigin} and
Figure~\ref{fig:SI_C-PSO_schaffer}, it's clear that all \ac{C-PSO}
versions perform better than both gbest and lbest for the Rastrigin
and Schaffer's $f_6$ test problems respectively. In both of them,
gbest gives the worst performance and converges prematurely in the
Rastrigin problem, followed by lbest as the second worst. All
\ac{C-PSO} versions give similar performance for the Rastrigin
problem and its hard to distinguish between them. As shown in
Figure~\ref{fig:SI_C-PSO_ackley} for the Ackley problem. \ac{C-PSO}
(10) outperforms all the other algorithms followed by lbest,
\ac{C-PSO} (20, 15) respectively, while gbest suffers premature
convergence again and falls by a long distance behind. The distances
to global optima after 10000 iterations of the optimizers are shown
in Table~\ref{tb:SI_dist_glob}.

\begin{table}
\caption[Distance to global optima]{Distance to global optimum after
10000 iterations \label{tb:SI_dist_glob}} \centering
\begin{tabular}{l||c|c|c|c|c}
\hhline{-||-----}%
Algorithm   &Sphere         &Rosenbrock &Rastrigin  &Schaffer's $f_6$   &Ackley\\%
\hhline{=::=====}%
lbest       &7.4e-77        &21.08      &60.37      &7.7e-4             &0.06\\%
gbest       &4.2e-93        &6.88       &75.32      &4.66e-3            &4.45\\%
C-PSO(20)   &1.3e-107       &{\bf 5.92} &35.10      &{\bf 0}            &0.10\\%
C-PSO(15)   &5.4e-137       &9.11       &34.34      &{\bf 0}            &0.20\\%
C-PSO(10)   &{\bf 1.1e-152} &6.07       &{\bf 34.30}&1.9e-4             &{\bf 0.03}\\%
\hhline{-||-----}
\end{tabular}
\end{table}

As expected, the performance of gbest and lbest depend on the
problem they optimize. For the first two problems which have a
single optimum, gbest performs better than lbest, as all the
particles get strongly attracted to the unique optimum due to the
fully connected social network in gbest. On the other hand, lbest
goes wandering and converges slowly. For the last three problems,
which have many local optima, lbest outperforms gbest. The partially
connected social network of lbest creates many points of attraction
for the particles in the swarm that help them escape some local
optima compared to gbest. Unlike gbest and lbest, \ac{C-PSO}
performance is much less problem dependent. \ac{C-PSO} (10)
outperformed both gbest and lbest for all problems.

The results obtained for the Sphere problem were unexpected. The
unique minimum and the non-deceptive landscape of the problem make a
perfect match with gbest. The fully connected social network should
do a better job in attracting the particles to the unique global
minimum than any other social network. These results necessitated
further investigation into the behavior of the optimizers and
specially the flow of influence through the swarm in unimodal and
multimodal problems. So the following experiment were conducted.

\subsection{Further Investigation of Optimizers' Behaviors}
\label{ch:SI_sec:C-PSO_sub_furinv} 
\begin{figure}
\begin{minipage}[b]{0.45\textwidth}
\centering
\includegraphics[width = \textwidth]{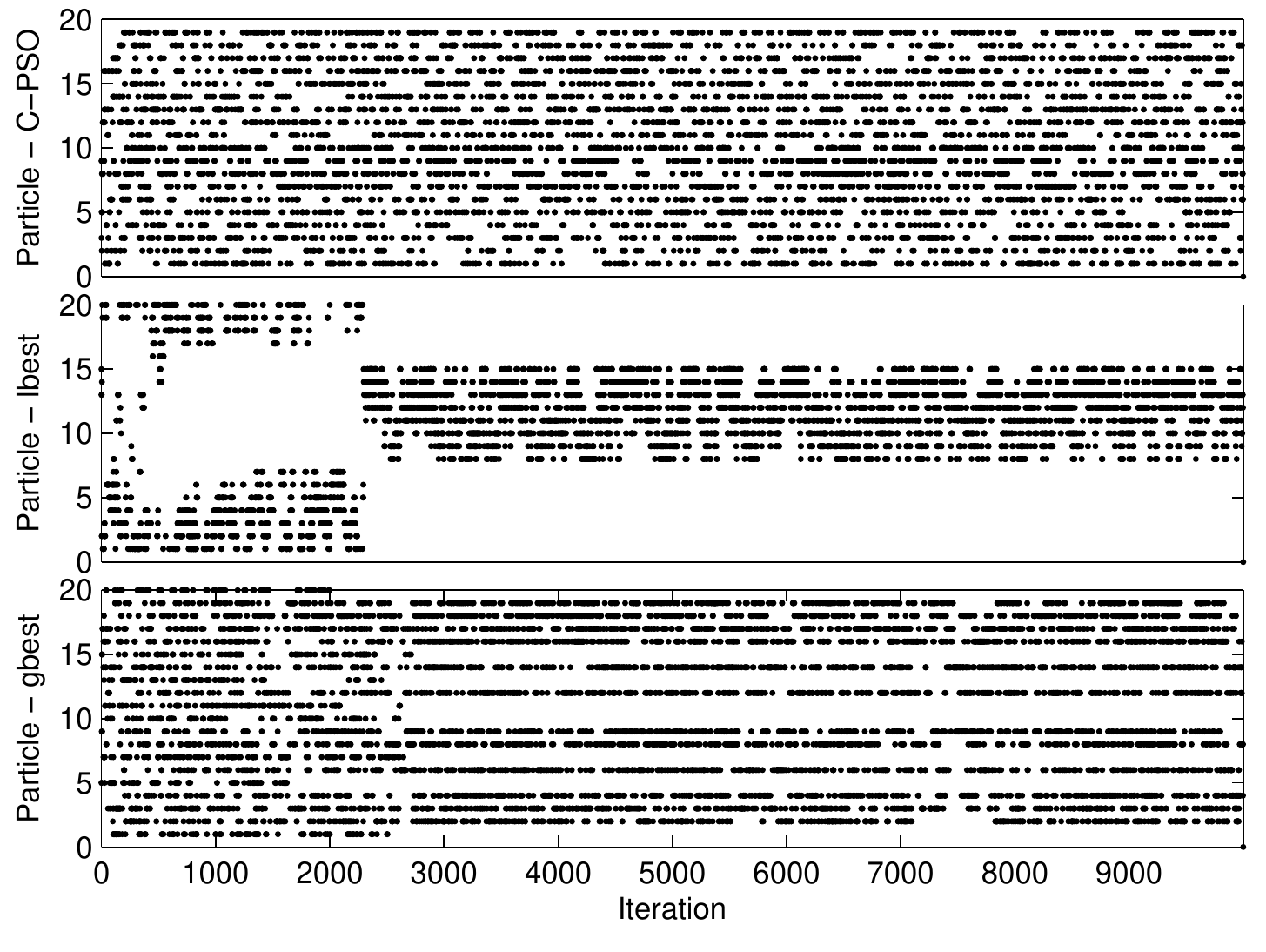}
\caption[Best particle in the swarm for Rosenbrock problem]{Best
particle in the swarm for Rosenbrock problem using C-PSO(10) (top),
lbest, and gbest (bottom) \label{fig:SI_C-PSO_pro2}}
\end{minipage}
\hfill
\begin{minipage}[b]{0.45\textwidth}
\centering%
\includegraphics[width = \textwidth]{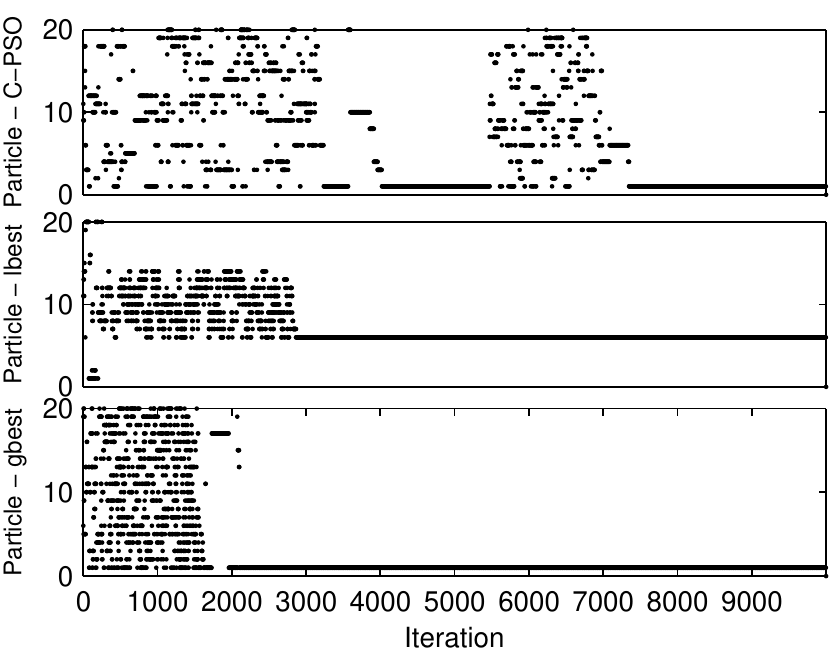}
\caption[Best particle in the swarm for Rastrigin problem]{Best
particle in the swarm for Rastrigin problem using C-PSO(10) (top),
lbest, and gbest (bottom) \label{fig:SI_C-PSO_pro3}}
\end{minipage}
\end{figure}

The three optimizers, \ac{C-PSO}
(10), lbest and gbest were run on the unimodal Rosenbrock test
problem and the multimodal Rastrigin problem. During the simulation
run the index of the best performing particle in the swarm was
recorded for each iteration. The same parameters used previously for
the first criterion were used for this experiment.
Figure~\ref{fig:SI_C-PSO_pro2} and Figure~\ref{fig:SI_C-PSO_pro3}
show the best performing particles in \ac{C-PSO} (10) (top), lbest
and gbest (bottom) for the Rosenbrock and Rastrigin problems
respectively. For each plot, the index of particles (20 particles)
is drawn against the number of iterations elapsed. A dot at (13,
5000) indicates that `particle 13' has the global best value in the
swarm during `iteration number 5000'.

First, considering Rosenbrock problem shown in
Figure~\ref{fig:SI_C-PSO_pro2}. The status of being the best
performing particle in the case of \ac{C-PSO} is almost uniformly
distributed over all particles, once a particle finds a good
solution, another particle finds a better one. A reason for this
behavior is that once a particle finds the good solution it becomes
the best particle in the swarm, making it the best in its
neighborhood as well. The particle shrinks its membership level one
by one and reduces its influence on other particles accordingly.
Neighbors of this superior particle will carry its influence to
other clubs they are member of, so other particles are still
indirectly guided by it, but are more free to find a steeper way
down the hill to the global optimum. Once a particle finds it, it
becomes the new best particle and continues or starts shrinking its
membership level (because it may become the best in its neighborhood
before becoming the global best). The particles which are no longer
the best in their neighborhood regain their default membership level
to increase their chance in learning from better particles to become
the new global best, and the cycle continues.

On the other hand, the best particle status in lbest goes bouncing
between two particles on the ring (note that particle 1 is connected
to particle 20) as shown in Figure~\ref{fig:SI_C-PSO_pro2} (middle).
The particles outside this arc search totally inefficient regions of
the search space. This is clear from the fact that none of them
shows up even once as the best particle in the swarm for the last
7500 iterations. This clustering mechanism may help the algorithm to
overcome local optima in multimodal problems, but in unimodal
problems it has detrimental effect. The reason that gbest algorithm
came second to \ac{C-PSO} in both unimodal problems is clear in
Figure~\ref{fig:SI_C-PSO_pro2} (bottom). After around 2600
iterations, 12 particles acted as guides for the other 8 particles
and literally dragged them behind. None of the 8 particles showed
superior performance till the end of the 10000 iterations.

Second, considering the Rastrigin problem presented in
Figure~\ref{fig:SI_C-PSO_pro3}, this multimodal problem requires
diversity in the swarm and a clever social network to overcome local
optima. The property of fully connected social network in gbest
provokes all the particles to jump to the best found position by all
particles in the swarm. This makes the first 1500 iterations for
gbest look almost the same for both unimodal and multimodal
problems. But after these 1500 iterations the algorithm prematurely
converges in the case of the multimodal Rastrigin problem as shown
in Figure~\ref{fig:SI_C-PSO_pro3} (bottom). On the other hand, lbest
algorithm presented in Figure~\ref{fig:SI_C-PSO_pro3} (middle)
maintains its diversity for a longer period than gbest does. Along
with its clustering property explained earlier, it manages to escape
local optima to some extent and get closer to global optimum than
gbest can get.

Finally for the \ac{C-PSO} optimizer as shown in
Figure~\ref{fig:SI_C-PSO_pro3} (top), the algorithm maintains
diversity longer than gbest and lbest do. Moreover, the best
performing particle status is distributed over the particles, unlike
lbest, and the particles do not jump over the best particle once it
emerges. This can be seen as the particles create more clusters than
in the case of lbest and gbest. These clusters represent local
optima found by the particles. The most interesting result found is
the ability of the \ac{C-PSO} to explore new regions after a period
of stagnation.

As can be seen \ac{C-PSO} finds better regions at around iteration
5200 after it has stagnated for nearly 2000 iterations. An
explanation for this behavior is that the best performing particles
in their neighborhood create different points of attraction for the
particles. The particles are grouped according to their clubs'
membership and search the space around these points of attraction.
At the same time, the worst particles on their neighborhood expand
their membership and bridge the influence between different groups
of searching particles. If a searching group finds a better
solution, its influence is transmitted over the bridge acting
particles to other groups and diverts them from searching
inefficient regions indefinitely. They start searching for other
optima which could be better than the best one found and create
different points of attraction, and the cycle goes on.

\subsection{Convergence Speed}
The second criterion to be considered is the convergence speed of
the algorithms. As explained earlier, it is being measured by the
number of iterations the algorithm takes to reach a certain degree
of closeness to the global optimum. This number of iterations should
be small enough to reflect the ability of the algorithm to converge
rapidly, and not its ability to escape local optima and achieve
better values at later stages of the run. On the other hand, the
closeness value chosen should lie close enough to the global optimum
to be efficient in practical applications. The closeness values for
the five benchmark problems that are satisfied by most algorithms
are chosen to be around the range of [500, 1000] iterations. These
closeness values are shown in Table~\ref{tb:SI_C-PSO_no_iter} next
to problem names.

\begin{table}
\centering%
\begin{minipage}[b]{10cm}
\caption[Convergence speed to global optima]{Number of iterations
needed to reach a certain degree of closeness to global optima for
the five optimizers. (best values are bold faced)
\label{tb:SI_C-PSO_no_iter}} \centering
\begin{tabular}{l||c|c|c|c|c}
\hhline{-||-----}%
Algorithm   &Avg.          &Med.            &Max.           &Min.
&Suc.\%\\%
\hhline{=::=====}%
            &\multicolumn{5}{c}{Sphere--(closeness = 0.0001)}\\%
\hhline{-||-----}%
lbest       &1030.8         &1036           &1103           &965            &{\bf 100}\\%
gbest       &684.88         &672            &1012           &489            &{\bf 100}\\%
C-PSO (20)  &611.68         &571            &1057           &421            &{\bf 100}\\%
C-PSO (15)  &528.18         &{\bf 506.5}    &711            &{\bf 417}      &{\bf 100}\\%
C-PSO (10)  &{\bf 518.14}   &513.5          &{\bf 652}      &443            &{\bf 100}\\%
\hhline{=::=====}%

            &\multicolumn{5}{c}{Rosenbrock--(closeness = 100)}\\%
\hhline{-||-----}%
lbest       &1429.6     &907        &7465       &604        &98\\%
gbest       &874.3      &425        &6749       &251        &{\bf 100}\\%
C-PSO (20)  &697.3      &424        &4537       &240        &{\bf 100}\\%
C-PSO (15)  &{\bf 569}  &473        &{\bf 1605} &{\bf 218}  &98\\%
C-PSO (10)  &725.8      &{\bf 376}  &6016       &226        &98\\%
\hhline{=::=====}%

        &\multicolumn{5}{c}{Rastrigin--(closeness = 50)}\\%
\hhline{-||-----}%
lbest       &1695.7         &1068           &8015       &500        &26\\%
gbest\footnote[1]{Not considered in comparison due to its very low
success rate}
            &250            &221            &313        &216        &6\\%
C-PSO (20)  &813.9          &702            &3396       &{\bf 254}  &88\\%
C-PSO (15)  &{\bf 695.4}    &{\bf 597.5}    &{\bf 1829} &262        &88\\%
C-PSO (10)  &753.3          &667            &1932       &299        &{\bf 96}\\%
\hhline{=::=====}%

        &\multicolumn{5}{c}{Schaffer's $f_6$--(closeness = 0.001)}\\%
\hhline{-||-----}%
lbest       &1076.2         &422            &7021       &84         &92\\%
gbest       &{\bf 791.1}    &{\bf 279.5}    &{\bf 4276} &60         &52\\%
C-PSO (20)  &1138.3         &524            &8462       &80         &{\bf 100}\\%
C-PSO (15)  &1120.1         &432            &4966       &88         &{\bf 100}\\%
C-PSO (10)  &945.6          &401            &9668       &{\bf 48}   &98\\%
\hhline{=::=====}%

        &\multicolumn{5}{c}{Ackley--(closeness = 0.01)}\\%
\hhline{-||-----}%
lbest       &968.2          &954.5          &1531       &827        &96\\%
gbest$^a$       &499            &499            &499        &499        &2\\%
C-PSO (20)  &831.1          &806            &1148       &610        &92\\%
C-PSO (15)  &{\bf 800.6}    &{\bf 793.5}    &{\bf 1141} &{\bf 570}  &84\\%
C-PSO (10)  &863.7          &841            &1151       &672        &{\bf 98}\\%
\hhline{-||-----}%
\end{tabular}
\end{minipage}
\end{table}

The figures presented in Table~\ref{tb:SI_C-PSO_no_iter} are
compiled from the same results data set collected for the first
criterion. They represent the Average, Median, Maximum, Minimum and
Success rate of 50 independent simulation runs for the five
optimizers. Only data of successful runs were used to evaluate these
values, so the sample number is not the same for all figures.

It can be seen from Table~\ref{tb:SI_C-PSO_no_iter} that \ac{C-PSO}
(15) achieves the overall best results. For the Sphere problem, all
the algorithms achieve the desired closeness in every single run,
although \ac{C-PSO} (10, 15) come ahead of them. The situation is
similar in the second unimodal Rosenbrock problem, however, the
success rate is lower for lbest and \ac{C-PSO} (10, 15).

Moving to multimodal problems, gbest shows poor performance in
reaching the closeness values. For Rastrigin and Ackley problems, it
only succeeds in six and two percent of the runs respectively,
compared to much higher success rates in all \ac{C-PSO} versions.
\ac{C-PSO} (15) outperform all the other algorithms for Rastrigin
and Ackley problems, except for the Rastrigin problem where it comes
second to \ac{C-PSO} (20) regarding the minimum number of iterations
in the 50 samples. gbest were not considered in the comparison for
Rastrigin and Ackley problems due to its very low success rate.
Finally for Shaffer's $f_6$ problem, gbest achieved the best results
for the mean, median and maximum number of iterations. It should be
noted however that it has a low success rate of 52\% which is almost
half the success rate for all \ac{C-PSO} versions. This low success
rate makes it unreliable in practical applications.

\section{Comparison to EAs}
The \ac{PSO} algorithm was compared to \acp{EA} once it was
developed \cite{kennedy:1995:PartSwarmOpt}. Kennedy and Eberhart
described it as an algorithm that stands somewhere between \acp{GA}
and \ac{EP}; the adjustment towards local and global best positions
exploits the accumulated knowledge by the swarm, analogously the
crossover operator recombines parts of good parents hopefully to
produce good offspring. While \ac{PSO} resembles \ac{EP} in its
reliance on stochastic processes ($rand_1$, $rand_2$). Some people
argue that \ac{PSO} is indeed an \ac{EA}, but Kennedy and Eberhart
do not share this opinion \cite{KennedyEberhart:SIbook:01}. However,
it is unquestionable that both of them are nature
inspired---population-based algorithms, they do operate a population
of complete solutions; at any moment during the run, the algorithm
could be stopped and $N$ solutions to the problem will be available,
where $N$ is the population size.

\acp{GA} employ the concepts of evolution and Darwinian selection,
while \ac{PSO} is powered by social interactions. In \acp{GA}, the
worst individuals \emph{perish} and get replaced by more fit ones,
while in \ac{PSO} the worst particles do \emph{learn} from their
neighbors, however subsequent positions of the swarm may be
perceived as different \ac{GA} generations; at the end of each
iteration of a \ac{PSO} algorithm, the old particles are killed and
replaced by their offspring at the next positions. But this analogy
entails many assumptions, such as assuming that every particle must
produce one, and only one offspring, and this offspring will replace
its parent.

The social interaction among swarm members brings them closer to
each other, the particles tend to pursue their superior neighbors
and thus converge to a point in one of the promising regions they
discovered. On the other hand, highly fit individuals tend to
produce more offspring than less fit individuals in the population
of a \ac{GA} algorithm, leading to convergence to a point in one of
the best found regions. \ac{PSO} particles do learn, while \ac{GA}
individuals do evolve.

The crossover operator of a \ac{GA} resembles the social
interactions of \ac{PSO} particles. When two individuals are
recombined, the produced offspring stands somewhere between the
parents, while social interaction between two particles at two
points in space brings one of them to an intermediate point.

The mutation operator of a \ac{GA} has its counterpart as well. The
stochastic variables $rand_1$, and $rand_2$ are their image in
\ac{PSO}. The mutation operator ensures that the probability of
sampling ever point in the solution space is never zero. It kicks an
individual to a point that other deterministic operators may never
take it to. Similarly, the stochastic variables in \ac{PSO} drives
the particles to regions that may seem poor when evaluated by their
personal and neighbors experience measures.

\acp{GA} was proposed in 1960s while \ac{PSO} was developed in 1995.
Due to this gap in time, \acp{GA} have been experimented and
developed more rigorously than \ac{PSO}. \acp{GA} are applied in
many real-world applications ranging from control engineering, water
treatment, and job scheduling, to image processing, and military
tactics. \ac{PSO} on the other hand has been applied less
extensively to real-world applications due to its young age and
scant mathematical analysis. The stability, reliability, and
availability of both algorithms are not verified so far by closed
form mathematics, thus their use is not advisable in critical
applications where its failure may result in injury or expensive
repair.

Despite many empirical studies and comparisons between \ac{PSO} on
one hand and \acp{EA} or \acp{GA} on the other hand, no definite
conclusion was shared among these different studies. Some of them
suggested the use of \ac{PSO} in some practical problems
\cite{Mous05}, while others were less enthusiastic about it
\cite{hassan05}. This different conclusions may further assert the
\ac{NFL} theory, which states that \cite{Wolpert:1995:NFL}:
\begin{quote}
\emph{``\ldots all algorithms that search for an extremum of a cost
function perform exactly the same, when averaged over all possible
cost functions''}
\end{quote}

%% file: ACE/ACE.tex
\chapter{Applications in Control Engineering}
\label{ch:EA_pro}

It was the feats accomplished by evolution, natural selection and
social interactions that inspired researchers to mimic these natural
mechanisms as computer paradigms to solve numerous problems. These
optimizers were already in the works long time ago, and their
obvious successful results is the strongest proof of their
applicability in real-world problems. But reverse engineering
\emph{nature} resulted in a new situation. In an ordinary situation,
an engineer who is faced with a problem works-out a
\emph{mathematical analysis} of the problem and proceeds step by
step, in a logical fashion, to create a problem \emph{solver}. The
applicability and efficiency of this solver are \emph{tested} and
\emph{verified} by practical application over a period of time. On
the other hand, the efficiency and applicability of a solver
employed by nature has already been \emph{tested} and
\emph{verified} through millennia. Researchers extracted the
\emph{solver} out of the system and applied it to different
problems. Today, many researchers and scientists work out
\emph{mathematical analysis} to find out why these solvers work, and
develop a closed-form mathematical explanation and proof for them.

No wonder why a generic problem solver such as \acp{GA} would be
applied to complex problems which proved to be unyielding to the
rigid, conventional, analytical, problem solvers, such as control
engineering problems. In fact, the application of \acp{GA} in
control engineering was proposed concurrently with the algorithm
itself when John Holland presented them in his PhD thesis titled
``Adaptation in Natural and Artificial Systems: An Introductory
Analysis with Applications to Biology, Control, and Artificial
Intelligence'' back in 1975 \cite{holland75:adapt}. Although
\ac{PSO} is still in its infancy, the increasing number of
publications proposing and reporting encouraging results of its
application in control engineering problems indicates its
applicability and efficiency in such problems \cite{Kro03, Cao06,
Yoshida01, Miranda02, Zhao05}.

\section{Why Computational Intelligence?}
\label{ch:ACE_sec:why}

The major power of \ac{CI} techniques stem from their generic
nature, which tolerates lack of information about the system in
study, can handle a problem with mixed type of variables, welcomes
ill-shaped system landscapes, provide flexible way in representing
decision vectors, can efficiently handle constraints by different
ways, and are not subject to many of the limitations that
traditional optimizers are subject to. All these merits directly
address many of the difficulties found in control engineering
problems.

If it is hard to create a mathematical model for the control system
under study due to lack of information about it, its high
non-linearity, or its stochastic nature, then conventional
techniques, which are problem specific, will not produce
satisfactory results. On the other hand, \ac{CI} methods, which are
generic, can deal with this lack of information and uncertainty, and
any acquired information about the system can be utilized by the
algorithm and incorporated in the model on-the-fly.

Yet another benefit of the generic nature of \ac{CI} techniques is
their ability to handle problems with mixed types and units of
variables. Since the algorithm works on encoded variables, it can
handle solution vectors such as ($\text{H}_2\text{O}$, $2\pi\,''$,
$-4\,^{\circ}\text{C}$, Italy, $\frac{5.6}{7}$Kg) with no problem.
However, this is not an easy task for a conventional method indeed.

Traditional optimization techniques perform badly on problems with
ill-shaped landscapes. Multimodality, non-differentiability,
discontinuity, time-variance, and noise may render a traditional
method inefficient, or may even block its application. However,
these properties are not a source of difficulty for \ac{CI} methods
\cite{Fleming01}.

Any convenient representation of solutions can be used with a
\ac{CI} method. However, the adopted representation can dramatically
affect the performance of the algorithm used, and henceforth, the
quality of solutions obtained \cite{Rothlauf:2006:RGE}. One
representation may encode decision vectors in such a way that help
exploring the solution space more effectively than another
representation, while a third one could be computationally less
expensive and runs faster than a fourth one. Moreover, the
representation itself may evolve over time, so the algorithm would
eventually produce the most efficient representation which encodes
optimum solutions. This evolving representation is depicted in
\ac{GP} models \cite{koza:GPbook92}. A good representation should
come with suitable variation operators, or else it will not be
possible to generate new solutions from the old ones, or produce
mostly infeasible or inferior solutions which require extra
computational cost to repair them, which may lead to random search
or even worse. The flexibility in representation offered by \ac{CI}
optimizers may be seen as a sort of shortcoming or advantage.

Almost all control and engineering problems in general are
constrained problems. These constraints can be soft constraints,
such as the riding comfort of an elevator, or hard constraints, such
as the stability of the elevator and the limit of its actuators.
Dealing with these constraints, specially in discontinuous problems
poses quite a challenge for traditional optimizers. But for \ac{CI}
techniques, these constraints can be handled easily in different
ways. These constraints may be embedded in the representation scheme
used, so evolving the encoded parameters always yields a valid
solution. Another way of dealing with constraints is penalizing
unfeasible solutions, different penalty strategies can be used
considering the degree of constraint violation and the number and
type of constraints violated and so forth. Alternatively, the
constraints can be imbedded into the problem as new objectives and
then by solving the new problem as a non-constrained problem,
feasible solutions to the original problem are obtained
\cite{coello00treating}.

\ac{CI} methods are not subject to many of the limitations and
constraints that traditional methods must adhere to. For example,
the evaluation of the discrepancy between the expected output of a
controlled system and the obtained one, which is known as the error,
is done in many traditional techniques by using the \ac{RMS} of the
error. However this evaluation technique is biased. As shown in
\eqref{eq:CIC_RMS}, \ac{RMS} underestimates errors below 1 and over
estimates errors above 1 \cite{Ursem03}. This bias does not exist in
the \ac{SAE} evaluation as shown in \eqref{eq:CIC_SAE}. However,
only the former is applicable in traditional methods due to the
discontinuity of the absolute function used in the later. \ac{CI}
techniques, on the other hand, have no problem in utilizing any one
of them.

\begin{alignat}{5}
& \text{RMS}\quad &\Rightarrow \quad &(0.9 - 0.6)^2 &&=0.09 \qquad &&(3 -
(-5))^2 &=&4 \label{eq:CIC_RMS}\\
& \text{SAE}\quad &\Rightarrow \quad &|0.9 - 0.6|   &&=0.3  \qquad
&&|3 - (-5)|   &=&2 \label{eq:CIC_SAE}
\end{alignat}


\section{When to opt out?}
\label{ch:ACE_sec:optout}

Despite many benefits obtained by using \ac{CI} techniques in
control engineering applications, in some situations it is not
advisable to use them.

If the system under study is simple, well-known, with low degree of
randomness and tolerable amount of noise, and can be approximated by
a linear system with a low degree of error if it is not a linear
system in the first place, then traditional techniques providing
analytical solution may be the method of choice, and it is unlikely
that \ac{CI} techniques will outperform it. The flexibility offered
by the Swiss-knife with its various tools does not make its tiny
scissors a powerful tool in cutting material when compared to
conventional scissors that can only handle this task
\cite{Michal00}.

\ac{CI} methods are known to be computationally intensive. The
algorithm that operates a population of agents working (at least
virtually) in parallel and evaluates their fitness every iterations
then classifies or orders them according to their fitness is indeed
resource intensive, both in memory, and computational power. This
resource intensive property of \ac{CI} techniques present two
handicaps. First, it is expensive (money perspective), which adds
extra cost to the control system and reduces its price
competitiveness when compared to other control systems using
traditional techniques which are less computationally intensive.
Second, the computational power needed to run some \ac{CI}
optimizers may not be available, or could be hard to fit in the
control system, which may lead to slow and unsatisfactory
performance.

Although \acp{EA} and \ac{SI} methods have been applied in many
practical applications, and some engineering consultancy firms are
specialized in providing \ac{CI} solutions for various engineering
problems. These methods are not guaranteed to succeed. Their
convergence and stability have not yet been proved using closed loop
mathematical analysis, albeit few mathematical proofs
\cite{holland75:adapt, ClercKennedy:2002} that relied on many
assumptions to facilitate the analysis, and proved the convergence
of a simple model which is quite different from the practical models
being used today. It is common in \ac{CI} literature to verify the
effectiveness of an optimizer by repeatedly running it on a
benchmark problem over 50 times to withdraw the possibility of
chance in the solutions obtained. This scepticism is far more less
in the application of traditional techniques. Due to lack of
mathematical analysis and presence of stochastic variables in
\ac{CI} techniques, they are not welcome in many critical
applications where any failure may result in injuries or leads to
expensive repair.

Traditional techniques are used in many on-line applications today,
however the use of \ac{CI} methods in this venue presents quite a
challenge. In on-line applications, the system must decide on the
action at every time step, which requires reaching a good decision
during this time frame, but as pointed out earlier \ac{CI} methods
are computationally intensive, and henceforth, the time it takes to
converge or reach a good decision may exceed the limited time frame
which is obviously not acceptable. Moreover if the best individual
in the population is chosen at each time step even if convergence
was not achieved, the system will perform poorly and unsatisfactory
results will be obtained, or even worse, the system may go unstable.

Another major of concern when using \ac{CI} methods for on-line
applications is the nature of these optimizers themselves. These
techniques work by learning from their past performance and
mistakes. So if they utilized the process which they operate
directly, sever consequences may result. For example, it is known
that most \ac{CI} techniques provide poor solutions early in the
run, then the quality of these solutions improve as the algorithm is
fed-back with their results. However, it is not acceptable in most
applications to waste materials, cause damage to equipments, or
reach unstability for the sake of teaching the algorithm. To
overcome this undesirable behavior, a traditional method may exist
in the system as a control scheme backup and whenever the decision
of the \ac{CI} technique goes beyond a predefined threshold, the
traditional backup optimizer is activated, and the \ac{CI} one steps
aside. However restricting the operation of a \ac{CI} optimizer in
such a way prevents it from learning and improving its performance,
so even after long running time, its performance may still be
unsatisfactory.

Another possible use of \ac{CI} techniques is to use them in
optimizing the parameters of a controller on-line. In this
situation, the optimizer gets its feedback from the process itself
and a model of the process is not required. After the parameters
have been properly tuned, they get fixed on these values and the
controller is put in real on-line operation.

However, in some situations it becomes extremely hard to evaluate
any of the population individuals using the real system. For
example, it is not possible to stop an electricity power plant
generating power for millions and operate it under varying
conditions for the sake of system identification. In such a
situation, the algorithm may watch and learn from the normal
input-output data of the operating system. However, as with the case
of the backup system presented earlier, the algorithm will not be
able to develop a good understanding of the model because its
knowledge was confined to a small range of the ordinary input-output
data generated under normal conditions.

\section{Applications}
\label{ch:ACE_sec:apps}

\ac{CI} techniques have been used in numerous applications, they
vary from controller design and system identification, to robust
stability analysis, fault diagnosis and robot path planning
\cite{Fleming01, Fonseca95a}. They can be used as a direct or
indirect design tool.

\subsection{Controller Design}
A \ac{CI} optimizer can be used in tuning controller parameters,
designing its structure, or doing them both. It can be used as the
only design tool, or assisted with other techniques in a hybrid
design system. Depending on the application of the controller, the
fitness function will be defined accordingly. In a dairy processing
factory, it will be desirable to abruptly change the milk
temperature for the pasteurization process. In such a case, the
fitness function will be inversely proportional to the rise time of
the milk temperature while its overshoot will be less significant.
On other applications involving passenger's comfort, the overshoot
of the vehicle's speed should be emphasized in the fitness function
used.

\subsubsection{Parameter Tuning}
For the parameter tuning problem, the algorithm operates a
population of individuals where each one of them encodes a set of
controller parameters. The fitness of each individual is determined
either by a the controller itself or by a model of it. If the
representation used allows infeasible solutions, such as those
leading to system instability, they are penalized according the
penalty system used. Depending on the efficiency of the algorithm
used, the optimizer may eventually discover a good set of controller
parameters.

Many researchers used \acp{GA} and \ac{PSO} to tune \ac{PID}
controller parameters. Among these efforts, Herrero {\it et al.}
\cite{Herrero02} used a \ac{GA} to tune an optimal \ac{PID}
controller for a nonlinear process model in various situaltions:
model errors, noisy input, IAE minimization, and following a
reference models. They concluded by recommending its use for
off-line parameter tuning due to high computational cost required by
the optimizer. The same problem was tackled in \cite{Selvan03} but
using a \ac{PSO} algorithm with some modifications. The modified
\ac{PSO} optimizer achieved encouraging results by achieving lower
settling time over various transfer functions when compared to the
performance of a \ac{PID} controller tuned using the traditional
Ziegler-Nichols method.

Alternatively, \ac{CI} optimizers may be used to tune the parameters
of a controller \emph{indirectly}. In this case, they are used to
tune the parameter values of a design technique, which in turn,
tunes the controller parameters. For example, a \ac{GA} may be used
to tune the \ac{LQG} method parameters, or tune the pre- and
post-plant weighting functions for the $\mathcal{H}_{\infty}$
method, then any one of them can be used to tune the controller
parameters. Using this procedure, the stability of the system will
be guaranteed by the \ac{LQG} and $\mathcal{H}_{\infty}$ methods,
while a \ac{GA} will be used to find their best parameters
\cite{Fleming01}.

\subsubsection{Structure Design}
The power of \ac{CI} techniques is unleashed when utilized in the
structure design of a controller. Unlike parameter tuning where
there are traditional straightforward, well known and trusted
optimization techniques, the controller structure design requires
human experience in the field. Henceforth, developing a \ac{CI}
method that can manipulate and develop a good controller structure
would provide a faster and probably cheaper alternative for the
human based design.

\ac{GP} has an advantage over many other \ac{CI} techniques when
applied to the automated design of controller structure because the
structure of \ac{GP} individuals evolve concurrently with the value
of the genes. So, it will be straightforward to encode each
individual as a variable length sequence of parallel and series
building blocks of control elements. The \ac{GP} optimizer will
evolve these individuals using a library of those control elements
building blocks provided by the user to find the best structure and
parameters for this structure using the limited set of elements.

Koza {\it et al.} used \ac{GP} for automatically synthesizing the
design of a robust controller for a plant with a second-order lag
\cite{koza99automatic}. They reported better for the \ac{GP} method
when compared to a PID compensator preceded by a low-pass pre-filter
regarding the integral time-weighted absolute error, rise time, and
disturbance suppression.

\subsection{Fault Diagnosis}
Another application of \ac{CI} methods is system fault diagnosis
where these algorithms can be used to detect the presence of a
fault, isolate it, and identify or classify the fault
\cite{Fleming01}.

Miller {\it et al.} used \acp{GA} to spot the fault in a system
\cite{miller93evaluation}. Given the probabilities that a particular
disorder causes a particular symptom, the algorithm was able to spot
the fault from a collection of faults given the set of symptoms that
indicate that a problem exists.

In an effort to increase the reliability of the system, Coit and
Smith used a \ac{GA} to find the best system configuration by
selecting components and levels of redundancy to collectively meet
reliability and weight constraints at a minimum cost \cite{Coit96}.

\subsection{Robust Stability Analysis}
Fadali {\it et al.} used a \ac{GA} in a stability analysis context
\cite{fadali:1999:rsadsuga}. They reduced the stability robustness
analysis for linear, time-invariant, discrete-time system to a
problem of searching for the roots of the system's characteristic
polynomial outside the unit circle. Since the presence of such a
point that lies outside the unit circle is a sufficient condition
for system instability, a \ac{GA} searching for such a point will
classify the system as unstable if that point is found. However, if
the point was not found the stability of the system is not
guaranteed.

The \ac{GA} robust stability analysis is a strong contender when
compared to the traditional analysis techniques which rely on simple
uncertainty structures, or other techniques with more complex
structures but are infeasible to implement.

\subsection{Robot Path Planning}
Yet another field that \ac{CI} techniques has successfully applied
to is robot path planning. The path planning problem is an
optimization problem that involves computing collision free path
between two locations. Beside this goal there are other criteria of
minimizing the travel distance, time, energy, safety, and smoothness
of the path. The collection of all this criteria in a dynamic system
where many robots may work concurrently on the same object makes
conventional approaches such as cell decomposition, road map, and
potential field impractical to apply.

Elshamali {\it et al.} used a \ac{GA} with a floating point variable
length chromosome representation to tackle this problem
\cite{Elshamli04}. Each chromosome represents a path as a sequence
of nodes, the first one is the starting point and the final one is
the final destination. The variable length chromosome allows a
flexibility of path creation. They used a weighted combination of
the path distance, smoothness, and clearance as the fitness
function. They used five operators to evolve the population with
different probabilities, and used a strategy to ensure population
diversity. This algorithm was effective and efficient in solving
different types of tasks in dynamic environments.

\section{System Identification}
\label{ch:sysident}

Understanding why a system behaves in a certain way and predicting
its future behavior is a major field of research in control
engineering. The system in question could be any thing from bacteria
growth and stock markets to global warming and galaxy movements.
Although some of these systems, such as the stock market, is a
man-made system, its exact behavior cannot be determined given
current and past inputs to this system. This uncertainty in such
systems is due, in large part, to the complexity of the system in
question. In other situations, some simple man-made systems deviate
from its designed behavior due to aging, wearing out, or change in a
system parameter that was assumed to be static. For example, the
performance of car brakes changes over time. The behavior of natural
systems such as galaxy movements and global warming is far more
complex and is much more harder to understand.

\begin{figure}
\centering%
\includegraphics[width=10cm]{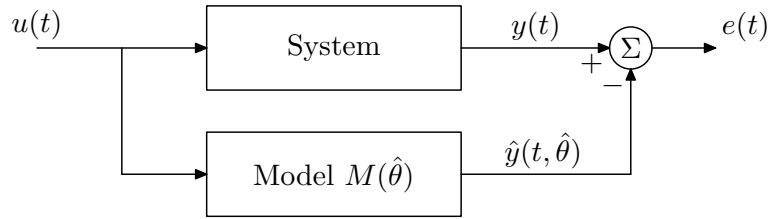}
\caption{Process of system identification \label{fig:CSSI_sysid}}
\end{figure}

System identification involves creating a model for the system in
question that, given the same input as the original system, the
model will produce an output that matches the original system output
to a certain degree of accuracy. The input or excitation to the
system and model, and their corresponding output are used to create
and tune that model until a satisfactory degree of model accuracy is
reached. As shown in Figure~\ref{fig:CSSI_sysid}, the input $u(t)$
is fed to both the system and the model $M(\hat{\theta})$, then
their corresponding outputs $y(t)$ and $\hat{y}(t,\hat{\theta})$ are
produced and matched. The error $e(t)$ reflects how much the model
matches the system; the lower the error, the more the model
resembles the system.

System identification of practical systems is not an easy task to be
accomplished by traditional techniques \cite{Ljung87a}. Most
real-world systems contain dynamic components. Due to this dynamic
nature of the system, the output of the system does not only depend
on the current input to the system, it depends on the past inputs
and outputs to and from the system. As the number of old data
affecting the system (maximum lag) increases, the number of terms
used in the classic system models increases substantially. Another
difficulty encountered in identifying real-world system is its
nonlinearity. In order to simplify the problem, many engineers
represent the non-linear system by a linear model, however the
performance of this simplified model will not be satisfactory in
mission critical applications. Although there are some classic
models that can represent non-linear models, such as the \ac{NARMAX}
model, the number of these model terms explodes as the degree of
nonlinearity increases. The number of terms in a \ac{NARMAX} models
for a modest real-world system with a nonlinearity order of 3 and a
maximum lag of 6 will have $\binom{15}{3}=455$ terms. Obviously, the
complexity of such a system and the huge volume of data required to
calculate the least-square estimates would render this model
impractical \cite{Fonseca95a}.

System identification is essential for many fields of study that
extend beyond control engineering. It is desirable to identify the
the global warming system in order to understand its mechanism and
try to slow it down or reverse it. It is crucial for an astronomer
who tries to understand the planetary movements and correlates it
with various astronomical phenomena. Creating models for human
organs helps physicians and medical engineers to create a substitute
for these organs. While understanding an industrial process helps an
engineer to control, maintain it, and diagnose its faults. The
applications of system identification is numerous, but for the sake
of brevity it will be approached here from a control engineering
perspective, though many of the concepts and procedures involved in
the identification process itself are essentially the same for
different applications.

\subsection{Identification Procedure}
The system identification process of constructing a system model can
be described by the following procedure \cite{Ljung87a}.
\begin{enumerate}[i)]
\item \emph{Data recording:} The input-output data of the system
are recorded. In some situations, it is possible to conduct an
experiment solely for this purpose. In this case, the engineer may
have choose when and what input and output data is to be recorded.
Furthermore, he may feed the input data of his choice that would
maximize the knowledge of the system. In other situations, the
engineer can only watch and record few input-output data that the
system allows him to monitor under normal operation. It is clear
that the data recorded in the later situation would be less
informative than the one recorded in the former.

\item \emph{Model Set Selection:} The next step is to choose a model set that
the system under study would be represented by one of its members.
This step is not a deterministic one, it is rather subjective. It
involves prior knowledge of the system, if available, and the
experience of the engineer plays a major here. The system could be
modeled by physical laws that reflect the dynamics of the system. A
model created by these laws which reflect the physical properties of
the system is called a \emph{white-box model}. However, creating a
white box model for real-world (complex) systems is a challenging
task. As a compromise to the lack of understanding of all physical
rules which drive the system, a model that imitates the real system
regarding input-output data is sought. A model that merely resembles
the system without reflecting a physical soundness is called a
\emph{black-box model}.

\item \emph{Model Selection:} After the model set is selected, the
best model in this set is selected using some of the input-output
data recorded previously. In this step, the model parameters are
tuned so that the model output would fit the system output as much
as possible. The quality of the model is based on a criterion chosen
a priori.

\item \emph{Model Validation:} The next step is to verify the
quality of the developed model. The quality assessment is done by
comparing the model output to the original system output when both
are fed with samples of the input-output data recorded previously
(in this case, the validation data are different from the data used
for model selection), or with the system in real operation situation
(not experimental mode). If the model meets the chosen criteria
which reflect the intended use of the model, the model is accepted,
otherwise, it is rejected and another model is created. This
procedure is repeated until a satisfactory model is created. It is
to be noted that a system model will only imitate the original
system in certain aspects of interest to the model designer. It will
never become a full and true description of the system
\cite{Ljung87a}.
\end{enumerate}
\begin{figure}
\centering%
\includegraphics{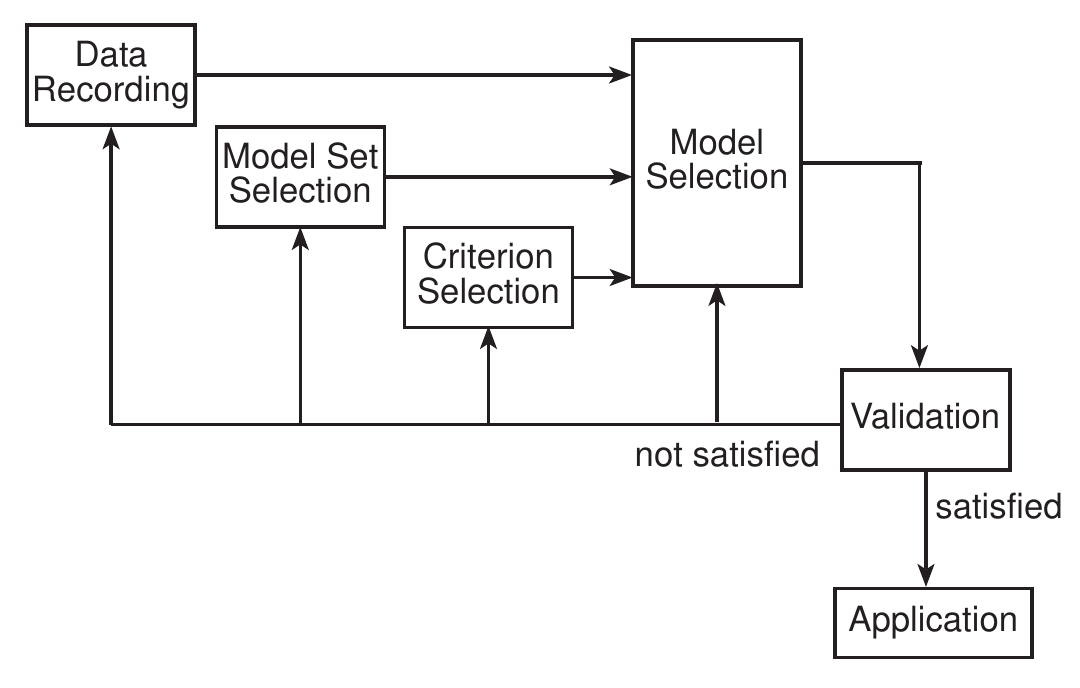}
\caption{Procedure of system identification
\label{fig:CSSI_sysid_proc}}
\end{figure}
The system identification procedure is depicted in
Figure~\ref{fig:CSSI_sysid_proc}, inspired from
\cite[pp.9]{Ljung87a}. The feedback from the `Validation' step to
other steps is used to refine or create a new model if the old one
did not provide a satisfactory performance. If the poor performance
is due to bad selection of criterion that does not reflect the
desired system attributes to be imitated by the model, then the
feedback to the `Criterion Selection' will be used to modify this
criterion or criteria. If deficiency in system matching is due to
bad model, then the feedback to `Model Selection' will be used to
change the order of the model or tune its parameters. If this
modification does not lead to a satisfactory behavior, then the
feedback to `Model Set Selection' will be used to select a different
model set. The mediocre performance could be attributed to bad
input-output data. These data may not be enough informative to be
used in model selection and tuning, henceforth, the feedback to the
`Data Recording' will be used to create another data set.

\subsection{Types of System Identification}
It is desirable in many control engineering problems to build a
model of the system under study. If the system is simple enough
(linear, time-invariant, deterministic, single-input single-output
system) the model can be built using building blocks representing
physical processes (white-box model), and by tuning the parameters
of these elements. However, most real-world problems are not that
simple. Non-linearity and time-invariance among other properties,
make it hard to create an acceptable model of the system using this
technique. Henceforth, System Identification can be decomposed into
identifying a structure for the system, and identifying the
parameters of a structure.

\subsubsection{Parameter Identification}
Because it is hard to create a white-box model for complex systems,
some parameterized models that can describe a system to some degree
of accuracy can be used instead (grey-box model). These models, such
as the ARX model and its variants, offer a reasonable degree of
flexibility so that if the model is well-tuned, the model output
will match the output of the real system to a high degree. The
process of identifying the values of these parameters are known as
\emph{parameter identification}.

\ac{CI} techniques can be used to tune the values of the
parameterized models in a similar way to that used in parameter
tuning of a controller presented earlier. The \ac{CI} algorithm
operates a population of potential solutions to the problem. The
individuals of this population encode different solutions to the
problem, and as the algorithm proceeds, it tries to find better
solutions according to some quality measure, which is typically the
difference between the predicted output of the system based on the
created model, and the measured output of the system (prediction
error). Depending on the efficiency of the algorithm used, it may
find a good set of parameters that produces a tolerable prediction
error. Following this approach, Voss and Feng used \ac{PSO} to find
the parameter set of the \ac{ARMA} parameterized model
\cite{Voss01}. They reported superior results of the \ac{PSO}-based
parameter tuning when compared to the International Mathematical
Libraries routines using noisy (real-world) data.

\subsubsection{Structure Identification}
In some problems the accuracy provided by parameterized models is
not satisfactory. This situation is encountered in complex systems
where the limited flexibility of the parameterized model does not
yield a tolerable error. Moreover, the choice of the parameterized
model to be used depends mainly on the experience of the designer
and is a matter of personal judgement.

The limited flexibility of the parameterized models and lack of an
objective method for selecting such a model makes \ac{CI} methods a
strong contender. \ac{CI} based techniques such as \ac{GP} offer
high degree of flexibility. From a limited set of elements, the
algorithm can develop and evolve different models of different
complexities that can resemble the real system to a high degree
(black-box models).

Gary {\it et al.} used \ac{GP} to identify parts of the nonlinear
differential equations and their parameters describing a model of
fluid flow through pipes in a coupled water tank system
\cite{gray96nonlinear}. The model created using this technique gave
an accurate representation of the real system.

\subsection{Identification Models}
As mentioned in the previous section, the selection of the model set
to be used in identification is one of the most important steps in
the system identification process, and probably the hardest one. The
selection decision does not follow a straightforward path and is
subject to experience and faith in previously tested and well-known
models. These models may include, but not limited to, the following
model sets:

\subsubsection{Functional Models}
One of the first non-linear system representations was the Volterra
series representation developed by the Spanish mathematician Vito
Volterra. Analogous to Taylor series, it provides an expansion of a
dynamic, non-linear, time-invariant system. It describes the system
output as the sum of the 1$^{\text{st}}$ order, 2$^{\text{nd}}$
order, 3$^{\text{rd}}$ order \dots etc. operators, and every
operator is described with a transfer function called a Volterra
kernel. Due to its general use, it is sometimes referred to as a
non-parametric model. A non-linear system can be described by the
following Volterra series:
\begin{equation}
y(t)=\sum_{n=1}^{\infty}\int_{-\infty}^{\infty}d\tau_1 \dotsm
\int_{-\infty}^{\infty}
g_n(\tau_1,\dotsm,\tau_n)\prod_{r=1}^{n}u(t-\tau_r) d\tau_n
\end{equation}
where $u(t)$ and $y(t)$ are the system input and output
respectively, $g_n$ are the Volterra kernels of the system, and
$\tau_i$ are time variables.

Volterra series representation may not be the choice for practical
non-linear systems for two reasons. First, the difficulty
encountered in practical measurement of Volterra kernels detracts
from the applicability of the technique. Second, the number of terms
required to represent a non-linear system explodes with the degree
of nonlinearity.

\subsubsection{Artificial Neural Network Models}
\acp{ANN} are a strong contender to other black-box models. With
their parallelism, adaptability, robustness, they can deal
efficiently with non-linear models, and with their repetitive
structure, they are resilient to failures, since any node can play
any other node's role by adjusting its weights
\cite{sjoberg94neural}. The \ac{RBF} and the \ac{MLP} networks are
among the most widely used \ac{ANN} schemes in system
identification.

\ac{RBF} networks consist of an input layer, one hidden layer with
\ac{RBF} activation function, and a linear output layer. The most
popular \ac{RBF} activation function takes the gaussian form:
\begin{equation}
\varphi_i(\v{x})=\exp\left(\frac{-\|\v{x}-\v{c}_i\|^2}{\sigma_i}\right)
\end{equation}
where $\varphi_i$ is the activation function excited by an input
pattern $\v{x}$, $\v{c}_i$ is a vector defining the center of the
\ac{RBF} $\varphi_i$, and $\sigma_i$ is a scaling factor for node
$i$.

After the input pattern is fed through the input layer, a hidden
node only responds to a pattern $\v{x}$ within a certain distance
(Euclidian sense) from its center $\v{c}_i$. The excitation values
of the hidden nodes are then passed to the output layer. The
weighted sums of these values are produced by the output layer to
make the model output according to the following rule:
\begin{equation}
h_j(\v{x})=\sum_{i=1}^m w_{ij} \; \varphi_i(\v{x})
\end{equation}
where $h_j$ is the output produced at the output node $j$, $w_{ij}$
is the weight between the hidden node $i$ and output node $j$.

The identification process involves tuning the weights $w_{ij}$, the
centers $\v{c}_i$, and the scaling factors $\sigma_i$, where
$i=(1,\dotsc,m)$, and $j=(1,\dotsc,n)$, so that the model outputs
$h_j(\v{x})$ would match the real system outputs $y_j(\v{x})$.

Training the network to optimize $c_i$ and $\sigma_i$ is done using
an unsupervised technique, such as the Hebbian and competitive
learning rules, while the optimization of the weights $w_{ij}$ is
done by a supervised method such as the following rules:
\begin{gather}
w_{ij}^{t+1} = w_{ij}^t + \Delta w_{ij}\\
\Delta w_{ij}= \eta \; (y_j(\v{x}) - h_j(\v{x})) \; \varphi_i(\v{x})
\end{gather}
where $w_{ij}^{t+1}$ is the new synaptic weight value, and $\eta$ is
the learning rate.

Alternatively, \ac{MLP} networks may be used instead of \ac{RBF}. In
this case, the network may contain more than one hidden layer and
the activation function may take other forms such as the logistic or
sigmoidal function.

\ac{RBF} networks are known to be local learning networks because
they can develop a good understanding of region when given few
learning data sets in that region, however, they cannot generalize.
Henceforth, they are suitable when few data sets are available, and
only the neighborhood of these sets is of interest (interpolation
applications). On the other hand, \ac{MLP} networks are global
learning networks. They do a better job in generalization, but they
need a high number of data sets to learn (extrapolation
applications).

\acp{ANN} are prone to over-fitting, they may even fit to the noise
accompanying the input leading to prediction far from the training
set data. Even in the absence of noise, \ac{MLP} may suffer
over-fitting and produce wild predictions.

Among its applications in system identification, \ac{ANN} was used
for lung cancer cell identification \cite{yuan-lung}, aerodynamic
identification \cite{raedt96aerodynamic}, sensor data fusion
modeling \cite{whittington90application}, and micromachined
accelerometers identification \cite{gaura99neural}.

\subsubsection{Fuzzy Modeling}
Fuzzy sets and fuzzy logic are used to incorporate experts'
knowledge in control and industrial applications. By reversing this
process, the data acquired from a system can be used to develop a
knowledge of this system and \emph{identify} it.

The available knowledge of the system to be identified may come from
two sources: An input-output data set, or an expert knowledge of the
system. Generally, there are two main approaches for system
identification based on fuzzy logic by using those two sources of
information \cite{babuska-fuzzy}:

In the first approach, the qualitative expert knowledge is
transformed into a set of if-then rules to build a model structure.
Then the parameters of this structure, such as the membership
functions, are tuned using the input-output data set. Since a fuzzy
model can be seen as a layered structure, similar to \ac{ANN},
standard learning algorithms can be applied to it as well.

In the second approach, the data are used to construct the structure
the model in the absence of expert knowledge initially. Later, if
such knowledge came through, they can be used to modify the rules or
create new ones.

Fuzzy modeling is appropriate for interpreting human knowledge about
the system which may be expressed in natural linguistic rules and
non-crisp sets. Moreover, this kind of modeling can process
imprecise data and deal with uncertainty. It provides a transparent
representation of the system based on a number of if-then rules
which are similar to human reasoning and can provide an intuitive
understanding of the system. On the other hand, fuzzy modeling
suffers from the curse of dimensionality.

\subsubsection{The Evolutionary Approach}
\ac{EA} and \ac{SI} algorithms are strong contender to the
previously mentioned techniques employed in system identification.
Unlike these techniques \ac{EA} can be used in structure
identification (black-box models), or in parameter identification
(white-box models) due to their robustness and flexibility explained
earlier in Chapter~\ref{ch:EAs} and~\ref{ch:SI}.

In parameter identification, the algorithm tunes the parameters of
the model until the output of the model matches the output of the
real system to a desired degree of accuracy. For example, in
identifying the values of different resistors and capacitors in an
electric motor, a model for this motor is created and the values of
these components are tuned by the algorithm until the output
produced by the model for a certain input matches the output of the
real motor fed with the same input.

In structure identification, instead of tuning the parameters of the
model, the algorithm modifies the structure of the model. This can
be achieved by selecting the terms of the Volterra series used to
model the system or by modifying the structure of the \ac{ANN} by
selecting the number of layers and nodes in each layer.

\section{Identification of an Induction Motor}
\label{ch:ACE_sec:paridentind}

Induction motors are the most widely used motors in industry because
they are simple to build and rugged, reliable and have good
self-starting capability. Due to their wide spread in industry, it
is of great importance to devise a technique that can estimate
different parameters of these motors which can't be measured
directly for different reasons. In this section, a model for the
induction motor will be created, then different \ac{EA}, \ac{SI},
and a traditional technique will be explained before being used to
identify six parameters of the motor. The results obtained using
these optimizers will be presented and analyzed in detail and a
conclusion will be reached.

\subsection{Induction Motor Model}
In this induction motor model, which is based on the model presented
in \cite{UrsemV04}, the three-phase voltages and currents are
transformed to complex notation for simplicity, and the model is
then treated as a two phase system. After the differential equations
of the system are solved, the three-phase currents are evaluated and
compared to their counterpart in the real system. The absolute value
of the difference between the estimated and measured currents over a
certain period of time indicates how well the model resembles the
real system, and henceforth, how well the estimated parameters match
their real counterparts. To create such a model, first, the
three-phase input voltages to the motor, $u_1$, $u_2$, and $u_3$,
are transformed to complex notation. The voltages applied to the
stator windings take the form:
\begin{align}
  u_s    &= u_{sd} + ju_{sq}\\
  u_{sd} &= \frac{1}{3}(2u_1 - u_2 - u_3)\\
  u_{sq} &= \frac{1}{\sqrt{3}}(u_2 - u_3)
\end{align}

The rotor voltages are not externally supplied, they are induced
from the relative rotation of the rotor with respect to the stator.
The equations describing the rate of change of the flux through the
stator and the rotor can be described by:
\begin{align}
  \dot{\psi}_s &= -R_s i_s + u_s\label{eq:CSSI_psisdot}\\
  \dot{\psi}_r - j\omega_r\psi_r &=-R_ri_r\label{eq:CSSI_psirdot}
\end{align}
Where $\psi_s$ and $\psi_r$ are the flux through the stator and
rotor windings respectively, $R_s$, $i_s$ are the stator resistance
and current respectively, and $R_r$ and $i_r$ are their counterparts
in the rotor, and $\omega_r$ is the relative speed of the rotor with
respect to the stator.

When the alternating stator voltages are applied to its windings, a
magnetic field is produced in the form of a traveling wave. This
field induces currents in the rotor windings, which in turn interact
with the traveling wave and produce torque which rotates the rotor.
This interaction between the stator and the rotor can be seen in the
following flux linkage equations:
\begin{align}
  \psi_s &= L_{sl}\;i_s + L_m(i_s + i_r)\\
  \psi_r &= L_{rl}\;i_r + L_m(i_s + i_r)
\end{align}
Where $L_{sl}$, $L_{rl}$, and $L_m$ are the stator, rotor, and
mutual inductances, respectively.

By separating $i_s$ and $i_r$, and applying them in
\eqref{eq:CSSI_psisdot} and \eqref{eq:CSSI_psirdot}, the currents
will be eliminated from those two differential equations. To solve
these equations, $\omega_r$ has to be evaluated by:
\begin{equation}
\label{eq:CSSI_wrdot}
  \dot{\omega} = \frac{3}{2JP}\;Im(\psi_s^*\:i_s)
\end{equation}
Where $J$ is the total moment of inertia for the rotor and the load, $P$ is the number of pole pairs,
and $\psi_s^*$ is the conjugate of $\psi_s$. For simplification, the
counteracting torque of the load and friction were ignored.

By substitution, the following set of differential equations are
produced and used to model the system.
\begin{align}
  \dot{\psi}_{sd} &= \frac{-R_s(L_{rl} + L_m)}{L_d}\; \psi_{sd} +
  \frac{R_sL_m}{L_d}\; \psi_{rd} + u_{sd}\label{eq:CSSI_ind_difeq1}\\
  \dot{\psi}_{sq} &= \frac{-R_s(L_{rl} + L_m)}{L_d}\; \psi_{sq} +
  \frac{R_sL_m}{L_d}\; \psi_{rq} + u_{sq}\\
  \dot{\psi}_{rd} &= \frac{-R_r(L_{sl} + L_m)}{L_d}\; \psi_{rd} +
  \frac{R_rL_m}{L_d}\; \psi_{sd} - \omega_r\psi_{rq}\\
  \dot{\psi}_{rq} &= \frac{-R_r(L_{sl} + L_m)}{L_d}\; \psi_{rq} +
  \frac{R_rL_m}{L_d}\; \psi_{sq} + \omega_r\psi_{rd}\\
  \begin{split}
  \dot{\omega}_r  &=
\frac{3}{2J}\left(\frac{\psi_{sq}(L_{rl}+L_m)}{L_d}-\frac{\psi_{rq}L_m}{L_d}
\right)\;
  \psi_{sd}\\ 
&\quad-\frac{3}{2J}\left(\frac{\psi_{sd}(L_{rl}+L_m)}{L_d}-\frac{\psi_{rd}L_m}{
L_d}\right)\;
  \psi_{sq}\label{eq:CSSI_ind_difeq5}
  \end{split}
\end{align}
where
\begin{equation}
  L_d = L_{sl}L_{rl} + L_{sl}L_m + L_{rl}L_m
\end{equation}

Given the input voltage values ($u_1$, $u_2$, and $u_3$), and the
estimated motor parameters ($R_s$, $R_r$, $L_{sl}$, $L_{rl}$, $L_m$,
and $J$), the state variables ($\psi_{sd}$, $\psi_{sq}$,
$\psi_{rd}$, $\psi_{rq}$, and $\omega_r$) can be evaluated by
solving the differential equations
\eqref{eq:CSSI_ind_difeq1}--\eqref{eq:CSSI_ind_difeq5}, and the
output, which is the three-phase current values ($i_1$, $i_2$, and
$i_3$), can be calculated by:
\begin{align}
  i_1 &= i_{sd}\\
  i_2 &= -\frac{1}{2}\;i_{sd} + j \frac{\sqrt{3}}{2}\;i_{sq}\\
  i_3 &= -\frac{1}{2}\;i_{sd} - j \frac{\sqrt{3}}{2}\;i_{sq}
\end{align}

\subsection{Algorithms}
Two types of algorithms are used here to carry out parameter
identification; Traditional algorithms and \ac{CI} techniques. The
\ac{LS} method is selected to represent the traditional algorithms, while four
other algorithms have been chosen to represent \ac{CI} techniques. The first one is a \ac{GA} which
falls under the umbrella of \acp{EA}, while the other three are
\ac{PSO} variants to represent a second category of \ac{CI}
techniques which is \ac{SI} methods.

\subsubsection{Line Search}
The \ac{LS} method is a simple deterministic local search technique.
In this method, the search space is discretized with a unit size of
$\delta_i$ for the $i^{th}$ dimension. A random point (\v{x}) is
chosen in the discretized search space and its fitness is evaluated
($f(\v{x})$), and the fitness of its neighboring points are
evaluated as well. If the fitness value of most fit neighbor is less
than or equals that of \v{x} (for a minimization problem), the
algorithm moves to this new point and evaluates the fitness of its
neighbors. But if the fitness of the most fit neighbor is higher
than that of \v{x}, the algorithm terminates.

The set of neighbors $\mathcal{N}_\v{x}$ for a point $\v{x}=(x_1,
x_2,\dotsc, x_n)$ in a $n$-dimensional space contains $2n$ unique
points. The position of each point in the set can be determined by
moving by a step of $\delta_i$ in both directions of the $i^{th}$
dimension; $\mathcal{N}_\v{x}=(x_1\pm\delta_1,
x_2,\dotsc,x_n),(x_1,x_2\pm\delta_2,\dotsc,x_n),\dotsc,(x_1,x_2,\dotsc,
x_n\pm\delta_n)$.

This algorithm contains one parameter which is the step size
$\delta_i$. For the current application, this value was set to 0.1\%
of the initialization range for the corresponding dimension as
explained in Table~\ref{tb:CSSI_para}.

The algorithm starts by randomly selecting a point in the range
shown in Table~\ref{tb:CSSI_para}. Throughout subsequent iterations,
this point is not allowed to fall below the lower limit, but is
allowed to take values beyond the upper limit.

\subsubsection{Genetic Algorithms}
The \ac{GA} used in this application is based on real value
representation as explained in Subsection~\ref{ch:EA_realrep}. The
parameters are encoded with real values during initialization to
take random values within the bounds given in
Table~\ref{tb:CSSI_para}. The real-valued representation is used to
alleviate roundoff errors in decimal-to-binary and binary-to-decimal
conversions, and to provide higher degree of accuracy. The Polyploid
model was not used here because the cost of storing extra
chromosomes and handling them during mating does not worth the
marginal benefits outlined in
Subsection~\ref{ch:EA_sec:Poly_subsec:conc}.

After initialization and fitness evaluation, a tournament selection
is used to select individuals for mating. The tournament selection
is used to decrease selection pressure and to help maintaining good
population diversity. Then a recombination operator is used to
create the offspring from the selected parents. In this application,
the \ac{SBX} operator \cite{deb95simulated} is used due to its
strong ability to produce a varied set of offspring which resemble
their parents to a certain degree defined by a parameter of this
operator ($\eta_c$).

After creating the offspring, a mutation operator is applied to the
original population, however the most fit individual is immune from
mutation. The mutation operator used here is the polynomial mutation
\cite{Raghuwanshi04:PolyMut} because it can produce mutations,
similar to those produced in binary \ac{GA}, with a parameter that
defines the severity of mutations ($\eta_m$).

The survival selection scheme used here relies on tournament
selection to reduce selection pressure and help preserve diversity.
However, instead of copying them to the mating pool, the selected
individuals were those who make the population of the next
generation. An elitism strategy is used here to ensure the survival
of the most fit individual to prevent a setback in the best found
fitness. The parameters of the \ac{GA} is presented in
Table~\ref{tb:CSSI_ind_GApara}.

\begin{table}
  \centering
  \caption[GA parameters' values]{GA parameters' values for the parameter
identification problem\label{tb:CSSI_ind_GApara}}
  \begin{tabular}{l l c}
    \hline
    Parameter   &Description        &Value\\
    \hline
    $N$         &Population size    &50\\
    $p_c$       &Crossover rate     &0.5\\
    $\eta_c$    &Crossover distribution index   &15\\
    $p_m$       &Mutation rate      &0.01\\
    $\eta_m$    &Mutation distribution index    &15\\
    $ts$        &Tournament size    &2\\
    \hline
  \end{tabular}
\end{table}

\subsubsection{Particle Swarm Optimization}
Three variants of the \ac{PSO} algorithm were used here for
parameter identification. They are based on the lbest (\ac{PSO}-l),
gbest (\ac{PSO}-g) topologies, and the \ac{C-PSO} algorithm.

The particles of the swarm of each one of the three variants are
randomly initialized within the initialization ranges of the
solution space given in Table~\ref{tb:CSSI_para}. Initial velocities
were randomly initialized as well.

Just as the case with the other algorithms, the particles were not
allowed to fly below the lower bounds of the search space but were
allowed to take any value above the upper bound. On the other hand,
the velocities were not restricted by any bound.

Based on the corresponding topology used in these variants, the
particles are affected by different neighbors and update their
positions accordingly. The parameters of those \ac{PSO} variants are
given in Table~\ref{tb:CSSI_ind_PSOpara}. It is to be noted here
that the inertia weight value for the \ac{C-PSO} algorithm
($w=1.458$) is twice as much as that for the two other topologies
because it is multiplied by a uniformly distributed random number
with a mean value of 0.5 leading to an expected value which is the
same as those for the two other topologies.

\begin{table}
  \centering
  \caption[PSO parameters' values]{PSO parameters' values for the parameter
identification problem \label{tb:CSSI_ind_PSOpara}}
  \begin{tabular}{l l l c}
    \hline
    Algorithm   &Parameter  &Description                &Value\\
    \hline
    \ac{C-PSO}  &$N$        &Swarm size                 &20\\
                &$w$        &Inertia weight             &1.458\\
                &$\chi$     &Constriction coefficient   &1\\
                &$\varphi_1$&Personal learning rate     &1.494\\
                &$\varphi_2$&Global learning rate       &1.494\\
                &$cn$       &Number of clubs            &100\\
                &$M_{avg}$  &Average membership         &10\\
                &$M_{min}$  &Min membership level       &4\\
                &$M_{max}$  &Max membership level       &33\\
    \hline
    \ac{PSO}-l  &$N$        &Swarm size                 &20\\
                &$w$        &Inertia weight             &0.729\\
                &$\chi      $&Constriction coefficient  &1\\
                &$\varphi_1$&Personal learning rate     &1.494\\
                &$\varphi_2$&Global learning rate       &1.494\\
    \hline
    \ac{PSO}-g  &$N$        &Swarm size                 &20\\
                &$w$        &Inertia weight             &0.729\\
                &$\chi      $&Constriction coefficient  &1\\
                &$\varphi_1$&Personal learning rate     &1.494\\
                &$\varphi_2$&Global learning rate       &1.494\\
    \hline
  \end{tabular}
\end{table}

\subsection{Experiments}
The five previously mentioned algorithms are used to estimate the
real parameters of an induction motor which are given in
Table~\ref{tb:CSSI_para}. It is to be noted that the stator and
rotor inductances ($L_{sl}$, $L_{rl}$) were combined in a single
variable because they are linearly dependent. All the five
optimizers used the same fitness function, which evaluates the
fitness of the solution passed to it by solving the differential
equations based on the parameters of this solution using Matlab's
\texttt{ode45} solver\footnote{This solver is based on the
Runge-Kutta (4,5) formula, the Dormand-Prince pair.} and accumulates
the error which is the difference between the estimated currents
($\hat{i}_1$, $\hat{i}_2$, and $\hat{i}_3$) and the measured
currents ($i_1$, $i_2$, and $i_3$). The error value is used as a
fitness measure.
\begin{equation}
  f(\hat{\theta}) = \int_0^T(|i_1 - \hat{i}_1| + |i_2 - \hat{i}_2| +
  |i_3 - \hat{i}_3|) dt
\end{equation}

The error was accumulated at 1000 equally spaced time samples for a
one second simulation of the start-up of the motor.

To make a fair comparison between the different optimizers, each one
of them were allowed to perform 100,000 function evaluations. So a
larger size for the population or the swarm meant lower number of
generations or iterations. Since \ac{CI} techniques are stochastic
algorithms, and the starting point of the \ac{LS} method is randomly
chosen, each one of the optimizers were run for 10 times on the same
problem and statistical results are produced to decrease the role of
chance in the obtained results.

\begin{table}
  \caption[Real values for motor parameters and their initialization
ranges]{Real values for motor parameters and their corresponding initialization
ranges\label{tb:CSSI_para}}
  \centering
  \begin{tabular}{l r@{.}l r@{.}l r@{.}l r@{.}l r@{.}l}
    \hline
                &\multicolumn{2}{c}{$R_s$}  &\multicolumn{2}{c}{$R_r$} 
&\multicolumn{2}{c}{$L_{sl} + L_{rl}$}
                  &\multicolumn{2}{c}{$L_m$}  &\multicolumn{2}{c}{$J$}\\
    \hline
    Real value   &9 &203    &6 &61      &0 &09718   &1 &6816    &0 &00077\\
    Min          &1 &0      &1 &0       &0 &002     &0 &05      &0 &00005\\
    Max         &20 &0     &20 &0       &1 &0       &5 &0       &0 &001\\
    \hline
  \end{tabular}
\end{table}

\subsection{Results}
\begin{figure}
 \centering%
  \includegraphics[width=0.8\textwidth]{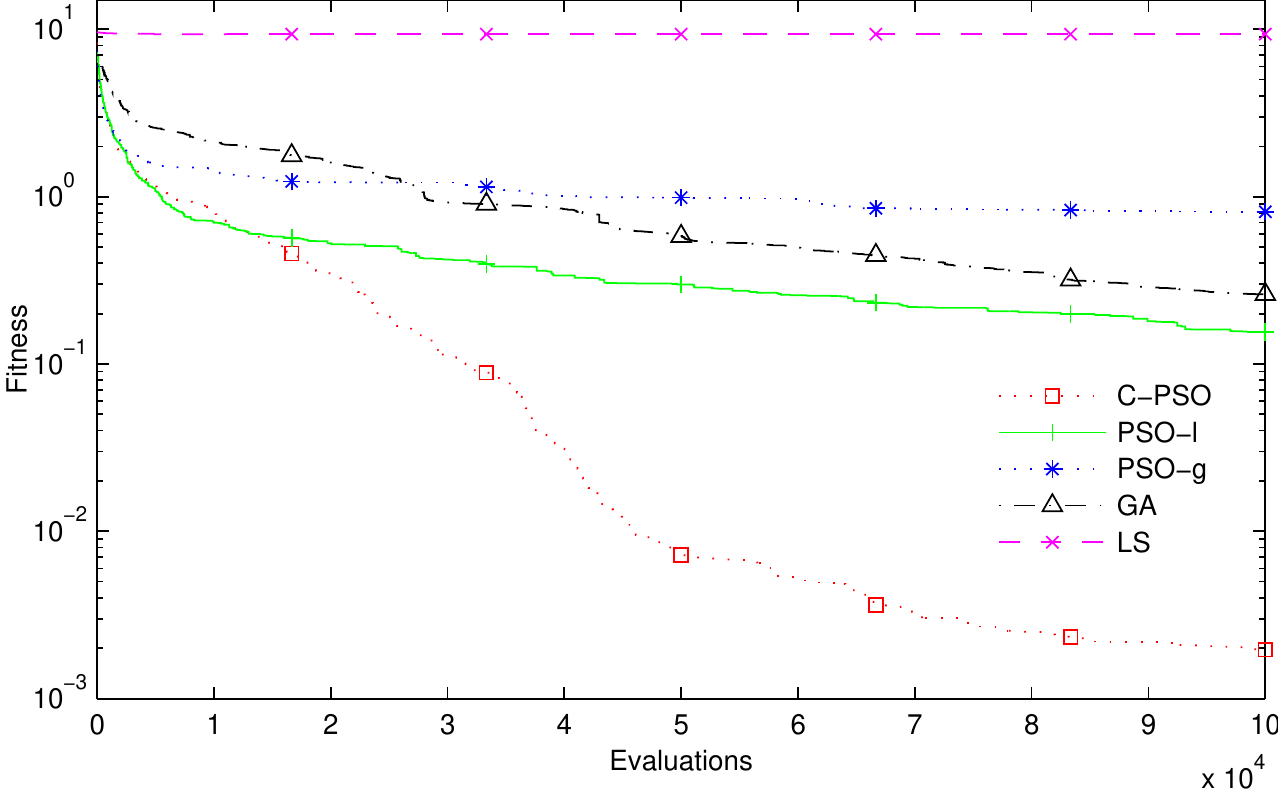}
  \caption[Fitness progress for the five optimizers]{Average fitness obtained by
the five optimizers against number of fitness function evaluations
\label{fig:CSSI_ind_err_prog}}
\end{figure}

Figure~\ref{fig:CSSI_ind_err_prog} shows the fitness values obtained
by the five optimizers through the 100,000 function evaluations of
the simulation. The values shown in the figure are the average of
the 10 independent runs for each one of the five optimizers. As can
be seen, the \ac{C-PSO} algorithm reached the lowest fitness value
among the five optimizers at the end of the 100,000 function
evaluations. It managed to reach a fitness value of 0.0019 which is
significantly better than a value of 0.1554 for the \ac{PSO}-l
algorithm which came in the second place. Little behind the
\ac{PSO}-l algorithm comes the \ac{GA} then the \ac{PSO}-g
optimizers, while the \ac{LS} algorithm lags behind them by a long
distance.

From the results obtained in Figure~\ref{fig:CSSI_ind_err_prog}, the
algorithms can be categorized into three groups according to their
performance. The first group contains the \ac{LS} algorithm which
totally lacks any capability of escaping local minima. In all the
runs, the algorithm was trapped in the first local minima it faced,
which happened very early in the run (in the first 5,000 function
evaluations in most cases), and subsequent evaluations were
unnecessary. In the second group come the \ac{PSO}-g, \ac{GA}, and
the \ac{PSO}-l algorithms. The ability of these algorithms to escape
local minima is much better than the \ac{LS} algorithm, which is
clear from their much lower final fitness values. Moreover, even at
the end of the 100,000 function evaluations, the fitness values of
these algorithms were steadily decreasing, however with a small
rate. The third group contains the \ac{C-PSO} algorithm which
outperformed all the other optimizers. This algorithm shows much
better ability than the two other groups in escaping local minima.
Moreover, the rate of fitness decrease for this algorithm is much
more higher than that of all the other algorithms, and it even
maintained a reasonable decrease rate at the end of the 100,000
function evaluations. These results show how the \ac{C-PSO}
algorithm exploits the available computation power much more better
than all the other algorithms.

Regarding convergence speed, it is clear from
Figure~\ref{fig:CSSI_ind_err_prog} that the \ac{C-PSO} algorithm is
the fastest converging algorithm. By using the number of evaluations
needed to reach a value equals 5\% of the initial fitness value
(roughly equals 10) as convergence speed measure, it can be seen
that the \ac{C-PSO} algorithm was the fastest converging algorithm
among all the optimizers. After around 15,000 function evaluations
it reached the desired fitness value which the \ac{PSO}-l algorithm
achieved after approximately 23,000 function evaluations. In the
third place comes the \ac{GA} algorithm which needed nearly 60,000
evaluations to reach that fitness value. However, neither the
\ac{PSO}-g nor the \ac{LS} algorithms achieved the fitness value in
question during the 100,000 function evaluations.

The results obtained here confirms the results obtained in a similar
parameter identification study presented in \cite{UrsemV04}, as the
\ac{PSO} algorithms (on average) achieved lower final fitness values
and higher convergence speed, which was the case here as well.

Further statistical analysis of the results is shown in
Table~\ref{tb:CSSI_ind_finfit}. As can be seen, the \ac{C-PSO}
algorithm achieves the best performance in all the five performance
measures shown in the Table. The lowest standard deviation value
achieved by the \ac{C-PSO} algorithm which is much more lower than
that of the \ac{PSO}-l algorithm which comes second shows how the
algorithm is much more reliable than all the other optimizers,
because its performance is less dependent on the stochastic
variables such as the starting point and the random weight
variables. The median value of the different independent runs is a
good representative of these runs because it is not affected by the
outlier values when compared to their average or mean value. For
this measure, the \ac{C-PSO} algorithm achieved the lowest fitness
value among all the optimizers as well.

\begin{table}
  \centering%
  \caption[Final fitness values]{Final fitness values after 100,000 function
evaluations \label{tb:CSSI_ind_finfit}}
  \begin{tabular}{l| r r r r r}
    \hline
    Algorithm   &Average    &Std. dev.  &Median &Min.   &Max.\\
    \hline
    C-PSO       &\bf{0.0019}    &\bf{0.0035}    &\bf{0.0003}    &\bf{2.5e-5}   
&\bf{0.0114}\\
    PSO-l       &0.1554         &0.1679         &0.0842         &5.6e-4        
&$\phantom{1}$0.5244\\
    PSO-g       &0.8125         &1.3091         &0.4261         &6.5e-4        
&$\phantom{1}$4.5732\\
    GA          &0.2607         &0.2250         &0.1735         &2.2e-2        
&$\phantom{1}$0.7459\\
    LS          &9.3531         &0.5663         &9.1654         &8.7e-0        
&10.4027\\
    \hline
  \end{tabular}
\end{table}

The \ac{C-PSO} algorithm continues to show superior performance over
the other algorithms regarding the average percentage deviation of
the estimated parameters from the actual induction motor parameters
which are shown in Table~\ref{tb:CSSI_ind_para_avg_er}. It achieves
much more lower deviation values in three out of the five parameters
(by an order of 100 in some situations), and comes second regarding
the other two parameters. As can be seen, the \ac{LS} algorithm did
a bad job in searching for the real parameter values. The lowest
deviation error it achieved, which is approximately 19\%, is an
unacceptable error in most real applications (above 5\% deviation
error\footnote{The Danish pumps manufacturer, Grundfos, which the
provided induction motor model is based on one of the motors they
produce, has set a tolerable deviation error value of 5\% using
conventional measurement techniques}). This deviation error value
reached a staggering value of 467\% in the case of the identified
inertia value ($J$). Second worst comes the \ac{PSO}-g algorithm
which achieved deviation errors ranging between 7.8\% and 25.9\%.
Although these values are better than those obtained by the \ac{LS}
algorithm, they are still unacceptable. Next come the \ac{PSO}-l and
the \ac{GA} algorithms showing similar performance, however the
\ac{PSO}-l is slightly better as it achieves lower deviation error
values in three out of the five parameters being identified. Those
two algorithms achieves a tolerable deviation error tolerance (below
5\%) in all the parameters. Ahead of all the other optimizers comes
the \ac{C-PSO} algorithm achieving a deviation error lower than 2\%
in all five parameters being identified.

\begin{table}
  \centering%
  \caption[Average percentage deviation for the estimated parameters]{Average
percentage deviation of the estimated parameters from the real
parameters\label{tb:CSSI_ind_para_avg_er}}
  \begin{tabular}{l| r r r r r}
    \hline
    Algorithm   &$R_s\phantom{0}$  &$R_r\phantom{0}$  &$L_{sl} + L_{rl}$ 
&$L_m\phantom{0}$  &$J$\phantom{00}\\
    \hline
    C-PSO       &\bf{0.024} &1.323      &\bf{0.652} &\bf{0.029} &1.684\\
    PSO-l       &1.976      &\bf{1.169} &3.051      &2.188      &2.814\\
    PSO-g       &17.111     &16.205     &25.889     &7.849      &17.698\\
    GA          &3.105      &2.517      &3.349      &0.051      &\bf{0.939}\\
    LS          &19.049     &63.178     &103.480    &28.869     &467.136\\
    \hline
  \end{tabular}
\end{table}

\begin{figure}
  \centering%
  \subfloat[C-PSO]{\label{fig:CSSI_C-PSO_ind_box}
    \begin{overpic}[width=0.3\textwidth]{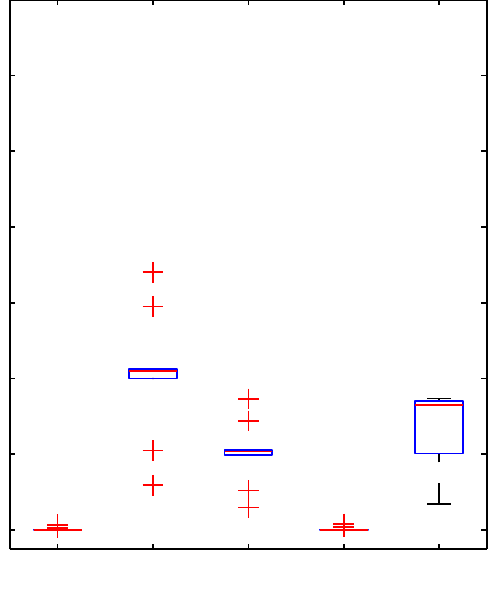}
      \put(6, 1){$R_s$}
      \put(22, 1){$R_r$}
      \put(33, 1){\scriptsize{$L_{sl} + L_{rl}$}}
      \put(56, 1){$L_m$}
      \put(71, 1){$J$}
      \put(-14, 40){\rotatebox{90}{\% error}}
      \put(-8,  8){$\phantom{1}$0}
      \put(-8, 21){$\phantom{1}$2}
      \put(-8, 34){$\phantom{1}$4}
      \put(-8, 47){$\phantom{1}$6}
      \put(-8, 60){$\phantom{1}$8}
      \put(-8, 72){10}
      \put(-8, 85){12}
    \end{overpic}}
  \subfloat[PSO-l]{\label{fig:CSSI_PSOl_ind_box}
    \begin{overpic}[width=0.3\textwidth]{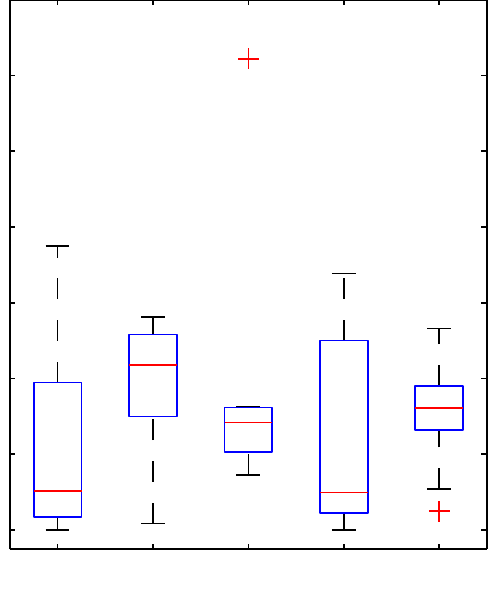}
      \put(6, 1){$R_s$}
      \put(22, 1){$R_r$}
      \put(33, 1){\scriptsize{$L_{sl} + L_{rl}$}}
      \put(56, 1){$L_m$}
      \put(71, 1){$J$}
    \end{overpic}}
  \subfloat[GA]{\label{fig:CSSI_GA_ind_box}
    \begin{overpic}[width=0.3\textwidth]{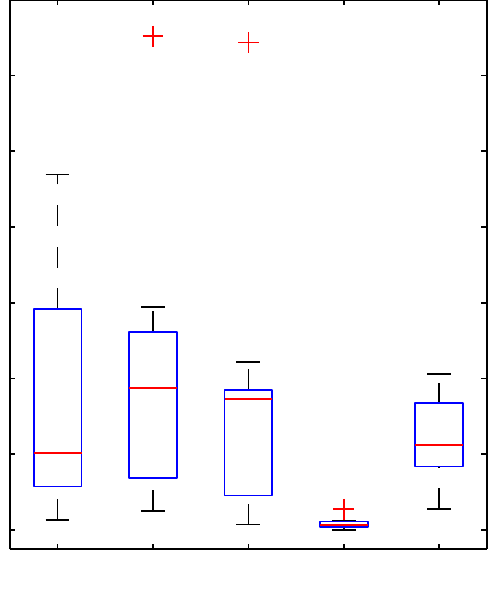}
      \put(6, 1){$R_s$}
      \put(22, 1){$R_r$}
      \put(33, 1){\scriptsize{$L_{sl} + L_{rl}$}}
      \put(56, 1){$L_m$}
      \put(71, 1){$J$}
    \end{overpic}}

  \subfloat[PSO-g]{\label{fig:CSSI_PSOg_ind_box}
    \begin{overpic}[width=0.3\textwidth]{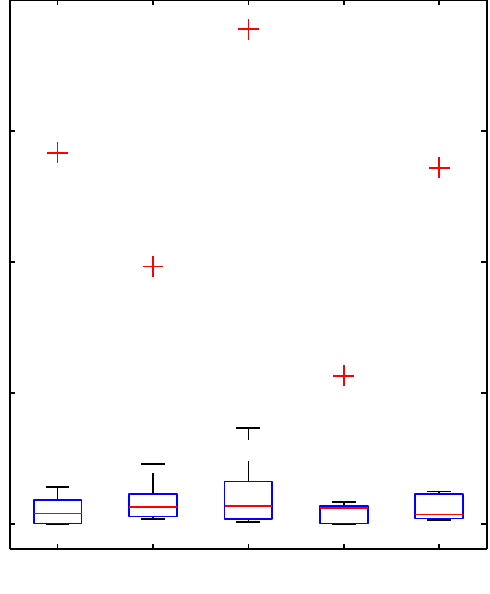}
      \put(6, 1){$R_s$}
      \put(22, 1){$R_r$}
      \put(33, 1){\scriptsize{$L_{sl} + L_{rl}$}}
      \put(56, 1){$L_m$}
      \put(71, 1){$J$}
      \put(-20, 40){\rotatebox{90}{\% error}}
      \put(-11, 9){$\phantom{15}$0}
      \put(-11, 31){$\phantom{1}$50}
      \put(-11, 52){100}
      \put(-11, 75){150}
    \end{overpic}}
  \subfloat[Fitness]{\label{fig:CSSI_errors_ind_box}
    \begin{overpic}[width=0.3\textwidth]{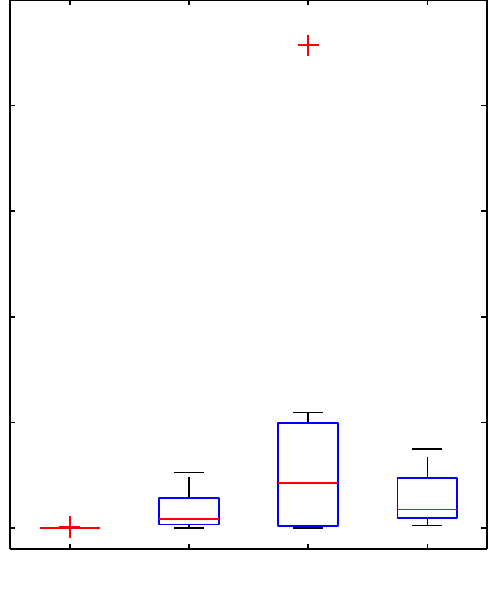}
      \put(6, 1){\tiny{C-PSO}}
      \put(26, 1){\tiny{PSO-l}}
      \put(48, 1){\tiny{PSO-g}}
      \put(69, 1){\tiny{GA}}
      \put(95, 65){\rotatebox{-90}{fitness}}
      \put(84, 9){0}
      \put(84, 26){1}
      \put(84, 44){2}
      \put(84, 62){3}
      \put(84, 80){4}
    \end{overpic}}
  \caption[Boxplots of the percentage error in the model parameters]{%
    Boxplots showing the performance of the CI algorithms: %
   
\protect\subref{fig:CSSI_C-PSO_ind_box}--\protect\subref{fig:CSSI_PSOg_ind_box}
    show the obtained percentage error in the parameters being tuned, while %
    \protect\subref{fig:CSSI_errors_ind_box} shows the fitness at the end of the
run. The error of the LS algorithm %
    was huge and henceforth was neglected in
\protect\subref{fig:CSSI_errors_ind_box} %
    to maintain a proper scale in the plot \label{fig:CSSI_ind_box}}
\end{figure}
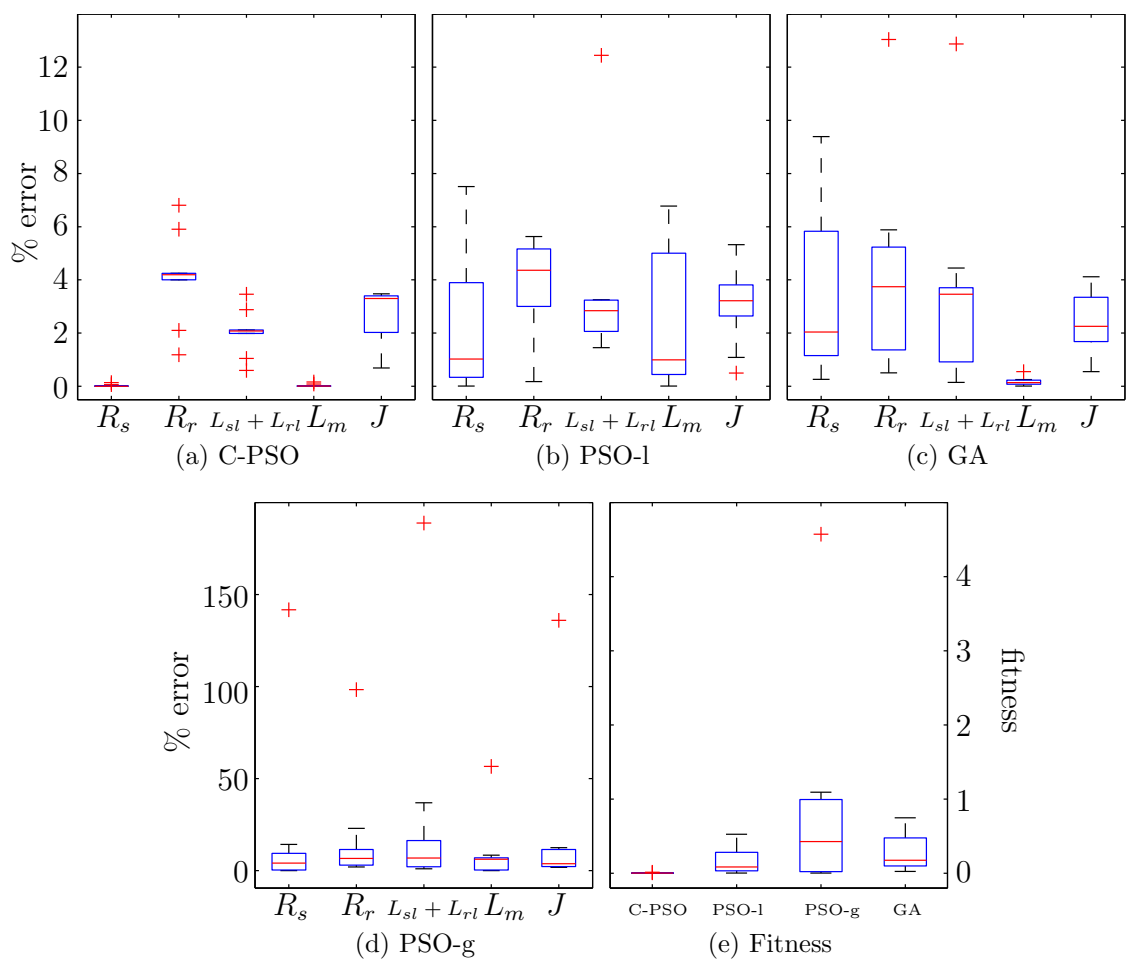

Further statistical analysis of the obtained results is presented in
Figure~\ref{fig:CSSI_C-PSO_ind_box} using boxplots. First,
Figures~\ref{fig:CSSI_C-PSO_ind_box}\subref{fig:CSSI_C-PSO_ind_box}--\subref{fig:CSSI_PSOg_ind_box}
show statistical data regarding the estimated parameters. As can be
seen,
Figure~\ref{fig:CSSI_C-PSO_ind_box}\subref{fig:CSSI_C-PSO_ind_box}
shows how the \ac{C-PSO} had more restricted outliers (lower
deviation from the mean) when compared to the other algorithms shown
in Figure~\ref{fig:CSSI_C-PSO_ind_box}\subref{fig:CSSI_PSOl_ind_box}--\subref{fig:CSSI_PSOg_ind_box}.
Moreover, the deviation from the mean is graphically shown to be
less in the case of the \ac{C-PSO} algorithm than in the case of the
other optimizers.
Figure~\ref{fig:CSSI_ind_box}\ref{fig:CSSI_errors_ind_box} shows
statistical data regarding the final fitness values obtained by the
three \ac{CI} techniques, again, \ac{C-PSO} is shown to show
superior performance; The obtained results are very close to the
mean value and there are no outliers compared to the other \ac{CI}
techniques with higher deviation from the mean and more outliers.

\begin{figure}
  \centering%
  \includegraphics[width = 0.8\textwidth]{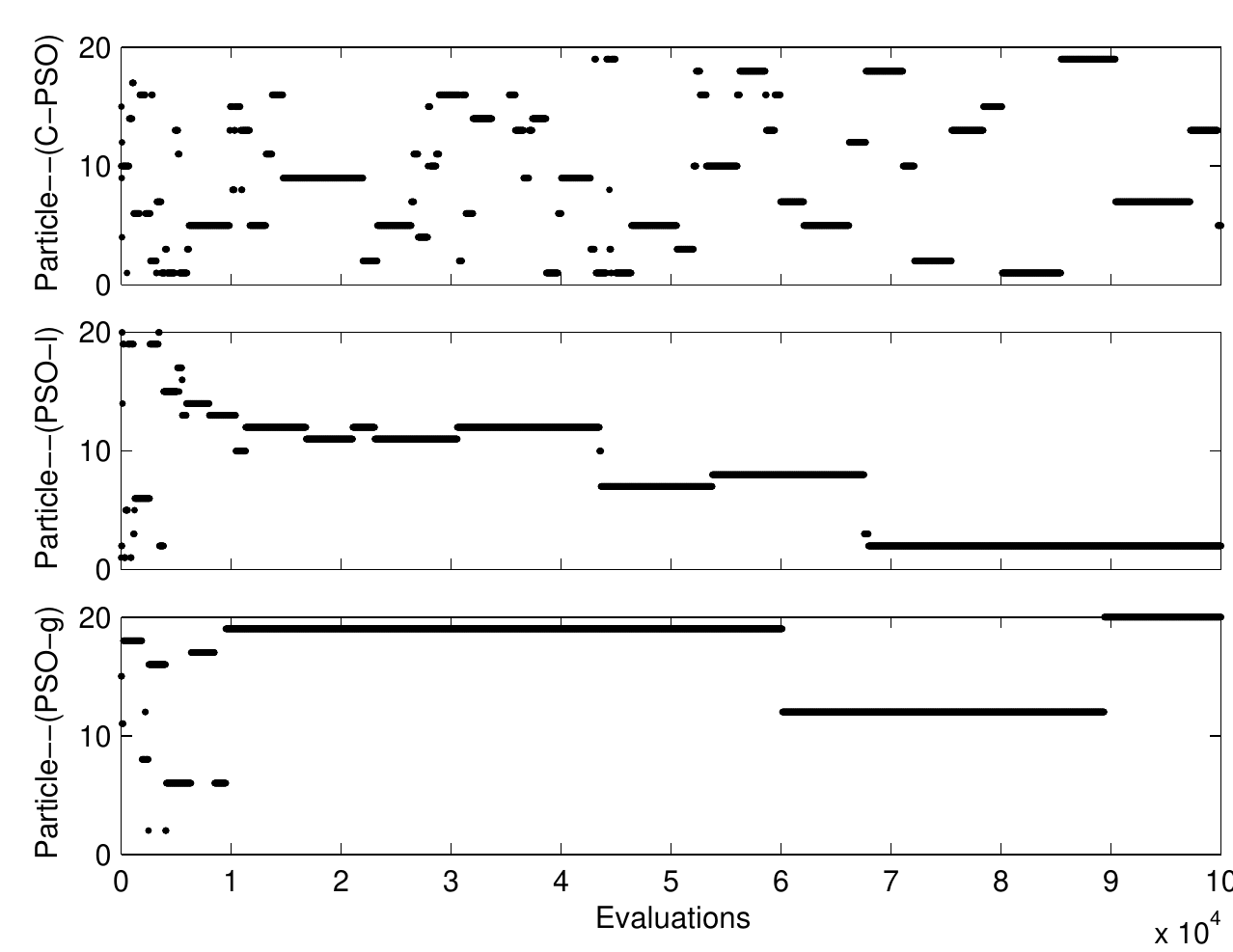}
  \caption[Best particle in swarm]
  {Best particle in the swarm of the three PSO algorithms used in parameter
identification, C-PSO (top), PSO-l (middle), and PSO-g (bottom)
\label{fig:CSSI_C-PSO_bstind}}
\end{figure}

Similar to the analysis done previously in
Subsection~\ref{ch:SI_sec:C-PSO_sub_furinv}, the index values of the
best performing particle in the swarm for the three \ac{PSO}
algorithms is presented in Figure~\ref{fig:CSSI_C-PSO_bstind}. As
can be seen, most of the particles in the \ac{C-PSO} swarm
participated in the search process as status of the best particle in
the swarm was alternating among almost all the particles
(Figure~\ref{fig:CSSI_C-PSO_bstind} up). On the other hand, the
status of the best particle in the \ac{PSO}-l algorithm was confined
to fewer particles (Figure~\ref{fig:CSSI_C-PSO_bstind} middle), and
each one of them claimed that status for a longer period of time (on
average) than the case of the \ac{C-PSO} algorithm. The effect of
the ring topology is clear in this case as best particle status
moves from a particle to its neighbor in the ring. Finally the
behavior of the particles of the \ac{PSO}-g algorithm is shown in
(Figure~\ref{fig:CSSI_C-PSO_bstind} bottom). Only three particles
(\#13, \#19, and \#20) were leading the swarm in the last 90,000
function evaluations.

The alternation of the best particle status as depicted in the
\ac{C-PSO} case shows that most of the particles of the swarm
participated effectively in the search process; While some particles
are searching for the global optimum in one region, the other
particles are searching for that optimum elsewhere, but are guided
by the experience of the other particles in the swarm. This
effective search mechanism was present but with less efficiency in
the case of the \ac{PSO}-l algorithm, and this efficiency is much
more less in the case of the \ac{PSO}-g as few particles are
effectively searching for the global minimum which the others are
being \emph{dragged} by them.

\subsection{Conclusions}
The problem of parameter identification presents quite a challenge
for any optimizer due to its multimodality and the availability of
constraints. \ac{CI} techniques are specifically made for such
challenging tasks. With their ability to escape local minima and
their problem independence they can efficiently exploit the
computation power to find the global optimum solution. This superior
ability was clear in the presented parameter identification problem
as all the \ac{CI} techniques outperformed the traditional \ac{LS}
techniques used by a big margin.

Among the different \ac{CI} techniques, two types of algorithms were
used. The \ac{SI} techniques represented by the \ac{PSO} algorithm
and the \ac{EA} techniques represented by a \ac{GA}. On average, the
\ac{PSO} variants outperformed the \ac{GA} regarding final fitness
values and convergence speed. One reason for the superiority of the
\ac{PSO} algorithms compared to the \ac{GA} is its ability to
maintain a diverge set of solutions. This was clear when the
population of solutions were monitored during the run of the
algorithms. The inertia weight used kept the particles reasonable
distanced from each other while still being guided by each other.
However in the case of the \ac{GA} optimizer, the most fit
individual took-over the population in a relatively small number of
generations, and caused most of the  population individuals to be
almost clones of itself, which badly affected the search process. A
possible remedy for this quick take-over effect is increasing the
crossover and mutation distribution indexes, however case must be
taken because the higher the values of these two parameters the more
the algorithm becomes a random search optimizer.

Comparing the different topologies used in the \ac{PSO} variants, it
was found that the \ac{C-PSO} topology achieved better results
regarding all performance metrics used. Further analysis of the
obtained results revealed that the dynamic nature of the
neighborhood employed in the \ac{C-PSO} topology resulted in a more
efficient search mechanism that employed the efforts of all
particles to search different regions of the search space instead of
following few particles or even one particle.

%% file: conc/conc.tex
\chapter{Conclusions}
\label{ch:conc}

The research presented in this thesis examined the efficiency of
\ac{CI} techniques and some proposed extensions for them. First, it
extends the simple \ac{GA} model and proposes a more biological
sound representation which mimics the structure of many living
organisms' chromosomes. This representation which adds more
redundant chromosomes to the simple---single-chromosome
representation produced marginal benefits regarding diversity of
solutions as the number of these redundant chromosomes increased.
However the convergence speed deteriorated at the meantime. It was
found that the algorithm was effectively optimizing the redundant
chromosomes as-well-as the primary ones, and this has caused the
slow progress of the optimization process. Although the Polyploid
algorithm used outperformed the \ac{NSGA-II} algorithm, it was clear
that the multi-chromosomal representation did not lead to this
superior performance because the less the number of redundant
chromosomes, the more the algorithm approaches the
simple---single-chromosome representation, the more the performance
improves.

An extension to the \ac{PSO} algorithms was presented to address
some deficiencies in current models. The proposed dynamic
neighborhood structure, in some sense, acts as an intermediate
solution to two currently well-known static neighborhood structures.
However, the results obtained showed that this proposed dynamic
neighborhood offers more than a simple compromise solution between
two extremes. The simulations on benchmark problems showed that the
proposed \ac{C-PSO} topology outperforms the other two structures
even in the characteristics they excel in. It was found that the
dynamic structure used in the \ac{C-PSO} topology makes better use
of the available computation power because almost all the particles
took part in the optimization process which was not the case in the
other two topologies.

The challenging, highly nonlinear, and multimodal problem of
induction motor's parameter identification exploits the good
characteristics of the \ac{C-PSO} algorithm. When different
optimizers were used for this problem, the results obtained
sustained previous findings, and offered even more. The \ac{C-PSO}
algorithm outperformed all the other algorithms, including the
\ac{GA}, in all performance measures used by a big margin. Not only
did the \ac{C-PSO} algorithm provide a higher convergence speed, it
also provided much lower percentage deviation of the identified
parameters from the real ones, and higher reliability as well.

A broad look on the algorithms showed that \ac{PSO} techniques, in
general, provided higher convergence speed, and lower error in the
identified parameters, however, the gbest topology presented an
exception to this rule. The gbest topology, which is known for its
fast convergence speed, was not a good choice for such a highly
multimodal problem. This topology leads to premature convergence and
causes the algorithm to get trapped very early in the run in a bad
local minima, which was not the case with the lbest and \ac{C-PSO}
topologies.

A reason for the relatively inferior performance of the \ac{GA} was
its low take-over time. The algorithm causes the population to lose
its diversity and converge rapidly to a single individual although
many precautions were taken to alleviate this well-known deficiency
of the \ac{GA} optimizer. On the other hand, the diversity of
solutions maintained by the \ac{C-PSO} and lbest topologies were
their winning horse, it allowed them to explore new regions and
avoid stagnation even at late stages of the search process.

Artificial intelligence models, and in particular, \ac{CI} models
show promising results in the system identification problem. The
results reported in this research answer some questions regarding
the suitability of some algorithms to some techniques, however, at
the same time they raise many questions to be answered by future
research. What could be a possible remedy to the innate deficiency
of the \ac{GA} technique regarding its low take-over time? Should
the parents be specifically matched instead of being randomly
assigned to each other? Since the dynamic neighborhood structure
used in this research for the \ac{PSO} algorithm resulted in such a
significant performance improvement, should other aspects of the
algorithm such as inertia weight and learning rates be made dynamic
as well? What about dynamic swarm size and multiple swarms? The
volume of questions exceeds the limited space of this thesis and
spurs more research in the applications and development of the
interesting field of \ac{AI}.

%% file: app/im.tex
\chapter{Induction Motor Model Derivation}

A space phasor $\v{x}$ is described as:
\begin{equation}
 \label{im:1}
 \underline{x} = x_d + jx_q
\end{equation}

Then for a reference frame rotating with speed $\omega_e$, the motor stator and rotor voltage equations. are:

\begin{equation}
\label{im:2}
\begin{aligned}
 \underline{v}_s^e &= r_s \underline{i}_s^e + \dot{\underline{\psi}}_s^e + j\,\omega_e \underline{\psi}_s^e\\
 \underline{v}_r^e &= 0 = r_r \underline{i}_r^e + \dot{\underline{\psi}}_r^e + j\,(\omega_e - \omega_m) \underline{\psi}_r^e
\end{aligned}
\end{equation}
Where $\omega_m$ is the electrical rotor speed.

To simplify \eqref{im:2}, we select a stator reference frame (superscript $s$) which is fixed to the machine stator, i.e. $\omega_e = 0$, to get:

\begin{equation}
\label{im:3}
\begin{aligned}
 \dot{\psi}_{sd}^s &= -r_s i_{sd}^s + v_{sd}^s\\
 \dot{\psi}_{sq}^s &= -r_s i_{sq}^s + v_{sq}^s\\
 \dot{\psi}_{rd}^s &= -r_r i_{rd}^s - \omega_m \psi_{rq}^s\\
 \dot{\psi}_{rq}^s &= -r_r i_{rq}^s - \omega_m \psi_{rd}^s
\end{aligned}
\end{equation}

Using
\begin{equation}
\label{im:4}
\begin{aligned}
 \underline{\psi}_s &= (L_{sl} + L_m) \underline{i}_s + L_m \underline{i}_r\\
 \underline{\psi}_r &= L_m \underline{i}_s + (L_{rl} + L_m) \underline{i}_r
\end{aligned}
\end{equation}
We get
\begin{equation}
\label{im:5}
\begin{aligned}
 \underline{i}_s &= \frac{L_{rl} + L_m}{L_d} \underline{\psi_s} - \frac{L_m}{L_d}\underline{\psi}_r\\
 \underline{i}_r &= -\frac{L_m}{L_d} \underline{\psi}_s + \frac{L_{ls} + L_m}{L_d} \underline{\psi_r}\\
 L_d &= L_{sl}L_{rl} + L_m (L_{sl} + L_{rl})
\end{aligned}
\end{equation}

Using ($d$ - $q$) components of $\underline{\psi}_s$, $\underline{\psi}_r$ to express ($d$ - $q$) components of $\underline{i}_s$, $\underline{i}_r$ according to \eqref{im:5}, then \eqref{im:3} becomes:

\begin{equation}
 \label{im:6}
 \begin{aligned}
  \dot{\psi}_{sd}^{s} &= -r_s x_1 \psi_{sd}^s + r_s \beta \psi_{rd}^s + v_{sd}^s\\
  \dot{\psi}_{sq}^{s} &= -r_s x_1 \psi_{sq}^s + r_s \beta \psi_{rq}^s + v_{sq}^s\\
  \dot{\psi}_{rd}^{s} &= -r_r x_2 \psi_{rd}^s + r_r \beta \psi_{sd}^s + \omega_m \psi_{rq}{s}\\
  \dot{\psi}_{rq}^{s} &= -r_r x_2 \psi_{rq}^s + r_r \beta \psi_{sq}^s + \omega_m \psi_{rd}{s}\\
  x_1 &= \frac{L_{rl} + L_m}{L_d}, \quad x_2 = \frac{L_{ls} + L_m}{L_d}\\
  \beta &= \frac{L_m}{L_d}
 \end{aligned}
\end{equation}

To get d.e. needed to solve for $\omega_m$, the generated electromagnetic torque is given by
\begin{equation}
 \label{im:7}
 T_e = \frac{3P}{2}(i_{sq} \psi_{sd} - i_{sd} \psi_{sq})
\end{equation}
where
\begin{equation}
 \label{im:8}
 \begin{aligned}
  i_{sd} = x_1 \psi_{sd} - \beta \psi_{rd}\\
  i_{sq} = x_1 \psi_{sq} - \beta \psi_{rq}\\
  P = \text{number of pole pairs}
 \end{aligned}
\end{equation}
and the motor mechanical equation is :
\begin{equation}
 \label{im:9}
 J \frac{d \omega'_m}{dt} = T_e - T_L - T_D
\end{equation}
where\\
\begin{tabular}{l@{:}l}
$\omega'_m$ 	&rotor speed in mechanical radian per sec.\\
$T_L$		&load torque\\
$T_D$ 		&damping torque
\end{tabular}

Using $\omega'_m = \frac{2}{p}\omega_m$ and for $T_L = T_D = 0$, we get
\begin{equation}
 \label{im:10}
 \frac{d\omega_m}{dt} = \frac{3P^2}{4J}(i_{sq} \psi_{sd} - i_{sd} \psi_{sq})
\end{equation}
And the induction motor model is represented by \eqref{im:6}, \eqref{im:8} and \eqref{im:10}.

To get $v_{sd}^s$, $v_{sq}^s$ given the line voltages $v_1$, $v_2$ and $v_3$, Park's transformation is used:

\begin{equation}
 \begin{bmatrix}
  f_d\\
  f_q\\
  f_o
 \end{bmatrix} = \frac{2}{3}
 \begin{bmatrix}
  \cos(\theta)	&\cos(\theta - 120)	&\cos(\theta + 120)\\
  -\sin(\theta)	&-\sin(\theta - 120)	&-\sin(\theta + 120)\\
  \frac{1}{2}	&\frac{1}{2}		&\frac{1}{2}
 \end{bmatrix}
 \begin{bmatrix}
  f_a\\
  f_b\\
  f_c
 \end{bmatrix}
\end{equation}
Where $f_o$ equation is added to yield a unique transformation, and $\theta$ is the electrical angle between the $d$-axis and the stator phase-$a$ axis as shown in Figure~\ref{im:axis}:

\begin{figure}
\centering
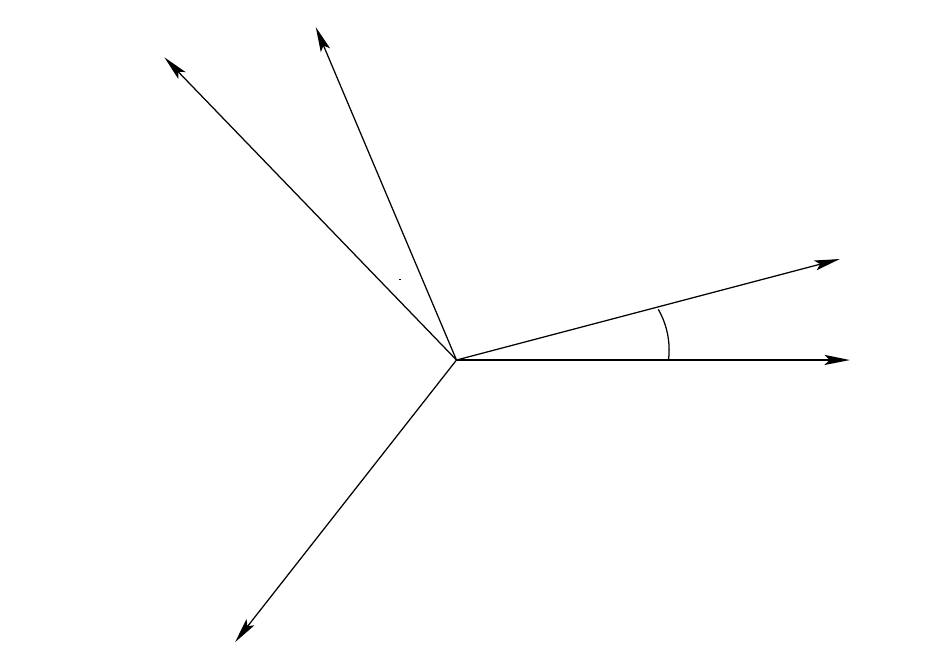
\caption{Axis transformation \label{im:axis}}
\end{figure}

The multiplier $(\frac{2}{3})$ is used to equalize the magnitude of mmf produced in $d$-$q$ frame with that of the 3-phase winding.

For $\theta = 0$, then

\begin{equation}
 \label{im:12}
 \begin{aligned}
  v_{sd}^s	&= \frac{2}{3}(v_1 + v_2\cos(-120) + v_3\cos(120))\\
	   	&= \frac{2}{3}(v_1 - \frac{1}{2}v_2 - \frac{1}{2}v_3)\\
		&= \frac{1}{3}(2v_1 - v_2 - v_3)
 \end{aligned}
\end{equation}

\begin{equation}
 \label{im:13}
 \begin{aligned}
  v_{sq}^s	&= \frac{2}{3}(-v_2\sin(-120) - v_3\sin(120))\\
		&= \frac{2}{3}(\frac{\sqrt{3}}{2}v_2 - \frac{\sqrt{3}}{2}v_3)\\
		&= \frac{1}{\sqrt{3}}(v_2 - v_3)
 \end{aligned}
\end{equation}

Also, the inverse transformation
\begin{equation*}
 \begin{bmatrix}
  f_a\\
  f_b\\
  f_c
 \end{bmatrix}=
 \begin{bmatrix}
  \cos(\theta)		&-\sin(\theta)		&1\\
  \cos(\theta - 120)	&-\sin(\theta - 120)	&1\\
  \cos(\theta + 120)	&-\sin(\theta + 120)	&1
 \end{bmatrix}
 \begin{bmatrix}
  f_d\\
  f_q\\
  f_o
 \end{bmatrix}
\end{equation*}
is used to get the line currents $i_1$, $i_2$ and $i_3$ given $i_{sd}^s$ and $i_{sq}^{s}$ by using $\theta = 0$ and $i_o = 0$,

\begin{equation}
 \label{im:14}
 \begin{aligned}
  i_1	&= i_{sd}^s\\
  i_2	&= i_{sd}^s\cos(-120) - i_{sq}^s\sin(-120)\\
	&= -\frac{1}{2}i_{sd}^s + \frac{\sqrt{3}}{2}i_{sq}^s\\
  i_3	&= -\frac{1}{2}i_{sd}^s - \frac{\sqrt{3}}{2}i_{sq}^s
 \end{aligned}
\end{equation}

and one iteration of the identification algorithm proceeds as follows:

\begin{enumerate}
 \item Initialize model parameters: $\hat{r}_s$, $\hat{r}_r$, $\hat{L}_{sl}$, $\hat{L}_{rl}$, $\hat{L}_m$ and $\hat{J}$.
 \item Get $v_{sd}^s$ and $v_{sq}^s$ given $v_1$, $v_2$ and $v_3$ using \eqref{im:12} and \eqref{im:13}.
 \item Solve \eqref{im:6}, \eqref{im:8}, \eqref{im:10} for a period of time $T$ \label{im:proc_s3}.
 \item Get $\hat{i}_{sd}^s$ and $\hat{i}_{sq}^s$ using \eqref{im:8}.
 \item Get $\hat{i}_1$, $\hat{i}_2$ and $\hat{i}_3$ using \eqref{im:14}.
 \item Evaluate the error (fitness function): $f = \int_0^T (|i_1 - \hat{i}_1| + |i_2 - \hat{i}_2| + |i_3 - \hat{i}_3|)dt$, where $i_1$, $i_2$ and $i_3$ are the true measured line currents over the interval: $t \in [0, T]$.
 \item If $f \leq e$ then stop, or else
 \item Update the model parameters ($\hat{r}_s$, $\hat{r}_r$, $\hat{L}_{sl}$, $\hat{L}_{rl}$, $\hat{L}_m$ and $\hat{J}$) using one of the optimizers mentioned in Chapter~\ref{ch:EA_pro}, then go to step~\ref{im:proc_s3}.
\end{enumerate}

An alternative model is used to represent the induction motor dynamics, especially for indirect field-oriented control applications as well as on-line identification of the rotor resistance \cite{Wang07}, is given as follows:

\begin{equation}
 \label{im:15}
 \begin{aligned}
  \dot{x} 	&= Ax + Bu\\
  x		&= [i_{sd}~ i_{sq}~ \lambda_{rd}~ \lambda_{rq}]^T\\
  u		&= [v_{sd}~ v_{sq}~ 0~ 0]^T\\
  A 		&= 
 \begin{bmatrix}
  -\gamma	&0		&\mu\:\eta	&\mu\:\omega_m\\
  0		&\gamma		&-\mu\:\omega_m	&\mu\:\eta\\
  \eta L_m	&0		&-\eta		&-\omega_m\\
  0		&\eta L_m	&\omega_m	&-\eta
 \end{bmatrix}\\
 B 		&=
 \begin{bmatrix}
  \frac{1}{\sigma L_s}\\
  \frac{1}{\sigma L_s}\\
  0\\
  0
 \end{bmatrix}
 \end{aligned}
\end{equation}

where

\begin{equation}
 \label{im:16}
 \begin{aligned}
  \eta 	&= \frac{r_r}{L_r}\\
  \sigma&= \frac{L_sL_r - L_m^2}{L_sL_r} = \text{the total leakage factor}\\
  \gamma&= \frac{1}{L_r(L_sL_r - L_m^2)}(r_sL_r^2 + r_rL_m^2)\\
  \mu	&=\frac{L_m}{L_sL_r - L_m^2}\\
  L_s	&= L_{sl} + L_m\\
  L_r	&= L_{rl} + L_m
 \end{aligned}
\end{equation}

To show how to derive this model from the previous one we start by:

\begin{equation}
 \label{im:2:3}
 \begin{aligned}
  v_{sd}	&= r_s i_{sd} + \dot{\psi}_{sd}\\
  v_{sq}	&= r_s i_{sq} + \dot{\psi}_{sq}\\
  0		&= r_r i_{rd} + \dot{\psi}_{rd} + \omega_m \psi_{rq}\\
  0		&= r_r i_{rq} + \dot{\psi}_{rq} - \omega_m \psi_{rd}
 \end{aligned}
\end{equation}

\begin{equation}
 \label{im:2:4}
 \begin{aligned}
  \underline{\psi}_s	&= L_s \underline{i}_s + L_m \underline{i}_r\\
  \underline{\psi}_r	&= L_m \underline{i}_s + L_r \underline{i}_r\\
  L_s		&= L_{sl} + L_m\\
  L_r		&= L_{rl} + L_m
 \end{aligned}
\end{equation}

Starting by:

\begin{equation*}
 \dot{\psi}_{rd} = -r_r i_{rd} - \omega_m \psi_{rq}
\end{equation*}

using

\begin{equation*}
 \begin{aligned}
  \psi_{rd} 		&= L_m i_{sd} + L_r i_{rd}\\
  \therefore i_{rd}	&= \frac{1}{L_r}\psi_{rd} - \frac{L_m}{L_r}i_{sd}\\
  \therefore \dot{\psi}_{rd}	&= \frac{r_r L_m}{L_r}i_{sd} - \frac{r_r}{L_r}\psi_{rd} - \omega_m \psi_{rq}
 \end{aligned}
\end{equation*}

and the same steps are followed to get the $\dot{\psi}_{rq}$ equation.

Now considering:

\begin{equation*}
 \dot{\psi}_{sd} = -r_s i_{sd} + v_{sd}
\end{equation*}

using \eqref{im:2:4}, we get:

\begin{equation*}
 \begin{bmatrix}
  i_{sd}\\
  i_{rd}
 \end{bmatrix}
 = \frac{1}{L_sL_r - L_m^2}
 \begin{bmatrix}
  L_r	&-L_m\\
  -L_m	&L_s
 \end{bmatrix}
 \begin{bmatrix}
  \psi_{sd}\\
  \psi_{rd}
 \end{bmatrix}
\end{equation*}
 
and

\begin{equation*}
 \begin{aligned}
  i_{sd}	&= \frac{L_r}{L_sL_r - L_m^2}\psi_{sd} - \frac{L_m}{L_sL_r - L_m^2}\psi_{rd}\\
		&= \frac{1}{\sigma L_s}\psi_{sd} - \mu\:\psi_{rd}
 \end{aligned}
\end{equation*}

differentiating, we get:

\begin{equation*}
 \begin{aligned}
  \dot{i}_{sd}	&= \frac{1}{\sigma L_s} \dot{\psi}_{sd} - \mu\:\dot{\psi}_{rd}\\
		&= \frac{1}{\sigma L_s}(-r_s i_{sd} + v_{sd}) - \mu(-r_r i_{rd} - \omega_m \psi_{rq})
 \end{aligned}
\end{equation*}

using

\begin{equation*}
 i_{rd} = \frac{1}{L_r}\psi_{rd} - \frac{L_m}{L_r}i_{sd}
\end{equation*}

then

\begin{equation}
 \dot{i}_{sd} = - \frac{r_s}{\sigma L_s}i_{sd} + \mu\:r_r(\frac{1}{L_r}\psi_{rd} - \frac{L_m}{L_r}i_{sd}) + \mu\:\omega_m\psi_{rq} + \frac{1}{\sigma L_s}v_{sd}
\end{equation}

i.e.

\begin{equation}
 \label{im:18}
 \dot{i}_{sd} = -(\frac{r_s}{\sigma L_s} + \frac{\mu\:r_rL_m}{L_r})i_{sd} + \frac{\mu\:r_r}{L_r}\psi_{rd} + \mu\:\omega_m\psi_{rq} + \frac{1}{\sigma L_s}v_{sd}
\end{equation}

where

\begin{equation}
 \label{im:19}
 \begin{aligned}
  \frac{r_s}{\sigma L_s} + \frac{\mu\:r_rL)m}{L_r} 	&= \frac{r_sL_sL_r}{(L_sL_r - L_m^2)L_s} + \frac{L_m^2r_r}{(L_sL_r - L_m^2)L_r}\\
  							&= \frac{1}{L_sL_r - L_m^2}(r_sL_r + \frac{L_m^2r_r}{L_r})\\
							&= \gamma
 \end{aligned}
\end{equation}

i.e.

\begin{equation}
 \label{im:20}
 \dot{i}_{sd} = -\gamma\: i_{sd} + \mu\:\eta\psi_{rd} + \mu\:\omega_m\psi_{rq} + \frac{1}{\sigma L_s}v_{sd}
\end{equation}

and following the same steps, we get the equation for $\dot{i}_{sq}$

Now for the torque equation, using

\begin{equation*}
 v_{sd} = \sigma L_s \dot{i}_{sd} + \gamma\: i_{sd} - \mu\:\eta\sigma L_s \psi_{rd} - \sigma L_s \omega_m\mu\:\psi_{rq}
\end{equation*}

\begin{equation}
 \label{im:21}
 \begin{aligned}
  v'_{sd}	&= -\sigma L_s \omega_m\eta\:\psi_{rq}\\
		&= -\frac{L_m}{L_r}\omega_m\psi_{rq}
 \end{aligned}
\end{equation}

where $v'_{sd}$ is the component of $v_{sd}$ contributing to the output mechanical power.

Also,

\begin{equation}
 \label{im:22}
 v'_{sq} = \frac{L_m}{L_r}\omega_m\psi_{rd}
\end{equation}

using

\begin{equation}
 \label{im:23}
 \begin{aligned}
  P_m = T\frac{\omega_m}{P}	&= \frac{3}{2}(v'_{sd}i_{sd} + v'_{sq}i_{sq})\\
				&= \frac{3}{2}\omega_m\frac{L_m}{L_r}(\psi_{rd}i_{sq} - \psi_{rq}i_{sd})
 \end{aligned}
\end{equation}

i.e.

\begin{equation*}
 \begin{aligned}
  T	&= \frac{3P}{2}\cdot\frac{L_m}{L_r}(\psi_{rd}i_{sq} - \psi_{rq}i_{sd})\\
	&= \frac{J}{P}\dot{\omega}_m + \frac{\beta}{P}\omega_m + T_L
 \end{aligned}
\end{equation*}

i.e.

\begin{equation}
 \label{im:24}
 \dot{\omega}_m = \frac{3P^2}{2J}\cdot\frac{L_m}{L_r}(\psi_{rd}i_{sq} - \psi_{rq}i_{sd}) - \frac{\beta}{J}\omega_m - \frac{P}{J}T_L
\end{equation}

where

\begin{tabular}{l@{:}l}
 $P$		&no. of pole-pairs\\
 $\omega_m$	&rotor speed in electrical rad./s\\
 $J$		&rotor moment of inertia\\
 $\beta$	&viscous damping coefficient\\
 $T_L$		&load torque\\
 $P_m$		&gross generated mechanical power
\end{tabular}

%% file: app/axis.pdf_t
\begin{picture}(0,0)%
\includegraphics{app/axis.pdf}%
\end{picture}%
\setlength{\unitlength}{3158sp}%
\begingroup\makeatletter\ifx\SetFigFont\undefined%
\gdef\SetFigFont#1#2#3#4#5{%
  \reset@font\fontsize{#1}{#2pt}%
  \fontfamily{#3}\fontseries{#4}\fontshape{#5}%
  \selectfont}%
\fi\endgroup%
\begin{picture}(5707,3906)(-374,-5224)
\put(1276,-5161){\makebox(0,0)[lb]{\smash{{\SetFigFont{11}{13.2}{\rmdefault}{\mddefault}{\updefault}{\color[rgb]{0,0,0}axis of phase $c$}%
}}}}
\put(3826,-3811){\makebox(0,0)[lb]{\smash{{\SetFigFont{11}{13.2}{\rmdefault}{\mddefault}{\updefault}{\color[rgb]{0,0,0}axis of phase $a$}%
}}}}
\put(3751,-3361){\makebox(0,0)[lb]{\smash{{\SetFigFont{11}{13.2}{\rmdefault}{\mddefault}{\updefault}{\color[rgb]{0,0,0}$\theta$}%
}}}}
\put(4126,-2686){\makebox(0,0)[lb]{\smash{{\SetFigFont{11}{13.2}{\rmdefault}{\mddefault}{\updefault}{\color[rgb]{0,0,0}$d$-axis}%
}}}}
\put(1651,-1486){\makebox(0,0)[lb]{\smash{{\SetFigFont{11}{13.2}{\rmdefault}{\mddefault}{\updefault}{\color[rgb]{0,0,0}$q$-axis}%
}}}}
\put(-374,-1561){\makebox(0,0)[lb]{\smash{{\SetFigFont{11}{13.2}{\rmdefault}{\mddefault}{\updefault}{\color[rgb]{0,0,0}axis of phase $b$}%
}}}}
\end{picture}%